%% file: main.tex
\pdfoutput=1
\documentclass[10pt,twocolumn,letterpaper]{article}
\usepackage[pagenumbers]{cvpr}
\input{preamble}
\usepackage{longtable}
\IfFileExists{placeins.sty}{\usepackage{placeins}}{\newcommand{\FloatBarrier}{\clearpage}}
\definecolor{cvprblue}{rgb}{0.21,0.49,0.74}
\usepackage[breaklinks,colorlinks,allcolors=cvprblue]{hyperref}
\ifdefined\pdfsuppressptexinfo\pdfsuppressptexinfo=-1 \fi
\input{meta}
\title{DyLoMo-DiT: Dynamic Looped Mixture-of-Experts Diffusion Transformers}
\author{\PaperAuthorBlock}
\graphicspath{{figs/}}
\makeatletter

\newcommand{\PaperLink}[1]{\if\relax\detokenize\expandafter{#1}\relax[link]\else\expandafter\url\expandafter{#1}\fi}
\providecommand{\PaperLinksLine}{Code: \PaperLink{\CodeURL}\qquad Weights: \PaperLink{\WeightsURL}\qquad Project page: \PaperLink{\ProjectURL}}
\newcommand{\PaperWeightsNote}{}
\patchcmd{\@maketitle}{\vskip .5em}{\vskip .5em{\normalsize\PaperLinksLine\par}\if\relax\detokenize\expandafter{\PaperWeightsNote}\relax\else{\small\PaperWeightsNote\par}\fi}{}{\PackageError{main}{Could not add the links line to the title block}{}}
\newcommand{\PaperAcknowledgmentsSection}{\if\relax\detokenize\expandafter{\PaperAcknowledgments}\relax\else\section*{Acknowledgments}\PaperAcknowledgments\par\fi}
\makeatother
\begin{document}
\maketitle
\begin{strip}
\centering
\includegraphics[width=\textwidth]{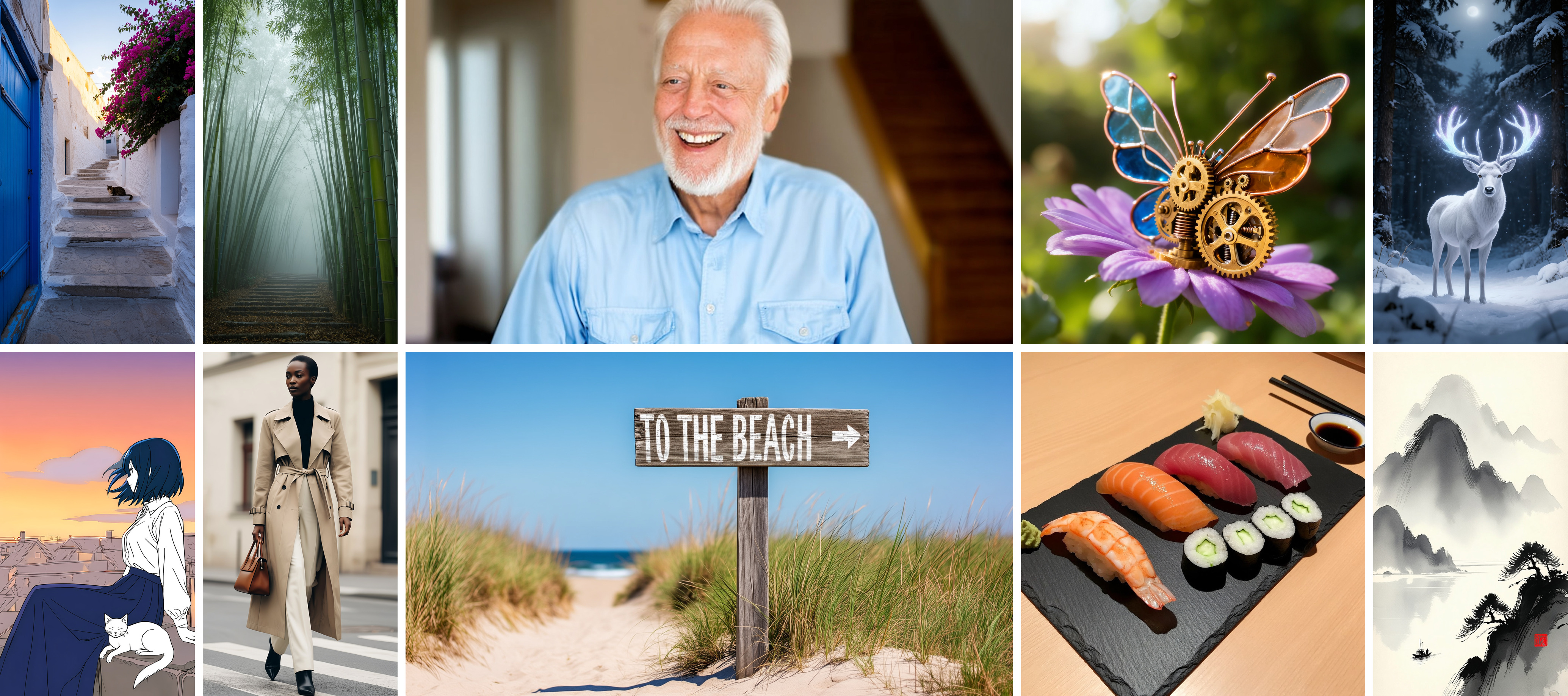}
\captionof{figure}{Selected text-to-image samples from DyLoMo-DiT (15.48B total parameters, 1.05B active per image token). Prompts are listed in \cref{sec:app_a}.}
\label{fig:head}
\end{strip}
\input{sec/00_abstract}
\input{sec/01_intro}
\input{sec/02_related}
\input{sec/03_method}
\input{sec/04_setup}
\input{sec/05_results}
\input{sec/06_conclusion}
\PaperAcknowledgmentsSection
{\small\bibliographystyle{ieeenat_fullname}\bibliography{refs}}
\appendix
\crefalias{section}{appendix}
\renewcommand{\thefigure}{\Alph{section}\arabic{figure}}
\renewcommand{\thetable}{\Alph{section}\arabic{table}}
\makeatletter\@addtoreset{figure}{section}\@addtoreset{table}{section}\makeatother
\section*{Appendix}
\noindent Sections, figures, tables and equations cited without a letter (for example Table~2 or Sec.~5.4) are those of the main paper; the ones of the appendix carry the letter of their appendix. Unless stated otherwise, ImageNet results follow the protocol of the main paper: $256\times256$, EMA weights, 250 DDPM sampling steps, guidance scale 1.5, FID over 50k images, at 400k training steps.
\input{supp/A_t2i_details}

\input{supp/A_gallery}
\FloatBarrier
\input{supp/B_params_compute}
\FloatBarrier
\input{supp/C_image_to_video}
\input{supp/D_in256_pairs}
\input{supp/E_scale_curve}
\FloatBarrier
\input{supp/F_routing_extra}
\FloatBarrier
\input{supp/G_loop_count}
\FloatBarrier
\input{supp/H_arch_details}
\FloatBarrier
\input{supp/I_ohc_full}
\FloatBarrier
\input{supp/J_ditmoe}
\FloatBarrier
\input{supp/K_flow}
\end{document}

%% file: preamble.tex
\usepackage{cuted}        %
\usepackage{currfile}
\usepackage{caption}
\usepackage{colortbl}     %
\usepackage{multirow}
\usepackage{bm}
\usepackage{siunitx}
\definecolor{tiergray}{HTML}{F0EFEC}   %
\newcommand{\tierrow}[2]{\rowcolor{tiergray}\multicolumn{#1}{l}{\textbf{#2}}\\}
\AtBeginDocument{\crefname{appendix}{App.}{Apps.}\Crefname{appendix}{App.}{Apps.}}
\newcommand{\hd}[1]{\multicolumn{1}{c}{\begin{tabular}[b]{@{}c@{}}#1\end{tabular}}}
\usepackage{microtype}  %

%% file: meta.tex
\newcommand{\PaperAuthorBlock}{Lernie Li \and Yuanliu Liu \and Pengfei Xiong \and Wei Zhang}
\newcommand{\PaperAcknowledgments}{}
\newcommand{\CodeURL}{https://github.com/OneBoxCream/DyLoMo-DiT}
\newcommand{\WeightsURL}{https://huggingface.co/OneCreamBox}
\newcommand{\ProjectURL}{https://oneboxcream.github.io/DyLoMo-DiT/}
\newcommand{\PaperIcon}[1]{\raisebox{-0.22em}{\includegraphics[height=1.15em]{#1}}}
\newcommand{\PaperLinkItem}[4]{\begin{tabular}[t]{@{}c@{}}\PaperIcon{#1}\hspace{0.4em}#2\\[0.15em]{\footnotesize\href{#3}{\nolinkurl{#4}}}\end{tabular}}
\newcommand{\PaperLinksLine}{\rule{0pt}{1.6em}%
  \PaperLinkItem{icon_github}{Code}{\CodeURL}{github.com/OneBoxCream/DyLoMo-DiT}\hspace{2.4em}%
  \PaperLinkItem{icon_hf}{Weights}{\WeightsURL}{huggingface.co/OneCreamBox}\hspace{2.4em}%
  \PaperLinkItem{icon_globe}{Project page}{\ProjectURL}{oneboxcream.github.io/DyLoMo-DiT}}

%% file: sec/00_abstract.tex
\begin{abstract}
Diffusion Transformers (DiTs) scale model capacity by increasing both parameter count and computation. Mixture-of-Experts (MoE) variants expand parameter capacity while keeping per-token computation roughly fixed, but still require substantial parameter storage.
In this paper, we propose Dynamic Looped MoE DiT (DyLoMo-DiT), which reuses parameters rather than adding them: a weight-shared loop repeatedly applies the middle layers, while sparse experts allocate parameter capacity across tokens. This design deepens the network without adding parameters, freeing the parameter budget for other uses, such as more experts or more distinct layers. It also allocates compute dynamically, varying the number of passes across noise levels. Experiments show that DyLoMo-DiT makes more effective use of parameter and compute budgets, with further gains from assigning more passes to low-noise steps.
On ImageNet $256\times256$, DyLoMo-DiT reduces FID by 11--24\% relative to DiT and 36--56\% relative to DiT-MoE at matched parameter and compute budgets. For text-to-image generation, a scaled DyLoMo-DiT achieves a higher GenEval score than FLUX.1 (0.714 vs.\ 0.673), with less than one tenth of its active parameters (1.05B vs.\ 12B).
\end{abstract}

%% file: sec/01_intro.tex
\section{Introduction}
\label{sec:intro}

\begin{figure*}[t]
  \centering
  \includegraphics[width=0.94\textwidth]{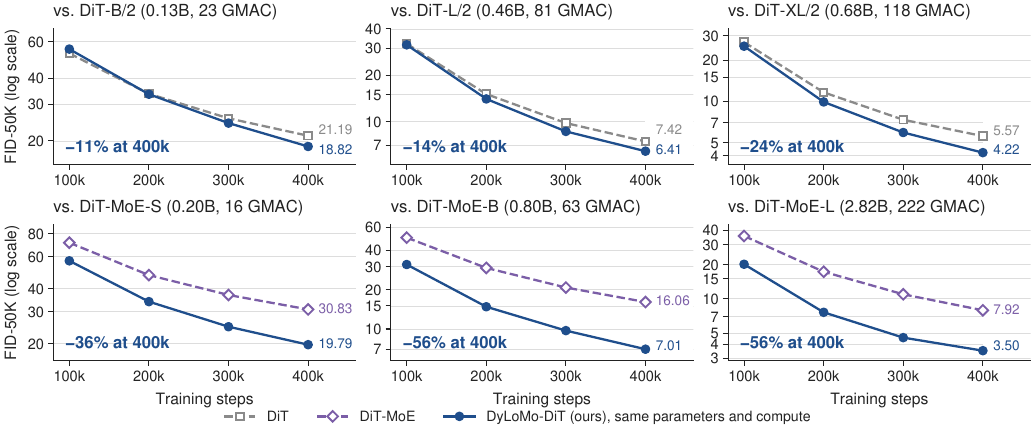}  %
  \caption{FID versus training steps on ImageNet-256: DiT (top; B, L, XL tiers) and DiT-MoE (bottom; S, B, L tiers) against the DyLoMo-DiT with the same parameters and forward compute. Percentages give the relative FID reduction at 400k steps.}
  \label{fig:teaser}
\end{figure*}

As diffusion models grow in size, scaling them to improve generation quality also increases computational and memory costs~\cite{esser2024sd3,flux2024,wan2025}. More efficient scaling therefore requires making better use of model parameters and computation, rather than simply increasing both.

DiT~\cite{peebles2023dit} applies the same dense layer stack at every denoising step. Increasing its width raises both the parameter count and per-token computation, while adding independently parameterized layers increases both depth and parameter count. Although timestep conditioning changes the network's activations, it does not change the layers or computation used at each step. MoE-DiTs such as DiT-MoE~\cite{fei2024ditmoe} use sparse experts~\cite{shazeer2017moe} to expand parameter capacity while keeping per-token computation roughly fixed, but still require substantial parameter storage. Both designs maintain uniform depth across noise levels, even though high-noise steps primarily establish global structure and low-noise steps refine detail~\cite{balaji2022ediffi}.

Depth offers another way to use model parameters more efficiently. Recurrently applying a group of weight-shared layers increases effective depth while storing only one set of parameters for the repeated layers. This reduces the parameter cost of added depth and frees budget for experts. Recurrent depth has been explored in language models and, more recently, in diffusion Transformers~\cite{goyal2026elt,chng2026loopeddit,derakhshani2026lift,wang2026loopreg}. However, recurrence alone does not determine where the additional computation is most useful: a fixed pass count applies the same extra computation at every noise level, while dense feed-forward networks lack MoE's token-dependent parameter capacity. In our matched-budget ablation, a uniform two-pass loop does not improve FID at the DiT-MoE-B budget (\cref{sec:design_ablation}). These observations motivate combining parameter-efficient recurrent depth with sparse experts and adapting computation across the denoising trajectory.

In this paper, we propose Dynamic Looped MoE DiT (DyLoMo-DiT), an architecture that jointly allocates parameter capacity, computation, and depth under a target parameter and compute budget. Its feed-forward networks use multi-head sparse experts to route each token to a small subset of experts, expanding parameter capacity across tokens while keeping per-token computation nearly fixed. A group of middle Transformer layers shares weights across recurrent passes, adding effective depth without storing a separate copy of the layer parameters. The resulting parameter savings can be reinvested in more experts. DyLoMo-DiT also uses a dynamic loop schedule that varies the number of passes with the noise level. Assigning more passes to low-noise inputs, rather than applying the same number of passes throughout denoising, brings further gains. This design differs from DiT, which uses a fixed dense layer stack, and MoE-DiT, which expands token-dependent parameter capacity but retains uniform depth across noise levels. The expert configuration, looped segment, routing heads, and pass schedule are jointly selected for the target budget.

We evaluate DyLoMo-DiT on ImageNet $256\times256$ and in text-to-image generation. At matched parameter and forward-compute budgets, DyLoMo-DiT reduces FID by 11\%, 14\%, and 24\% relative to DiT in the B, L, and XL tiers, respectively, and by 36\%, 56\%, and 56\% relative to DiT-MoE in the S, B, and L tiers under the same training protocol. The advantage persists against scaled-up baselines: a DiT with three times the parameters and compute still has 13\% higher FID, while the original DiT-MoE-XL configuration, with six times the parameters, has 63\% higher FID. Further analyses show how the joint design contributes to these gains: at the DiT-MoE-B budget, progressively adding the design choices lowers FID from 16.06 to 5.28 (\cref{sec:design_ablation}). Extra passes improve results primarily at low noise, while routers without explicit timestep inputs learn to specialize by noise level. Finally, a scaled text-to-image variant scores higher than FLUX.1 on GenEval at $1024^2$ (0.714, compared with 0.673 for FLUX.1-schnell and 0.671 for FLUX.1-dev) with less than one tenth as many active parameters (1.05B vs.\ 12B).

Our contributions are summarized as follows:
\begin{itemize}\setlength{\itemsep}{0pt}\setlength{\parskip}{0pt}\setlength{\parsep}{0pt}
    \item \textbf{Parameter-efficient recurrent depth for diffusion Transformers.} DyLoMo-DiT reuses a group of middle layers to increase effective depth without duplicating their parameters, freeing budget for sparse experts.

    \item \textbf{Dynamic loop allocation across noise levels.} DyLoMo-DiT assigns different numbers of passes along the denoising trajectory, with more passes at low noise.

    \item \textbf{Joint allocation of parameters, computation, and depth under a shared budget.} Sparse experts allocate capacity across tokens, weight sharing controls the parameter cost of added depth, and the loop schedule distributes computation across noise levels.

    \item \textbf{A scalable design space for dense and sparse-expert diffusion Transformers.} DyLoMo-DiT scales across model sizes and includes DiT, MoE-DiT, and looped DiT as special cases.
\end{itemize}

%% file: sec/02_related.tex
\section{Related Work}
\label{sec:related}
\textbf{Parameters without compute: sparse experts in Transformers.} DiT-MoE~\cite{fei2024ditmoe} introduces per-token sparse experts into DiT. Subsequent work improves expert routing or extends sparse experts to other backbones and tasks~\cite{yuan2025race,sun2025ecdit,shi2025diffmoe,wei2025promoe,jia2026mmoe,akiti2026nucleus}. DiffMoE and ProMoE report that token-choice sparse experts often underperform dense models when active parameter counts are matched. Another line of work assigns experts explicitly to different noise stages~\cite{balaji2022ediffi,feng2023ernievilg,fang2024remixdit,park2024dtr,park2024switchdit,wan2025wan22,diffmoe_timestep}. Our method does not use such an assignment, but \cref{sec:mechanism} shows that its routing learns a similar split. Per-head routing has been explored in Multi-Head Mixture-of-Experts~\cite{wu2024mhmoe}, Multi-Head LatentMoE~\cite{cui2026mhlatentmoe}, and MAGI-2~\cite{magi2}. None of these methods shares weights across depth, so depth remains tied to parameter count.

\textbf{Depth without parameters: weight-shared Transformers.}
The Universal Transformer~\cite{dehghani2019universal} applies the same transformation recurrently across depth. SMELT~\cite{wang2026smelt} compares looped and non-looped MoE language models at matched compute, parameter count, and KV cache, and recommends looping through roughly half of the layers—located in the middle of the network—for two passes, with residual updates scaled by $1/r$. For visual generation, Fixed Point Diffusion Models~\cite{bai2024fpdm} incorporate fixed-point solver layers into the denoiser. ELT, Looped DiT, LiFT, and LoopREG~\cite{goyal2026elt,chng2026loopeddit,derakhshani2026lift,wang2026loopreg}, three of which are concurrent with this work, apply loops to some or all layers of a DiT. However, they retain dense feed-forward networks, leaving parameter count tied to computation.

\textbf{Removing both couplings.}
Several language-model studies combine sparse experts with recurrence~\cite{tan2023sut,csordas2024moeut,chen2026loopmoe,gao2026loopie,wang2026smelt}. Lee~\etal~\cite{lee2026sparse} find that looped MoE models benefit more from scaling than the standard baseline, whereas looped dense models do not. Subsequent work explores using more passes, the roles of the two visits, scaling laws with respect to pass count, experts shared across layers, and per-token recursion depth~\cite{he2026loom,he2026latentloop,chen2026scalinglooped,jaggi2026tying,qiu2026more,bae2025mor}. To our knowledge, no diffusion Transformer has removed both couplings while allocating depth by noise level. This work does so under the standard DiT protocol. We evaluate multi-stream residuals~\cite{zhu2024hyper,xie2025mhc,guo2026ohc,hy4preview,ge2026chimera} separately; they are not part of our method (\cref{sec:app_i}).

%% file: sec/03_method.tex
\section{Method}
\label{sec:method}

\subsection{Three Resources and the Design Space}
\label{sec:overview}

\begin{table}
  \centering\small\setlength{\tabcolsep}{3pt}
  \caption{Ways of allocating parameters and depth that each architecture offers, tested in \cref{sec:results}. $^\dagger$Only when trained with a mixed number of passes.}
  \label{tab:freedoms}
  \begin{tabular}{@{}lccc@{}}
    \toprule
    Degree of freedom & DiT & DiT-MoE & DyLoMo-DiT \\
    \midrule
    Parameters per token & -- & \checkmark & \checkmark \\
    Depth at fixed parameters & -- & -- & \checkmark \\
    Depth by noise level & -- & -- & \checkmark \\
    Parameters by noise level & -- & \checkmark & \checkmark \\
    Depth chosen at inference & -- & -- & \checkmark$^\dagger$ \\
    \bottomrule
  \end{tabular}
\end{table}

\begin{figure*}[t]
  \centering
  \includegraphics[width=0.94\textwidth]{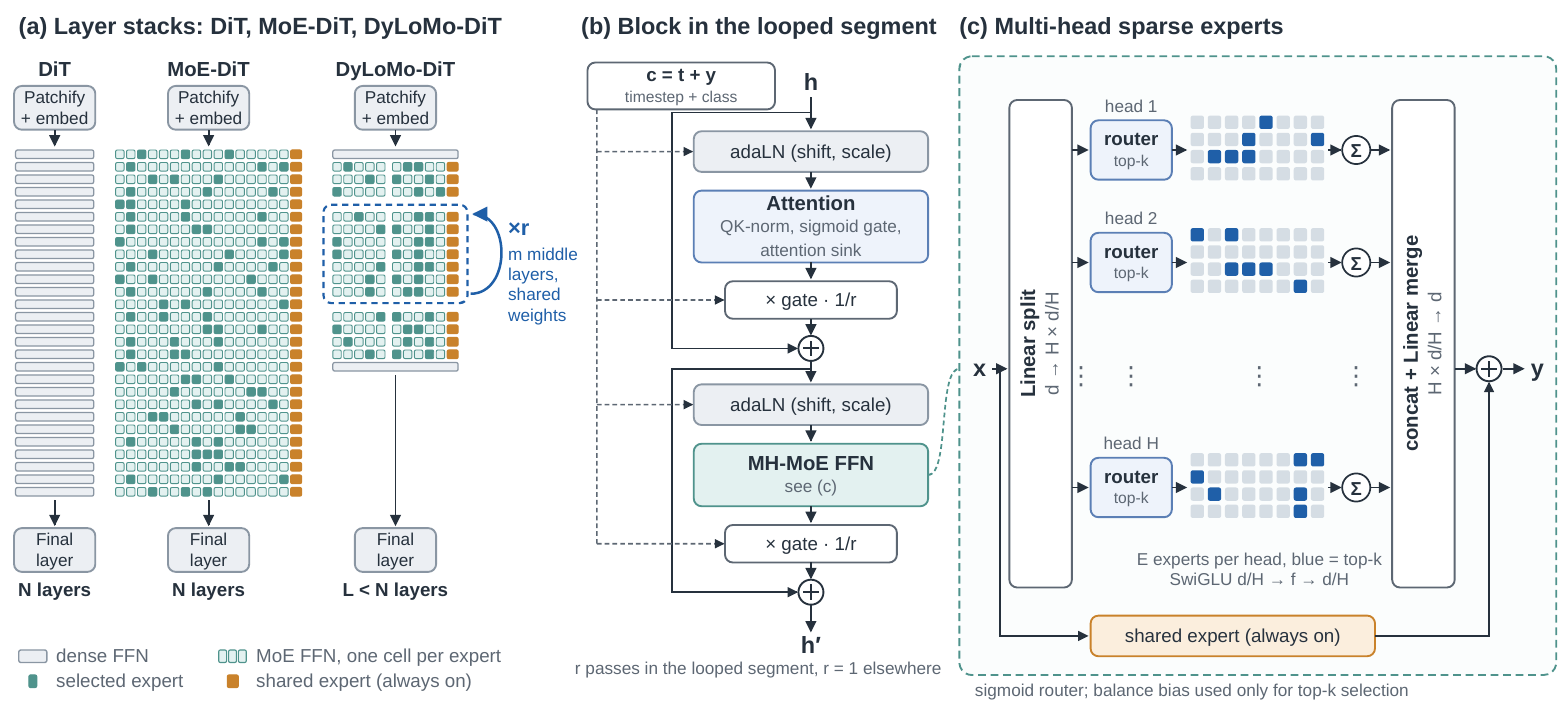}
  \caption{Architecture of DyLoMo-DiT. (a) DiT, an MoE-DiT, and DyLoMo-DiT (column width sketches the stored parameters): the MoE-DiT keeps the $N$ layers of DiT and adds experts to each; DyLoMo-DiT has fewer physical layers ($L<N$), with $m$ middle layers sharing weights over $r$ passes. (b) A looped layer: adaLN-Zero modulates the input and gates the output, scaled by $1/r$. (c) Multi-head sparse experts: $H$ heads, each router selecting $k$ of its $E$ experts, plus a shared expert (values: \cref{fig:arch_detail}).}
  \label{fig:arch}
\end{figure*}

DiT ties parameter count, forward compute, and depth (the number of layers each token traverses) to its width and layer count, using the same configuration for every token and noise level. DyLoMo-DiT makes these choices separately configurable (\cref{fig:arch}): width $d$; $L$ physical layers, including a middle segment of $m$ weight-shared layers repeated according to the noise-dependent schedule $r(t)$; and $H$ routing heads per feed-forward network, each with $E$ experts and $k$ selected per token. Weight sharing adds effective depth without duplicating parameters, freeing budget for experts. DiT, MoE-DiT, and looped DiT correspond to one expert and one pass, one pass, and one expert, respectively (\Cref{tab:freedoms}).

Otherwise, DyLoMo-DiT follows DiT~\cite{peebles2023dit}: patchified VAE latents are conditioned through adaLN-Zero using timestep and class embeddings, and the output layer predicts noise and variance. Training uses the DiT objective without auxiliary losses, loop-specific objectives, or pass embeddings. Further implementation details are in \cref{sec:app_h}.

\subsection{Parameters per Token: Multi-Head Sparse Experts}
\label{sec:mhmoe}

Except in the first and last layers, every feed-forward network is a multi-head sparse mixture of experts: a $d\times d$ linear layer transforms the sublayer input $x$ and splits it into $H$ heads of $d/H$ dimensions, each with its own router and $E$ experts. Each router gives its experts sigmoid scores, selects the top $k$ after adding a load-balancing bias, and weights them by the scores $L_1$-normalized over the $k$ and multiplied by a fixed constant $\alpha$ (\cref{sec:app_h}). The head outputs are concatenated, passed through a linear layer, and added to the output of a shared expert that is always on:
\begin{equation}
  \begin{aligned}
    y &= W_{\mathrm{merge}}\Big[\textstyle\sum_{e\in\mathcal{T}^{(j)}} w^{(j)}_e\, E^{(j)}_e\big(u^{(j)}\big)\Big]_{j=1}^{H} + E_{\mathrm{shared}}(x),\\
    u &= W_{\mathrm{split}}\,x.
  \end{aligned}
  \label{eq:mhmoe}
\end{equation}
Each expert is a gated feed-forward network $d/H\to f\to d/H$ with a gpt-oss-style~\cite{openai2025gptoss} clamped SwiGLU~\cite{shazeer2020glu} activation; at the XL tier, $H=4$, $E=16$, $k=6$ and $f=640$, so each token activates 24 routed experts per layer (other tiers and single-head models: \cref{sec:app_h}). The structure draws on Multi-Head Mixture-of-Experts~\cite{wu2024mhmoe} and the public code and model of MAGI-2~\cite{magi2}, the shared expert follows DeepSeekMoE~\cite{dai2024deepseekmoe}, and the balancing follows DeepSeek-V3~\cite{wang2024auxfree,deepseek2024v3}. No auxiliary loss is used: after each optimizer step the expert biases, which take part only in selection, are updated from the routing counts.

\subsection{Depth from a Weight-Shared Loop}
\label{sec:loop}

The $L$ physical layers form a front, a middle and a back segment, the middle one of $m$ consecutive layers: a forward pass executes the front segment, the middle segment $r$ times, and the back segment, with the $r$ passes sharing one set of parameters (attention, experts, routers and adaLN modulation), so depth grows with $r$ while the parameter count does not. Each sublayer of the middle segment writes to the residual with a factor of $1/r$:
\begin{equation}
  h \leftarrow h + \tfrac{1}{r}\, g(c) \odot F\big(\mathrm{adaLN}(h, c)\big),
  \label{eq:loop}
\end{equation}
where $F$ is the attention or feed-forward sublayer and $g(c)$ the adaLN-Zero gate; layers outside the segment have $r=1$. Routing is recomputed in every pass, so a token may select different experts in different passes. The recipe follows SMELT~\cite{wang2026smelt}, whose ablation favors looping about half of the layers, in the middle, over the full stack, and the factor $1/r$ follows Loop-$\mu$P~\cite{wang2026loopmup}, keeping the updates of the shared weights from accumulating across visits; under the ImageNet protocol the middle segment is also the best among front, back, middle and all layers (\cref{sec:app_g}).

\subsection{Depth by Noise Level: Allocating the Passes}
\label{sec:alloc}

The number of passes is a function of the noise level: the recommended schedule runs $r=3$ for $t<500$ and $r=1$ for $t\ge500$, two passes on average over uniformly spaced sampling steps, so the mean compute equals that of two passes at every timestep while a token passes through 30 layers at low noise and 16 at high noise (XL tier). The schedule is part of the model: the timestep of each training sample decides its number of passes, sampling uses the same schedule, and a model trained under one schedule must be run under that schedule; when the number of passes is instead drawn from $\{1,2,3\}$ at every training step, one set of weights can be run under several schedules, so depth can be chosen at inference. \Cref{sec:allocation} compares schedules; the return on extra passes (\cref{sec:loop_role}) is also observed on an in-house video generation model (\cref{sec:app_g}).

\subsection{Configuring a Model for a Budget}
\label{sec:config}

Because the three resources share one budget, a member is configured as a whole: the width and the number of layer visits set the forward compute, matched within 1\%; the loop decides how many of these visits are distinct layers (a shared middle segment of a third to a half of the layers leaves fewer physical layers than a loop-free model has at the same compute); and the number of experts fills the parameter budget, matched within about 2\%, so the parameters that the loop saves go to the experts. At the budget of DiT-XL/2 (28 layers, 0.68B parameters, 118.6 GMAC), DyLoMo-DiT-XL has 16 physical layers and 23 layer visits on average; removing the loop and spending the saved parameters on independent layers gives the single-pass member at the same budget (23 layers), which \cref{sec:vs_dit,sec:design_ablation} use to isolate the loop (all configurations: \cref{sec:app_h}).

%% file: sec/04_setup.tex
\section{Experimental Setup}
\label{sec:setup}
To ensure a fair comparison with DiT and DiT-MoE, we train all class-conditional models on ImageNet~\cite{deng2009imagenet} using the same training setup. We also train a single text-to-image model, without ablations, to assess whether DyLoMo-DiT extends to text-conditioned generation. This model scales up the architecture in \cref{sec:method}: it has width 1024, 32 physical layers with the middle 16 looped, 4 routing heads with 128 experts each, 15.48B total parameters, and 1.05B active parameters per image token. It differs from the class-conditional models in three respects: text and image tokens form a single sequence, with conditioning written into each token's input rather than injected through adaLN; it uses a 4-stream residual, which trains normally without adaLN and improves performance (\cref{sec:app_i}); and it is trained with the rectified-flow objective~\cite{liu2022rectified}.

\textbf{Training details.}
All ImageNet models follow a common pipeline based on the DiT recipe: VAE latents~\cite{rombach2022ldm}, the DDPM objective~\cite{ho2020ddpm,nichol2021improved}, a global batch size of 256, 400k training steps, AdamW~\cite{loshchilov2019adamw} with a constant learning rate of $10^{-4}$, and EMA with decay 0.9999 (\cref{sec:app_h}). We train the DiT baselines from scratch using the official model definitions; their unguided FID agrees with Table~4 of the DiT paper (\cref{sec:app_h}). We implement DiT-MoE~\cite{fei2024ditmoe} based on its public code (\cref{sec:vs_ditmoe}, \cref{sec:app_j}).

\textbf{ImageNet-256.}\label{sec:in256}
We evaluate class-conditional generation on ImageNet at $256\times256$, following the DiT protocol~\cite{peebles2023dit}. We use EMA weights and sample with 250-step DDPM and classifier-free guidance~\cite{ho2022cfg} at scale 1.5. Every 100k training steps, we generate 50k images per checkpoint and evaluate FID, sFID~\cite{nash2021sfid}, IS~\cite{salimans2016is}, Precision, and Recall~\cite{kynkaanniemi2019pr} using the ADM evaluator~\cite{dhariwal2021adm}.

\textbf{Compute matching.}
Because both model capacity and inference cost matter, we match each DyLoMo-DiT model to its baseline in total parameter count and forward compute, unless stated otherwise. We measure forward compute as the number of multiply-accumulates per image in one forward pass. GMAC denotes $10^9$ multiply-accumulates, counted in the same way as the Gflops reported in the DiT paper. For models with noise-dependent loop schedules, we average compute over the 250 uniformly spaced sampling steps. Parameter counts include the adaLN modulation layers, which account for about one third of DiT's parameters while adding negligible compute; these layers are shared within the looped segment. Excluding adaLN, DyLoMo-DiT has 1.2--1.3$\times$ the parameters of the DiT in the same tier (\cref{sec:app_b}). All reported results are means over seeds, and differences and percentages are computed from unrounded values.

\textbf{Text-to-image evaluation.}\label{sec:t2i_setup}
We report GenEval~\cite{ghosh2023geneval}; DPG-Bench~\cite{hu2024ella} and COCO FID~\cite{lin2014coco} are reported in \cref{sec:app_a}. We rerun and rescore public models using the same evaluation pipeline at their native resolutions. Differences smaller than 0.03 are treated as not meaningful (\cref{sec:app_a}).

%% file: sec/05_results.tex
\section{Results}
\label{sec:results}
The experiments have four stages. We first compare DyLoMo-DiT with DiT and DiT-MoE on ImageNet-256 at matched total parameters and forward compute (\cref{sec:vs_dit,sec:vs_ditmoe}), and then against scaled-up baselines (\cref{sec:scale_baselines}). Next, we ablate the design choices and analyze the loop's parameter--compute trade-off, the noise levels that benefit from extra passes, and the learned routing (\cref{sec:design_ablation,sec:loop_role,sec:allocation,sec:mechanism}). Finally, we test a text-to-image model at a larger scale (\cref{sec:t2i}). 

\subsection{Comparison with DiT at Equal Parameters and Compute}
\label{sec:vs_dit}

\begin{table*}[t]
\caption{ImageNet-256, 400k steps, EMA weights, guidance scale 1.5, 50k images. Each tier has three rows at the same total parameters and forward compute: the DiT of the tier, the same recipe without the loop, and DyLoMo-DiT (three passes for $t<500$, one for $t\ge500$). ``vs.\ DiT'': FID change relative to the DiT of the tier; best per column within a tier in bold. F1: harmonic mean of precision and recall.}
\label{tab:main}
\centering\footnotesize\setlength{\tabcolsep}{4pt}
\begin{tabular}{lrrrrrrrrrr}
\toprule
Model & \hd{Total\\params (M)} & \hd{Active\\params (M)} & GMAC & FID $\downarrow$ & vs.\ DiT & sFID $\downarrow$ & IS $\uparrow$ & Prec.\ $\uparrow$ & Rec.\ $\uparrow$ & F1 $\uparrow$ \\
\midrule
\tierrow{11}{B tier}
DiT-B/2 & 130.5 & 130.5 & 23.0 & 21.19 & --- & 6.01 & 68.9 & 0.643 & 0.553 & 0.595 \\
Same recipe, no loop (18 layers) & 130.2 & 114.8 & 23.1 & 19.45 & $-$8\% & 5.63 & \textbf{71.0} & 0.669 & \textbf{0.555} & 0.607 \\
DyLoMo-DiT-B (12 layers, 6 looped) & 128.5 & 75.8 & 23.1 & \textbf{18.82} & $-$11\% & \textbf{5.56} & 70.9 & \textbf{0.684} & 0.551 & \textbf{0.610} \\
\tierrow{11}{L tier}
DiT-L/2 & 458.1 & 458.1 & 80.7 & 7.42 & --- & 4.82 & 136.6 & 0.777 & 0.527 & 0.628 \\
Same recipe, no loop (18 layers) & 456.2 & 423.2 & 80.9 & 6.97 & $-$6\% & \textbf{4.48} & 136.8 & 0.786 & \textbf{0.531} & 0.634 \\
DyLoMo-DiT-L (12 layers, 6 looped) & 451.3 & 278.3 & 80.1 & \textbf{6.41} & $-$14\% & 4.50 & \textbf{140.5} & \textbf{0.799} & 0.526 & \textbf{0.635} \\
\tierrow{11}{XL tier}
DiT-XL/2 & 675.1 & 675.1 & 118.6 & 5.57 & --- & 4.67 & 158.2 & 0.801 & \textbf{0.526} & 0.635 \\
Same recipe, no loop (23 layers) & 678.4 & 595.8 & 117.7 & 5.02 & $-$10\% & 4.34 & 163.8 & 0.814 & 0.523 & 0.637 \\
DyLoMo-DiT-XL (16 layers, 7 looped) & 686.6 & 411.3 & 117.8 & \textbf{4.22} & $-$24\% & \textbf{4.30} & \textbf{174.4} & \textbf{0.824} & 0.524 & \textbf{0.640} \\
\bottomrule
\end{tabular}
\end{table*}

\begin{figure}[t]
\centering
\includegraphics[width=\linewidth]{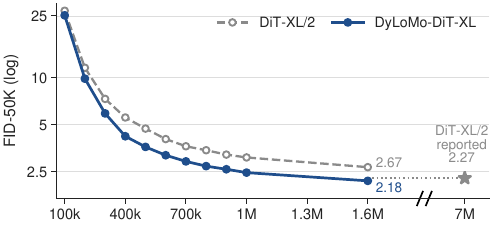}
\caption{Training to 1.6M steps, XL tier (FID-50K on a log scale, guidance 1.5). From 400k on, the FID ratio stays within 0.76--0.82; points after 400k are single runs (\cref{sec:app_b}).}
\label{fig:1m}
\end{figure}

\begin{figure}[t]
\centering
\includegraphics[width=0.9\linewidth]{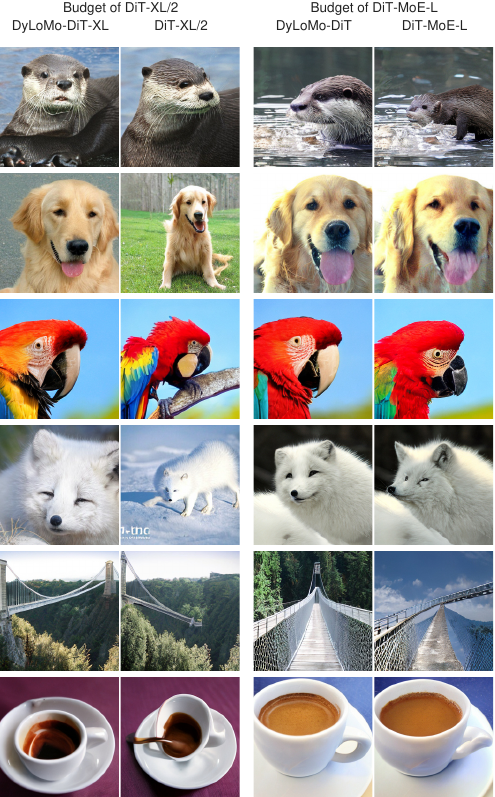}
\caption{Class-conditional samples (400k steps, guidance scale 4.0); rows: otter, golden retriever, macaw, arctic fox, suspension bridge, espresso. Left: at the compute of DiT-XL/2; right: at the budget of DiT-MoE-L. Each pair shares the class label and noise; more in \cref{sec:app_d}.}
\label{fig:compare}
\end{figure}

At matched parameter and compute budgets, DyLoMo-DiT reduces FID relative to DiT by 11\%, 14\%, and 24\% in the B, L, and XL tiers, respectively (\Cref{tab:main}). Its advantage grows during training: the FID ratio of DyLoMo-DiT to DiT declines from 100k to 400k steps in every tier (B: $1.05\rightarrow0.89$; L: $0.98\rightarrow0.86$; XL: $0.93\rightarrow0.76$), as shown in \Cref{fig:teaser}. The no-loop variant accounts for part of the improvement; adding the shared loop with noise-dependent pass allocation further reduces FID by 3\%, 8\%, and 16\% in the B, L, and XL tiers, with larger gains at larger scales. The advantage persists with longer training: at 1M steps, DyLoMo-DiT-XL reaches an FID of 2.46 versus 3.08 for DiT-XL/2 (\cref{fig:1m}); at 1.6M steps, it reaches 2.18, below the 2.27 reported for DiT-XL/2 after 7M steps~\cite{peebles2023dit}. \Cref{fig:compare} compares samples generated from the same class label and noise.

\subsection{Comparison with DiT-MoE at Its Own Budget}
\label{sec:vs_ditmoe}
At matched total parameter counts and forward compute, DyLoMo-DiT reduces FID by 36\%, 56\%, and 56\% relative to DiT-MoE in the S, B, and L tiers, respectively (\Cref{tab:ditmoe}; \cref{fig:teaser}, bottom). We compare at each DiT-MoE tier's own budget because these models retain the width and depth of the corresponding DiT while adding experts to every layer, resulting in about six times as many parameters and 2.7 times the compute. For each tier, we configure DyLoMo-DiT to match DiT-MoE's total parameter count and compute: one routing head, 16 experts per layer with 2 selected, 16 layers with the middle 6 looped, and the same pass allocation as above (\cref{sec:config}). We train DiT-MoE under the same protocol, using 16 experts per layer with 4 selected. Thus, both models have the same number of experts per layer. These results reflect our DiT-MoE implementation under this training protocol, rather than the results reported in its paper (\cref{sec:app_j}).

We also evaluate the original DiT-MoE configuration, which uses 8 experts per layer, selects 2, and includes a shared expert. It obtains FIDs of 30.80, 12.91, and 7.50 in the S, B, and L tiers; DyLoMo-DiT reduces these by 36\%, 46\%, and 53\%, respectively (XL results: \cref{sec:scale_baselines}). We further compare against ProMoE~\cite{wei2025promoe}. At the ProMoE-L budget and after 500k training steps, matching the schedule in its paper, DyLoMo-DiT reaches an FID of 4.05, compared with 4.86 for our ProMoE-L rerun and 5.12 as reported. At 400k steps, our ProMoE-XL rerun reaches an FID of 4.78, compared with 4.22 for DyLoMo-DiT-XL, despite using $2.3\times$ as many parameters (\cref{sec:app_j}).

\begin{table*}[t]
\caption{Comparison at the DiT-MoE budget (ImageNet-256, 400k steps, EMA weights, guidance scale 1.5, 50k images). Each tier has two rows at the same total parameters and forward compute: the DiT-MoE of the tier and DyLoMo-DiT. ``vs.\ DiT-MoE'': FID change relative to the DiT-MoE of the tier; best per column within a tier in bold. F1: harmonic mean of precision and recall.}
\label{tab:ditmoe}
\centering\footnotesize\setlength{\tabcolsep}{4pt}
\begin{tabular}{lrrrrrrrrrr}
\toprule
Model & \hd{Total\\params (M)} & \hd{Active\\params (M)} & GMAC & FID $\downarrow$ & vs.\ DiT-MoE & sFID $\downarrow$ & IS $\uparrow$ & Prec.\ $\uparrow$ & Rec.\ $\uparrow$ & F1 $\uparrow$ \\
\midrule
\tierrow{11}{S tier}
DiT-MoE-S & 199.3 & 71.9 & 16.0 & 30.83 & --- & 8.02 & 48.5 & 0.573 & \textbf{0.558} & 0.565 \\
DyLoMo-DiT (budget of DiT-MoE-S) & 199.6 & 56.9 & 16.1 & \textbf{19.79} & $-$36\% & \textbf{5.90} & \textbf{68.5} & \textbf{0.666} & 0.544 & \textbf{0.599} \\
\tierrow{11}{B tier}
DiT-MoE-B & 795.9 & 286.3 & 62.9 & 16.06 & --- & 5.90 & 86.9 & 0.683 & \textbf{0.540} & 0.603 \\
DyLoMo-DiT (budget of DiT-MoE-B) & 796.2 & 229.0 & 62.9 & \textbf{7.01} & $-$56\% & \textbf{4.57} & \textbf{135.0} & \textbf{0.784} & 0.531 & \textbf{0.633} \\
\tierrow{11}{L tier}
DiT-MoE-L & 2824.0 & 1012.0 & 222.5 & 7.92 & --- & 4.94 & 130.4 & 0.765 & \textbf{0.540} & 0.633 \\
DyLoMo-DiT (budget of DiT-MoE-L) & 2821.8 & 712.0 & 222.5 & \textbf{3.50} & $-$56\% & \textbf{4.22} & \textbf{198.0} & \textbf{0.836} & 0.526 & \textbf{0.646} \\
\bottomrule
\end{tabular}
\end{table*}

\subsection{Scaling Up the Baselines}
\label{sec:scale_baselines}

To test whether the gains can be recovered simply by increasing baseline size, we scale up DiT and DiT-MoE (\cref{sec:app_b}). DiT-XL/2 scaled to $1.6\times$, $2.2\times$, and $3.0\times$ gives FIDs of 5.22, 4.57, and 4.75. Even at three times the parameters and compute, DiT remains 13\% worse than DyLoMo-DiT-XL at its original scale (4.22 FID). The original XL DiT-MoE, with $6.2\times$ the parameters and $2.8\times$ the compute, reaches 6.87 FID, 63\% worse than DyLoMo-DiT-XL. DiT-MoE scaled to $2.2\times$ in both parameters and compute reaches 6.65 FID, 58\% worse.

\subsection{Ablating the Design at the DiT-MoE-B Budget}
\label{sec:design_ablation}

\begin{table}[t]
\caption{From DiT-MoE-B to DyLoMo-DiT at one budget (setting of \Cref{tab:ditmoe}; every row 780--797M parameters, 62.9--63.1 GMAC). ``Experts'': heads $\times$ experts per head (selected per head); ``Step'': FID change from the row above.}
\label{tab:path}
\centering\footnotesize\setlength{\tabcolsep}{2.2pt}
\begin{tabular}{@{}lrlrr@{}}
\toprule
Model & Layers & Experts & FID $\downarrow$ & Step \\
\midrule
DiT-MoE-B & 12 & 16 (4) & 16.06 & --- \\
DyLoMo-DiT, no loop & 22 & 1$\times$10 (2) & 7.93 & $-$51\% \\
\ + loop, two passes at every $t$ & 16 & 1$\times$16 (2) & 7.94 & $\pm$0\% \\
\ + passes allocated by noise level & 16 & 1$\times$16 (2) & 7.01 & $-$12\% \\
\ + three routing heads & 16 & 3$\times$16 (2) & 5.61 & $-$20\% \\
\ + finer experts & 16 & 3$\times$46 (6) & \textbf{5.28} & $-$6\% \\
\bottomrule
\end{tabular}
\end{table}

To isolate each design choice, \Cref{tab:path} adds them incrementally at the DiT-MoE-B budget. The single-pass model reduces FID from 16.06 to 7.93. A uniform two-pass loop leaves FID essentially unchanged (7.94), showing that recurrence alone does not improve this sparse-expert model at the matched budget. Noise-dependent pass allocation then lowers FID by 12\%, with three routing heads and finer-grained experts reducing it further to 5.28. The final model's FID is 67\% lower than DiT-MoE-B; other layouts (layer count, width, and routing heads) are in \cref{sec:app_b}.

\subsection{The Loop's Parameter--Compute Trade-off}
\label{sec:loop_role}

\begin{figure*}[t]
\centering
\includegraphics[width=0.94\textwidth]{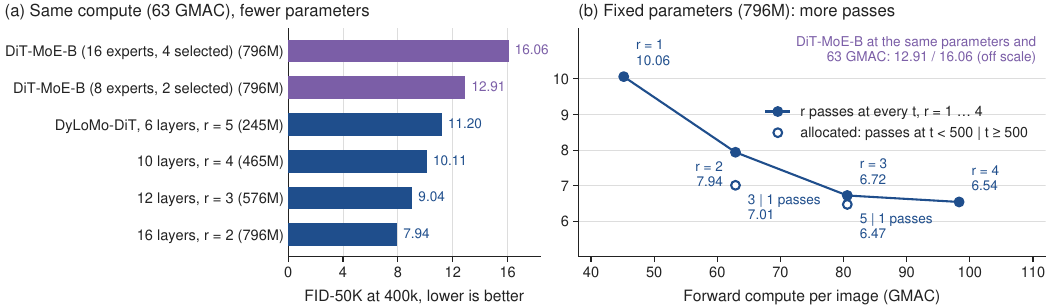}
\caption{The two roles of the loop (B tier, single routing head). (a) Fewer parameters at fixed compute: at $0.31\times$ the parameters, the model is still ahead of both DiT-MoE-B configurations. (b) More compute at fixed parameters: one to four passes at every timestep (filled) and passes allocated by noise level (open; 3 or 5 passes for $t<500$, 1 for $t\ge500$), each close to or level with one more pass.}
\label{fig:loop}
\end{figure*}

We next examine the loop's two resource trade-offs (\cref{fig:loop}). At fixed compute (63 GMAC), reducing parameters to $0.72\times$, $0.58\times$, and $0.31\times$ of the two-pass model raises FID by 14\%, 27\%, and 41\%, respectively. Even with less than one third of its parameters, the smallest model achieves 13\% and 30\% lower FID than the two DiT-MoE configurations at the same compute. Conversely, at fixed parameters (796M), increasing passes from one to four reduces FID by 21\%, 15\%, and 2.7\% across successive increases. Allocating passes by noise level is worth roughly one extra pass: at two-pass compute, it approaches three passes throughout (7.01 vs.\ 6.72 FID), and at three-pass compute, it matches four passes throughout (6.47 vs.\ 6.54).

\subsection{Allocating Passes by Noise Level}
\label{sec:allocation}

\begin{figure}[t]
\centering
\includegraphics[width=0.93\linewidth]{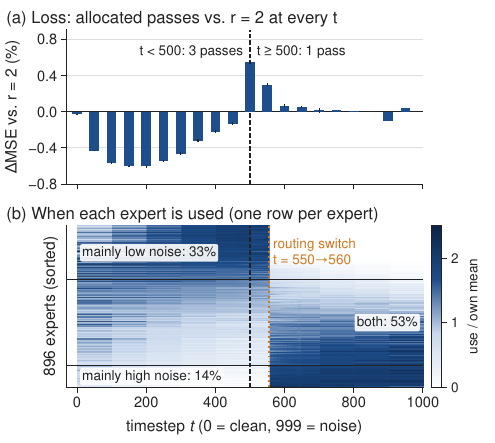}
\caption{Passes and routing by noise level (DyLoMo-DiT-XL). (a) Per-timestep loss of allocated passes relative to two passes at every $t$: the extra pass pays only at low noise. (b) When each expert is used (one row per expert; color relative to its own mean): a low-noise and a high-noise group, switching at $t=550\rightarrow560$ (dotted), away from the allocation boundary $t=500$ (dashed).}
\label{fig:noise}
\end{figure}

We next examine where additional loop computation is useful along the denoising trajectory. Compared with two passes at every timestep, allocated-pass DyLoMo-DiT-XL has lower noise-prediction error at every $t<500$ (by up to 0.6\% around $t=150$--200), slightly higher error just above 500, and similar error beyond 600 (\cref{fig:noise}a). At the DiT budget, noise-dependent allocation lowers FID by 4.2\%, 4.7\%, and 3.4\% in the B, L, and XL tiers, using all available seeds; at the DiT-MoE budget, it improves the single-head B model by 12\% (\cref{sec:app_b}). Reversing the allocation worsens the B-tier FID by 33\% (to 26.14), while a three-band schedule performs similarly to the two-band schedule (\cref{sec:app_g}).

The pass schedule can also be chosen after training (\cref{sec:app_g}). A DyLoMo-DiT-XL trained with pass counts sampled from $\{1,2,3\}$ can run under different schedules; any schedule with at least two passes at low noise reaches nearly the same FID (within 3\%). When run with two passes, this model reaches 4.68 FID, 7\% worse than the model trained with two passes throughout (4.36), reflecting the cost of choosing depth at inference.

\subsection{Routing Specialization by Noise Level}
\label{sec:mechanism}
We analyze routing in the DyLoMo-DiT-XL model in \Cref{tab:main} (protocol and two-pass control: \cref{sec:app_f}).

\textbf{The second pass reuses most of the experts.}
On the second pass, each token reuses an average of 4.61 of the 6 experts per head selected on the first pass, compared with 2.57 for the same position in another image and 2.25 under random selection. The two passes also produce similarly directed residual writes, with cosine similarities of 0.68 for attention and 0.62 for feed-forward networks. These aligned writes motivate the $1/r$ scaling across passes in \cref{sec:loop}.

\textbf{Routing specializes by noise level and can switch between adjacent steps.}
Across adjacent timesteps within either $t\le550$ or $t\ge650$, the expert distribution changes little (JS divergence of about 0.005 nats). Between $t=550$ and $t=650$, it shifts sharply, reaching 0.30, and every layer switches (\cref{sec:app_f}). On a finer grid with step size 10, the switch occurs between $t=550$ and $t=560$ (JS divergence 0.287); all other adjacent pairs on this grid have divergence at most 0.0005. Experts specialize across the two ranges: 47\% receive more than 80\% of their calls in one range (\cref{fig:noise}b), compared with 36\% in the two-pass control (\cref{sec:app_f}). The number of active experts also falls from 15.0 at $t=50$ to 9.2 at $t=950$.

The router receives no explicit timestep input; timestep information reaches it only through adaLN-modulated features, and the learned timestep embedding itself changes abruptly where the routing switches, unlike that of DiT (\cref{sec:app_f}). The routing switch does not align with the pass-allocation boundary at $t=500$ (the JS divergence from $t=499$ to $t=500$ is 0.002), and a model without the loop also switches within one step (\cref{sec:app_f}). These findings suggest that the router's noise-level specialization is learned from its input features, rather than imposed by the pass schedule or the loop.

\begin{table}[t]
\caption{GenEval recomputed with our pipeline at each model's native resolution. Parameters: denoiser only (DyLoMo-DiT: total / active per image token). GMAC: one forward pass of the denoiser at the native resolution. Checkpoints and sampling settings: \cref{sec:app_a}.}
\label{tab:geneval}
\centering\footnotesize\setlength{\tabcolsep}{3pt}
\begin{tabular}{@{}llcrc@{}}
\toprule
Model & \hd{Parameters\\(B)} & \hd{Native\\resolution} & GMAC & GenEval $\uparrow$ \\
\midrule
DyLoMo-DiT & 15.48 / 1.05 active & $1024^2$ & 8,333 & \textbf{0.714} \\
FLUX.1-schnell & 12 & $1024^2$ & 34,733 & 0.673 \\
FLUX.1-dev & 12 & $1024^2$ & 37,192 & 0.671 \\
Sana-1.6B~\cite{xie2024sana} & 1.6 & $1024^2$ & 1,483 & 0.632 \\
SDXL~\cite{podell2023sdxl} & 2.6 & $1024^2$ & 3,381 & 0.551 \\
PixArt-$\alpha$~\cite{chen2023pixartalpha} & 0.6 & $512^2$ & 618 & 0.483 \\
\bottomrule
\end{tabular}
\end{table}

\subsection{Text-to-Image Generation}
\label{sec:t2i}
Scored with the same pipeline at their native resolutions, DyLoMo-DiT achieves a GenEval score of 0.714 at $1024^2$, surpassing FLUX.1-schnell (0.673), FLUX.1-dev (0.671), and the other models in the comparison (\Cref{tab:geneval}). It has 15.48B total parameters, of which 1.05B are active per image token. DyLoMo-DiT uses 24\% and 22\% as much forward-pass compute as FLUX.1-schnell and FLUX.1-dev, respectively. Sample generations are shown in \Cref{fig:head}.

%% file: sec/06_conclusion.tex
\section{Conclusion}
\label{sec:conclusion}

DyLoMo-DiT improves matched-budget efficiency by combining sparse experts with weight-shared recurrence, with further gains from dynamically assigning more passes to low-noise steps. Routers also learn noise-level specialization without explicit timestep inputs. On ImageNet, DyLoMo-DiT reduces FID by 11--24\% versus DiT and 36--56\% versus DiT-MoE at matched budgets; at equal compute, it outperforms DiT-MoE-B with less than one third of its parameters, and a DiT with three times the parameters and compute remains 13\% behind; at 1.6M steps, it surpasses the FID reported for DiT-XL/2 after 7M steps. Scaled up to 15B parameters, it becomes a production-scale text-to-image model that scores above FLUX.1 on GenEval with less than one tenth of its active parameters, and it transfers to video (\cref{sec:app_c}).

%% file: supp/A_t2i_details.tex
\section{Sampling and Evaluation Details of the Text-to-Image Experiments}
\label{sec:app_a}

This appendix gives the details behind \cref{sec:t2i_setup} (text-to-image setup) and \cref{sec:t2i}, and lists the prompts of the head figure (\cref{fig:head}). The experiment asks whether the same architecture carries over from class-conditional generation on ImageNet to text-to-image; only one model is trained, with no ablation. Images are encoded with the Wan2.2 VAE~\cite{wan2025wan22} and text with a frozen Qwen2.5-VL-7B~\cite{bai2025qwen25vl}; sampling uses 50 Euler steps with APG~\cite{sadat2024apg} at guidance scale 5.

\paragraph{Model.}
The first and last layers use dense feed-forward networks, and the 30 layers in between use sparse-expert feed-forward networks; each token selects 6 experts per head, and there is one always-on shared expert (width 1280). The looped segment is the middle 16 layers, with 8 layers before it and 8 after it, so one forward pass executes $8+16\times2+8=48$ layers. Active parameters are counted as follows: the layers of the looped segment are counted once; the routers, which score all experts, are counted in full; the image input and output adapter layers are counted once. The timestep is written into the first 64 channels of the input as a 64-dimensional sinusoidal encoding.

\paragraph{Encoders.}
The VAE has 48 channels and downsamples spatially by $16\times$; there is no further patching, so one latent position is one token. Text features are taken from the output of the third-to-last layer of the text encoder, before its final normalization layer, with at most 1,000 tokens.

\begin{figure}[t]
\centering
\includegraphics[width=\linewidth]{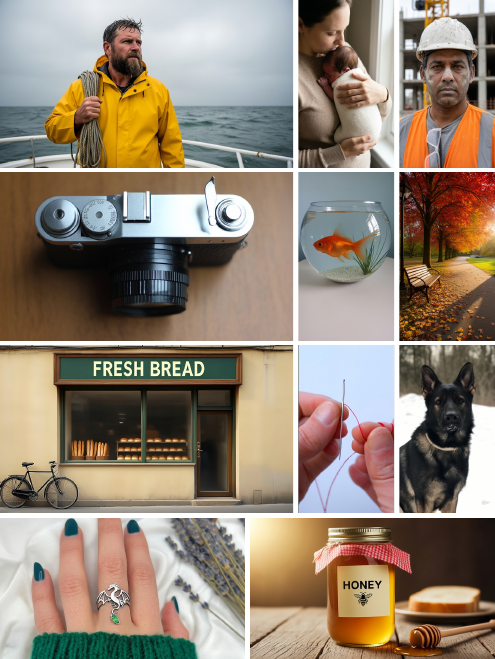}
\caption{Text-to-image samples from DyLoMo-DiT (landscape $784\times448$, portrait $448\times784$; APG, guidance scale 5, 50 Euler steps). Prompts are listed in \Cref{tab:a2}.}
\label{fig:t2i}
\end{figure}

\paragraph{Checkpoints.}
\Cref{tab:geneval} uses the checkpoint evaluated at $1024\times1024$ (the checkpoint of the head figure and of the gallery below). Earlier checkpoints score 0.705 at $448\times256$ and 0.700 at $784\times448$; the samples in \cref{fig:t2i} come from the latter.

\paragraph{Sampling.}
Euler method with 50 steps; $u$ is taken at equally spaced points from 1 to 0 and then passed through the shift transform $\sigma=s\cdot u/(1+(s-1)\cdot u)$ with $s=1.6$. APG~\cite{sadat2024apg} acts on the predicted clean image, with guidance scale 5, momentum $-0.5$, parallel-component weight $\eta=0.15$ and no rescaling; the unconditional branch uses the empty text, and no negative prompt is used. All three metrics use this one set of settings.

\paragraph{Metrics.}
GenEval~\cite{ghosh2023geneval} has 553 prompts with 4 images each, 2,212 images in total; we use the official implementation, with Mask2Former (Swin-S)~\cite{cheng2022mask2former} as the detector and colors judged by CLIP~\cite{radford2021clip}. DPG-Bench has 1,065 prompts with 4 images each, scored with the script released by ELLA~\cite{hu2024ella} and mPLUG-large~\cite{li2022mplug}; the official script reads a $2\times2$ grid and splits it, whereas we feed the 4 single images directly; we also fixed one issue in the upstream parsing script, which skipped the first question of every prompt. COCO FID uses the first 10,000 captions of the commonly used 30,000-caption MS-COCO~\cite{lin2014coco} val2014 list, one image per caption, with the original images of these captions as the reference set; generated and reference images are both resized directly to $256\times256$ (no cropping, so the aspect ratio changes), and FID is computed with clean-fid~\cite{parmar2022cleanfid}. DyLoMo-DiT scores 83.6 on DPG-Bench at $448\times256$ and 85.2 at $784\times448$, and its COCO FID is 20.5 at $448\times256$.

\begin{table}[t]
\caption{DPG-Bench and COCO FID-10K at $1024\times1024$ with the checkpoints and sampling settings of \Cref{tab:geneval}. DPG-Bench: 1,065 prompts, 4 images each; intervals are 95\% percentile intervals over 10,000 bootstrap resamples of the prompts, and differences are paired by prompt. COCO FID-10K: the protocol above, one generation run per model.}
\label{tab:a_dpg}
\centering\footnotesize\setlength{\tabcolsep}{2.3pt}
\begin{tabular}{@{}lccccc@{}}
\toprule
 & \multicolumn{2}{c}{DPG-Bench $\uparrow$} & \multicolumn{2}{c}{DPG vs.\ DyLoMo-DiT} & COCO \\
\cmidrule(lr){2-3}\cmidrule(lr){4-5}
Model & score & interval & value & interval & FID-10K $\downarrow$ \\
\midrule
DyLoMo-DiT & 85.0 & $[83.9, 86.0]$ & --- & --- & 17.9 \\
FLUX.1-schnell & 84.8 & $[83.7, 85.8]$ & $-0.2$ & $[-1.1, 0.7]$ & 22.4 \\
Sana-1.6B & 83.9 & $[82.9, 84.9]$ & $-1.0$ & $[-1.9, -0.2]$ & 27.0 \\
FLUX.1-dev & 83.6 & $[82.6, 84.7]$ & $-1.3$ & $[-2.2, -0.4]$ & 28.6 \\
\bottomrule
\end{tabular}
\end{table}

\paragraph{DPG-Bench and COCO FID against the public models.}
With the DPG-Bench procedure above, we also score DyLoMo-DiT and the three public models with the highest GenEval in \Cref{tab:geneval}, each at $1024\times1024$ with the checkpoint and sampling settings of that table (\Cref{tab:a_dpg}). For DyLoMo-DiT, this evaluation uses the checkpoint with its routed-expert parameters stored in bf16. Every model generates 4 images for each of the 1,065 prompts; every image has its own seed, assigned by the same rule for all models. DyLoMo-DiT scores 85.0, FLUX.1-schnell 84.8, Sana-1.6B 83.9 and FLUX.1-dev 83.6. The differences to Sana-1.6B and FLUX.1-dev exclude zero; the difference to FLUX.1-schnell does not, so we do not interpret it. With the COCO protocol above, at the same resolution and settings and one image per caption, DyLoMo-DiT reaches an FID-10K of 17.9, against 22.4 for FLUX.1-schnell, 27.0 for Sana-1.6B and 28.6 for FLUX.1-dev; each value comes from one generation run per model. Center-cropping both image sets before resizing changes the values (16.3, 21.1, 25.8 and 27.1) but not their order.

\paragraph{Magnitude of fluctuation.}
Only one model was trained on the text-to-image line, so the only available reference for fluctuation comes from consecutive checkpoints, whose GenEval scores differ by at most about 0.03. On this basis, we do not interpret GenEval differences within 0.03.

\paragraph{Comparison models.}
The public models are models with open weights and a published GenEval score, covering 0.6B to 12B denoiser parameters and including FLUX.1~\cite{flux2024} (12B), the strongest open dense model of a scale close to ours. Parameters in \Cref{tab:geneval} count the denoiser only (excluding the text encoder and VAE), as given in each model's paper or official release; for DyLoMo-DiT, total parameters / active parameters per image token. Prompts are not rewritten, and no negative prompt is used. The sampling settings of each model are: SDXL~\cite{podell2023sdxl}, 50 steps, guidance scale 9.0 (the value in the official GenEval example script); SD 1.5~\cite{rombach2022ldm}, 50 steps, 7.5; PixArt-$\alpha$~\cite{chen2023pixartalpha}, 20 steps, 4.5; Sana-1.6B~\cite{xie2024sana}, 20 steps, 4.5; FLUX.1-dev~\cite{flux2024}, 50 steps, 3.5; FLUX.1-schnell, 4 steps, no guidance. GenEval recomputed by our pipeline at each model's native resolution: SDXL 0.551 (published 0.55), SD 1.5 0.423 (published 0.43), PixArt-$\alpha$ 512 0.483 (published 0.48), FLUX.1-dev 0.671 (published 0.66 or 0.67), all within 0.011 of the published values; Sana-1.6B 0.632 (published 0.66) is 0.03 lower, and FLUX.1-schnell 0.673 (published 0.69 or 0.71) is 0.02 to 0.04 lower. Two further public models were evaluated but not listed in \Cref{tab:geneval}: SD 1.5 scores 0.423 at its native $512^2$ and PixArt-$\alpha$ (256) scores 0.491 at $256^2$. At $1024^2$, DyLoMo-DiT (0.714) is above the recomputed values of both FLUX.1 models. At $448\times256$, DyLoMo-DiT scores 0.705 and the public models 0.701 (FLUX.1-schnell), 0.614 (FLUX.1-dev), 0.412 (PixArt-$\alpha$), 0.133 (Sana-1.6B) and 0.021 (SDXL); the last two do not work properly at this resolution. The GMAC column of \Cref{tab:geneval} counts one forward pass of the denoiser at the native resolution, in the way of \cref{sec:setup} (text encoder and VAE excluded). Text tokens are counted as each sampling pipeline feeds them to the denoiser: 512 for FLUX.1-dev, 256 for FLUX.1-schnell, 300 for Sana-1.6B, 120 for PixArt-$\alpha$ and 77 for SDXL, all padded, and the unpadded prompt for DyLoMo-DiT (12.9 tokens on average over the GenEval prompts). There are 4,096 image tokens for DyLoMo-DiT and FLUX.1 and 1,024 for Sana-1.6B and PixArt-$\alpha$. Sources of the published values: SDXL and SD 1.5 from the official GenEval repository; PixArt-$\alpha$ ($512\times512$) and Sana-1.6B from the comparison table of the Sana paper; FLUX.1-dev and FLUX.1-schnell are 0.67 and 0.71 in the comparison table of the Sana paper and 0.66 and 0.69 in Table~1 of the rCM paper~\cite{zheng2025rcm}.

\input{supp/A_prompts}

\paragraph{First-page samples.}
The samples in the head figure of the main paper (\cref{fig:head}) come from the same model (architecture and parameter count unchanged), using the checkpoint evaluated in \Cref{tab:geneval}. The sampling settings are the same as above (APG with guidance scale 5, Euler with 50 steps), and the shift is either 1.6 or the resolution-dependent value given in the gallery paragraph below; 4 images are generated per prompt (two seeds $\times$ two shift values), and one is selected. The prompts of the ten samples are listed in \Cref{tab:a3}.
\input{supp/A_prompts_row}

%% file: supp/A_prompts.tex
\begin{table*}[t]
\caption{Prompts of the 11 text-to-image samples in \cref{fig:t2i} (position = row and column in that figure). All samples: APG with guidance scale 5, Euler 50 steps.}
\label{tab:a2}
\centering\footnotesize\setlength{\tabcolsep}{4pt}
\begin{tabular}{llp{13.2cm}}
\toprule
Position & Size & Prompt \\
\midrule
row1-1 & 784$\times$448 & A middle-aged bearded man stands on the deck of a small fishing boat, gripping a coiled rope in his left hand and looking toward the horizon with narrowed eyes. He is wearing a bright yellow rain jacket with the hood down, and his dark hair is wet and pushed back from his forehead. Drops of water cling to his beard and the shoulders of the jacket. The sea behind him is gray and choppy under a heavy overcast sky, and the white railing of the boat crosses the lower part of the frame. The photo is taken at chest height from a few steps away. \\
row1-2 & 448$\times$784 & A woman holds a sleeping newborn baby against her chest, shown from the waist up. The baby lies upright, with its head resting under her chin and its tiny fist closed around the collar of her beige sweater. Her left hand supports the baby's bottom, and her right hand lies flat across the baby's back, covering most of the white knitted blanket wrapped around it. She looks down at the baby with her eyes almost closed. Soft daylight from a window on the right falls across her face and the baby's fine dark hair. \\
row1-3 & 448$\times$784 & A close-up portrait of a middle-aged male construction worker, wearing a scratched white hard hat and an orange high-visibility vest over a gray t-shirt. His tanned face is streaked with pale dust, and he has dark stubble and tired brown eyes that look straight into the lens. A pair of clear safety glasses hangs from the collar of his shirt. The light is flat and overcast, without strong shadows. Behind him, heavily blurred, are the gray concrete frame of an unfinished building and the yellow arm of a crane. \\
row2-1 & 784$\times$448 & A vintage 35mm film camera lies on a wooden table, shown from above and slightly from the front. The top plate is brushed silver metal and holds a film advance lever on the right, a round shutter speed dial with small engraved numbers next to it, and a rewind crank on the left. The lens points toward the bottom of the frame and has a ribbed focus ring and a narrower aperture ring. The body is wrapped in black textured leather. No brand name is visible. Light from a nearby window reflects softly in the front glass of the lens. \\
row2-2 & 448$\times$784 & A single orange goldfish swims in a round glass fishbowl that stands on a white table. The fish is in the center of the bowl, seen from the side and facing left, with a long, flowing, semi-transparent tail and small fins. The bowl is filled with clear water to about four fifths of its height, and a layer of small white pebbles covers the bottom. One green water plant with thin leaves grows from the pebbles on the right. The curved glass slightly magnifies the fish and carries a bright window reflection. The wall behind is pale blue. \\
row2-3 & 448$\times$784 & A park path in autumn, seen along its length. On the left side of the path, a wooden bench with iron armrests stands under a black lamppost. Along the right side, a row of maple trees with red and orange leaves leans over the path and forms a canopy. In the foreground, the paved path is covered with fallen leaves in yellow, orange, and brown. In the background, the path curves gently to the right and a small stone bridge is visible in the distance. Low sunlight shines through the leaves from the right and casts long shadows across the path. \\
row3-1 & 784$\times$448 & A small bakery storefront is shown straight on from across a narrow street. Above the large display window hangs a long wooden sign painted dark green, with the words ``FRESH BREAD'' in large cream capital letters. The window shows shelves of round loaves and baguettes standing in baskets. To the right of the window is a glass door with a brass handle, and a black bicycle leans against the wall on the left. The wall is pale yellow plaster, and soft morning light falls evenly across the front of the shop. \\
row3-2 & 448$\times$784 & A close-up of two hands threading a sewing needle against a plain light blue background. The left hand holds a thin steel needle upright between the tips of the thumb and index finger, with the eye of the needle at the top. The right hand holds a length of red thread between thumb and index finger and guides its end toward the eye, with the thread just about to pass through. The other fingers of both hands are loosely curled. The nails are short and unpainted. Bright, even light shows the fine lines of the skin and the fibers of the thread. \\
row3-3 & 448$\times$784 & A black and tan German Shepherd dog is sitting in the snow. The dog has erect ears, dark eyes, and a black nose with some snow on its muzzle. It wears a light-colored collar. The background consists of blurred trees and snow-covered ground. \\
row4-1 & 784$\times$448 & A hand with dark teal nail polish is shown wearing a silver dragon ring on the middle finger. The ring features a stylized dragon with wings spread, and a small green gemstone at the tip of its tail. The hand is emerging from a green knitted sleeve. In the background, there are dried lavender sprigs on a white fabric surface. \\
row4-2 & 784$\times$448 & A glass jar of golden honey stands on a rustic wooden table. A cream paper label is glued to the front of the jar, with the word ``HONEY'' printed in bold black capital letters, and a small drawing of a bee below it. The jar has a metal lid covered with a square of red checkered cloth tied with string. A wooden honey dipper lies on the table to the right of the jar, with a small shining pool of honey under its tip. A slice of bread on a plate is blurred in the background. Warm sunlight from the left glows through the honey. \\
\bottomrule
\end{tabular}
\end{table*}

%% file: supp/A_prompts_row.tex
\begin{table*}[t]
\caption{Prompts of the 10 text-to-image samples in Fig.~1 (position = row--column, rows from top, columns from left). All samples: APG with guidance scale 5, Euler 50 steps.}
\label{tab:a3}
\centering\footnotesize\setlength{\tabcolsep}{4pt}
\begin{tabular}{llp{13.4cm}}
\toprule
Position & Size & Prompt \\
\midrule
1-1 & 1024$\times$1808 & A narrow stone alley in an old Mediterranean town early in the morning, with whitewashed walls, a blue painted wooden door on the left, and pink bougainvillea spilling over a wall on the right. Stone steps lead up and curve out of sight. A cat sits on one of the steps. The sun has just risen and lights the top of the walls while the lower alley is in cool shadow. No people are present. \\
1-2 & 1024$\times$1808 & A path winds through a dense bamboo forest in morning mist, with tall green stalks rising on both sides and leaning slightly over the path. Fallen leaves cover the stone steps in the foreground. The mist softens the deeper rows of bamboo into pale gray-green silhouettes, while the nearest stalks show their smooth segmented surfaces. Soft diffused light filters down from above, with no direct sun, and the path curves out of sight in the distance. \\
1-3 & 1808$\times$1024 & An older man with white hair and a white beard is smiling broadly at the camera. He is wearing a light blue button-up shirt with two chest pockets. The background is softly blurred, showing an indoor setting with a wooden staircase to the right and a window or door on the left. \\
1-4 & 1024$\times$1024 & A close-up of a mechanical butterfly made of brass gears, thin copper wire, and translucent stained-glass wings in amber and blue, resting on a real purple flower. Tiny screws and springs are visible in its body, and the wings catch the sunlight like real glass. The background is a soft blur of green garden. The lighting is bright daylight from the upper left, with sharp detail on the gears and a shallow depth of field. \\
1-5 & 1024$\times$1808 & A white stag with antlers that glow faintly with soft blue light stands in a snowy forest at night, surrounded by gently falling snow. Small points of light drift around its antlers. The trees are dark pines heavy with snow, and the ground is untouched white. Moonlight from above and the glow of the antlers light the snow around the stag. The rendering is photorealistic, with fine fur detail and visible breath in the cold air. \\
2-1 & 1024$\times$1808 & An anime-style illustration of a young woman with short blue-black hair, wearing a white blouse and a long navy skirt, sitting on a stone wall beside a sleeping white cat, looking out over a town at sunset. The sky is painted in gradients of orange, pink, and violet with a few soft clouds, and the rooftops below are drawn in clean lines with warm highlights. Her hair and skirt move slightly in the breeze. The line art is crisp and the shading is cel-style with soft gradients on the sky. This is a cartoon or anime illustration. \\
2-2 & 1024$\times$1808 & A tall woman in her thirties with dark skin and close-cropped hair walks across a zebra crossing in a city street, wearing a long beige trench coat belted at the waist over a black turtleneck and straight-leg cream trousers, with pointed black leather boots. She carries a structured brown leather handbag in her right hand and looks slightly off to the side. The buildings behind her are pale stone, softly out of focus. The light is bright and overcast, giving even tones with soft shadows. The photo is a full-length shot taken at knee height. \\
2-3 & 1808$\times$1024 & A weathered wooden sign on a post stands at the edge of a sandy path leading to a beach, with the words ``TO THE BEACH'' painted in faded white capital letters and a small arrow pointing right. Tall dune grass sways around the post, and the sea is visible in the distance under a clear blue sky. Bright midday sun casts a short shadow from the post onto the sand. The letters are clear and readable. \\
2-4 & 1024$\times$1024 & An assorted sushi platter arranged on a rectangular black slate board, photographed at a slight angle from above. The platter holds pieces of salmon nigiri, tuna nigiri, a shrimp nigiri, and a row of six cucumber maki rolls, with a small mound of pickled ginger and a dab of wasabi at one corner. Each slice of fish shows its glossy surface and marbling. A small dish of soy sauce and a pair of black chopsticks sit beside the board on a light wooden table. The lighting is soft and even. \\
2-5 & 1024$\times$1808 & An ink wash painting of misty mountains rising above a river, in a traditional East Asian style, with layered gray washes suggesting distant peaks and sharp black brushstrokes defining the pine trees on the nearest cliff. A small fishing boat with a single figure drifts on the pale water. Large areas of the paper are left empty to represent mist and sky. A single red seal mark sits in the lower corner. The brushwork varies from dry, scratchy strokes on the rocks to soft, wet gradients in the clouds. This is a painting. \\
\bottomrule
\end{tabular}
\end{table*}

%% file: supp/A_gallery.tex
\input{supp/A_gallery_figs}

\paragraph{Gallery of selected samples.}
The gallery uses the checkpoint of the head figure of the main paper (\cref{fig:head}). The sampling settings are those of the sampling paragraph above (APG~\cite{sadat2024apg} with guidance scale 5, momentum $-0.5$ and $\eta=0.15$; Euler method with 50 steps); the shift is either 1.6 or a value that depends on the resolution (3.88 at $1024\times1024$, 4.87 at $1808\times1024$ and $1024\times1808$). The prompt set has 223 prompts in 13 categories (portrait, people, fashion, animals, food, product, nature, city, interior, light and macro, illustration, fantasy, and text), and every prompt is generated with two seeds and both shift values, 892 images in total. The 51 samples in \GalAllFigs{} are selected from these images and ordered by category; the ten images of \cref{fig:head} come from the same selection and are not repeated here. For this document the images are downsampled to at most 1400 pixels on the long side.

\paragraph{Samples by category.}
For each of the 13 categories, \GalCatFigs{} show four further samples from the same 892 images; none of them is among the 51 samples above. \Cref{tab:a4} lists the category and prompt of every sample of both galleries under the number printed in the lower-left corner of the image.

\clearpage
\onecolumn
\input{supp/A_gallery_prompts}
\clearpage
\twocolumn

%% file: supp/A_gallery_figs.tex
\begin{figure*}[p]
\centering
\includegraphics[width=\textwidth]{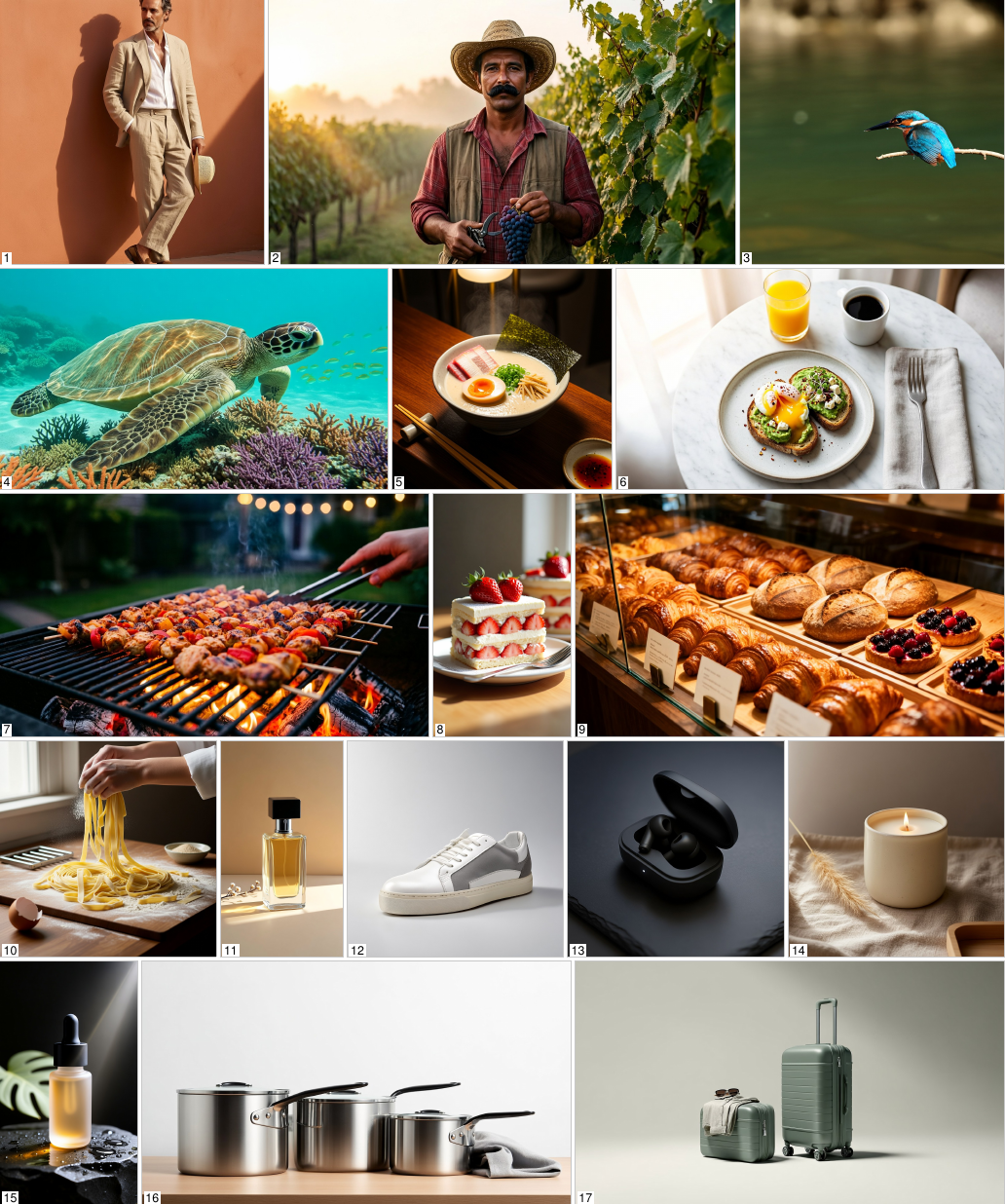}
\caption{Text-to-image samples from DyLoMo-DiT, part 1 of 3: samples 1--17 of 51, ordered by prompt category and shown at their native aspect ratios. The number in the lower-left corner of an image is its row in \cref{tab:a4}.}
\label{fig:a_gal_all_1}
\end{figure*}
\begin{figure*}[p]
\centering
\includegraphics[width=\textwidth]{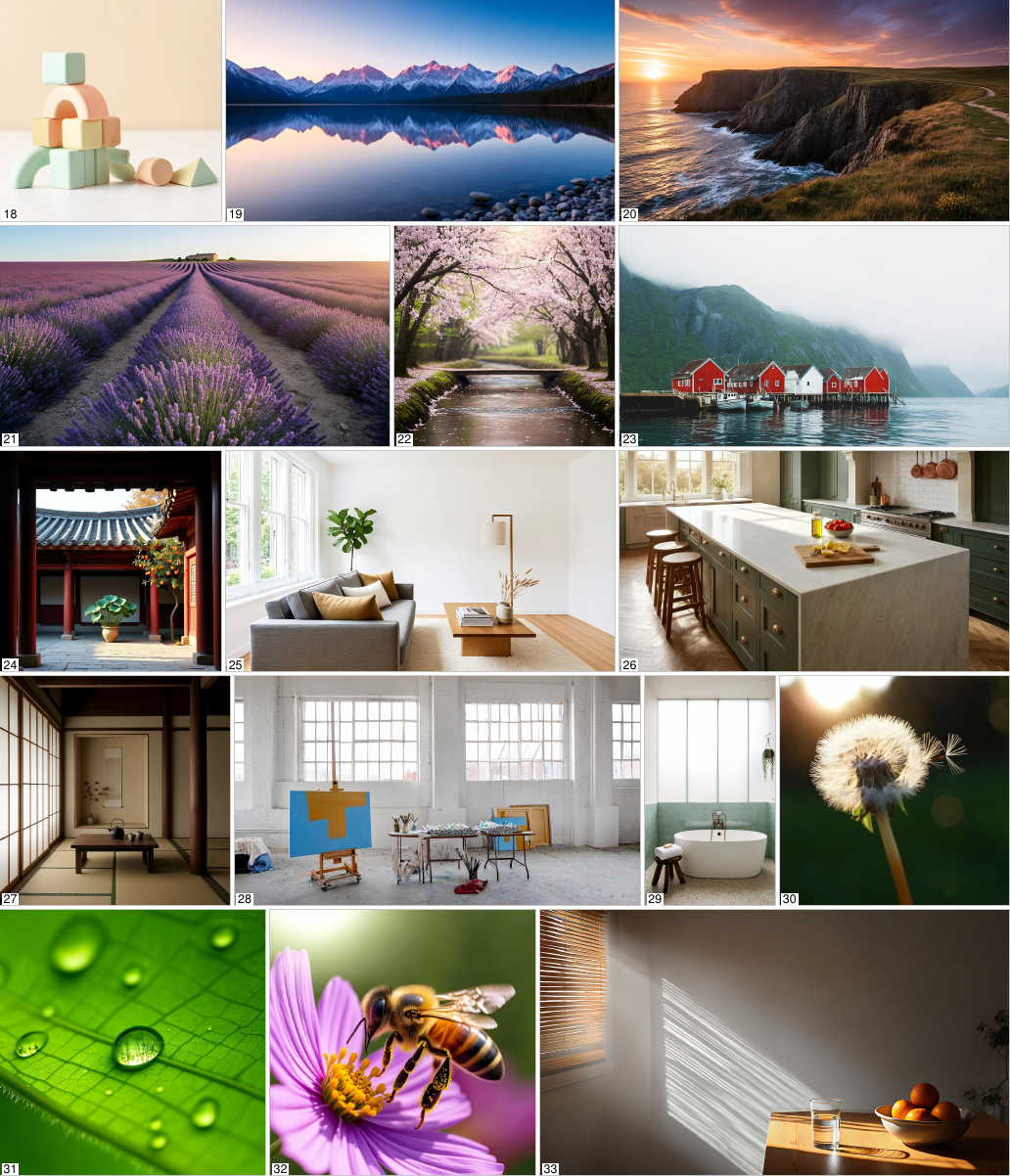}
\caption{Text-to-image samples from DyLoMo-DiT, part 2 of 3: samples 18--33 of 51, ordered by prompt category and shown at their native aspect ratios. The number in the lower-left corner of an image is its row in \cref{tab:a4}.}
\label{fig:a_gal_all_2}
\end{figure*}
\begin{figure*}[p]
\centering
\includegraphics[width=\textwidth]{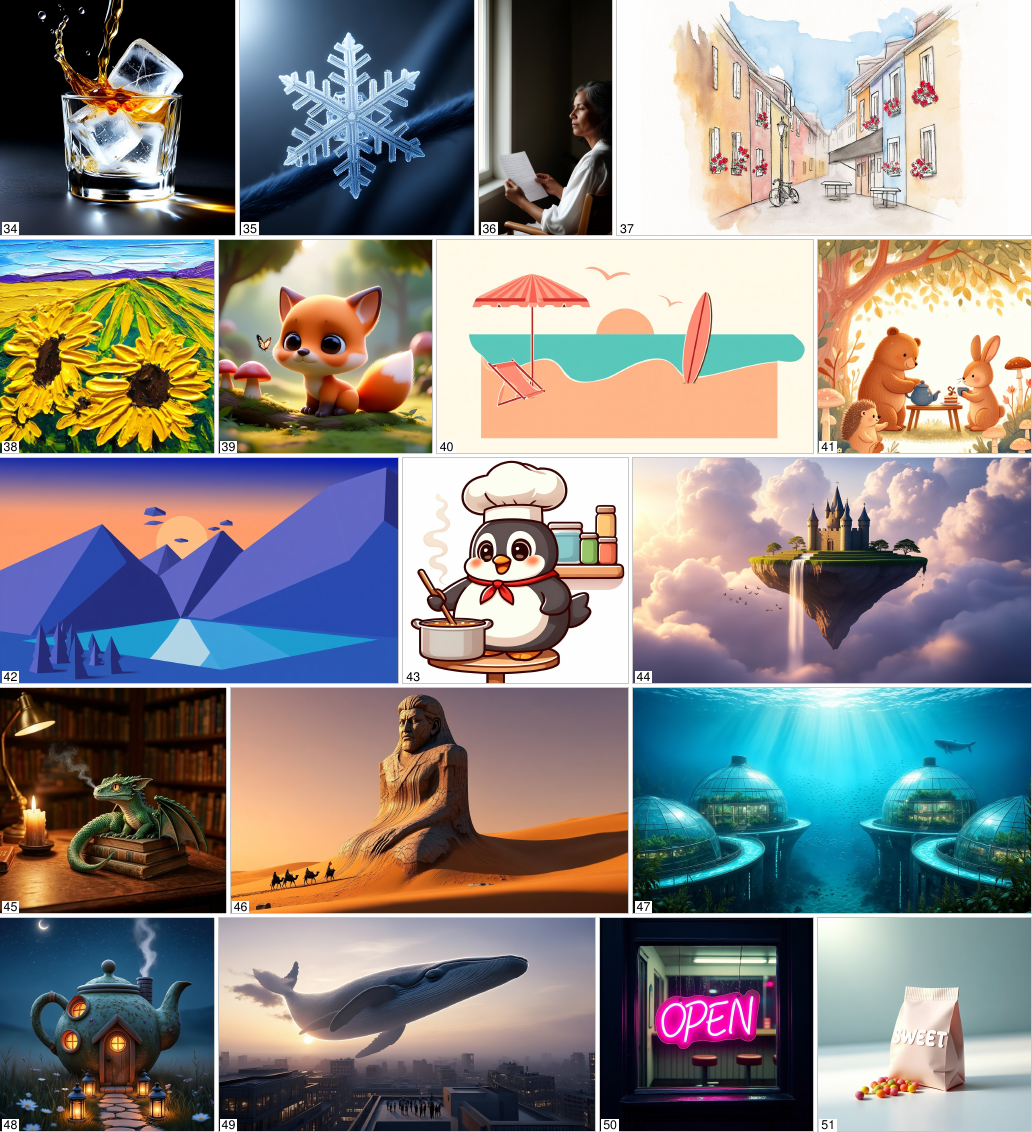}
\caption{Text-to-image samples from DyLoMo-DiT, part 3 of 3: samples 34--51 of 51, ordered by prompt category and shown at their native aspect ratios. The number in the lower-left corner of an image is its row in \cref{tab:a4}.}
\label{fig:a_gal_all_3}
\end{figure*}
\begin{figure*}[p]
\centering
\includegraphics[width=\textwidth]{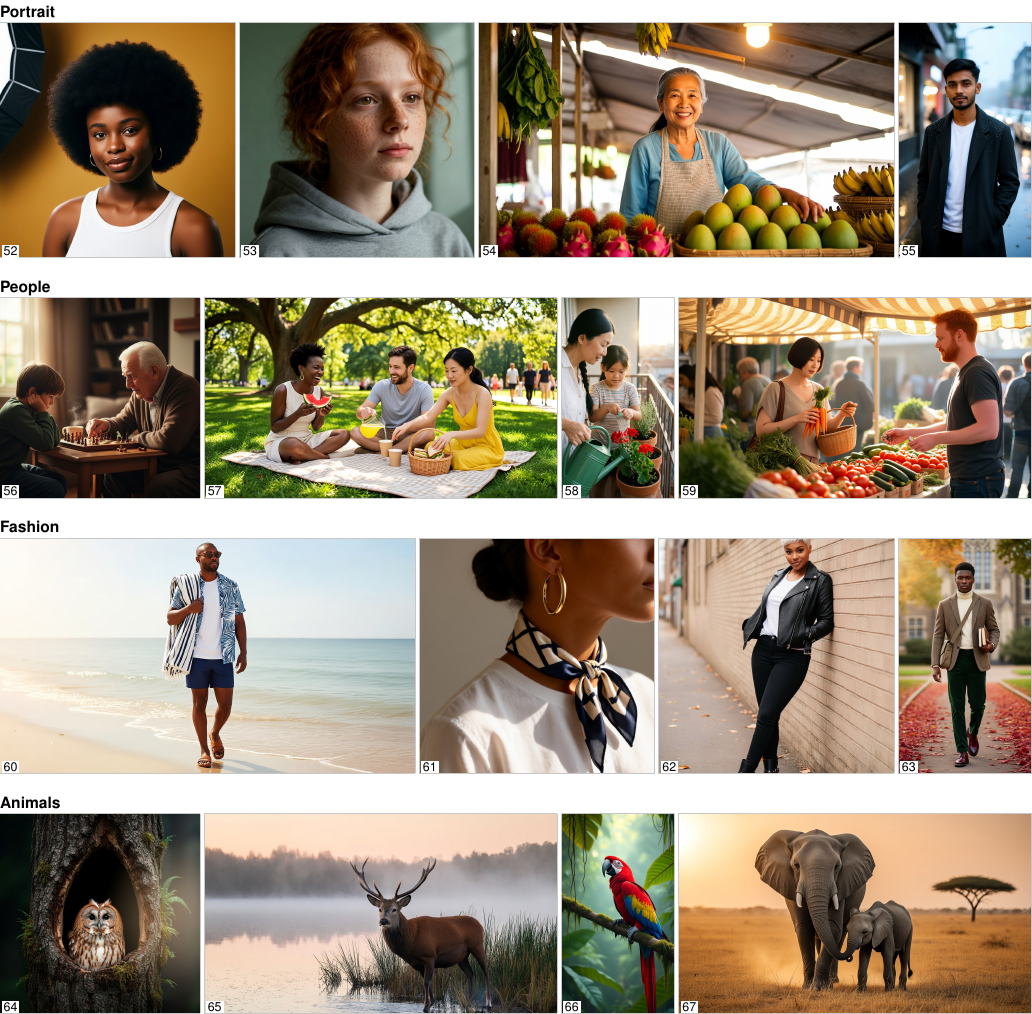}
\caption{Text-to-image samples from DyLoMo-DiT by prompt category, part 1 of 3: samples 52--67, four samples per category. Numbers refer to \cref{tab:a4}.}
\label{fig:a_gal_cat_1}
\end{figure*}
\begin{figure*}[p]
\centering
\includegraphics[width=\textwidth]{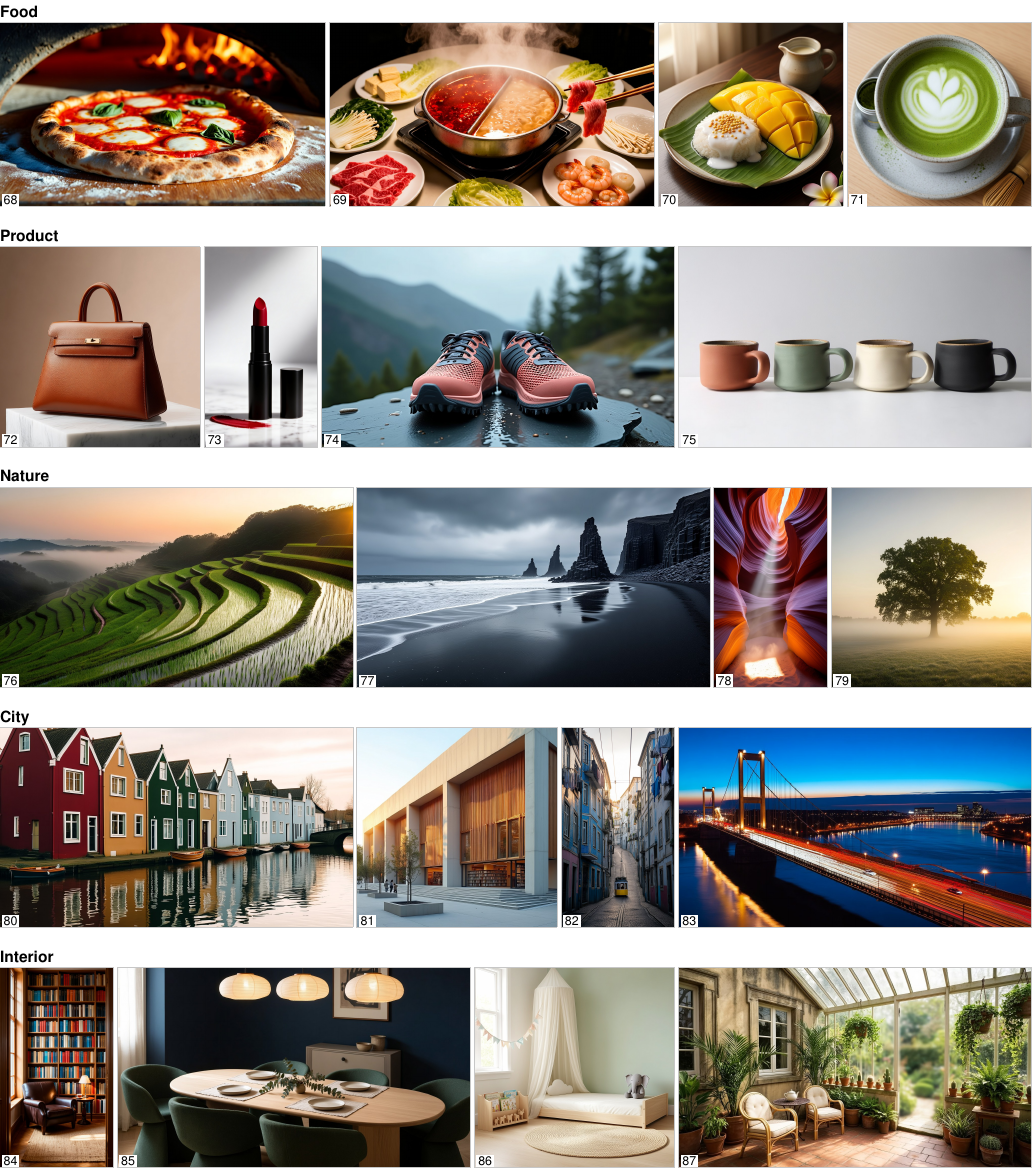}
\caption{Text-to-image samples from DyLoMo-DiT by prompt category, part 2 of 3: samples 68--87, four samples per category. Numbers refer to \cref{tab:a4}.}
\label{fig:a_gal_cat_2}
\end{figure*}
\begin{figure*}[p]
\centering
\includegraphics[width=\textwidth]{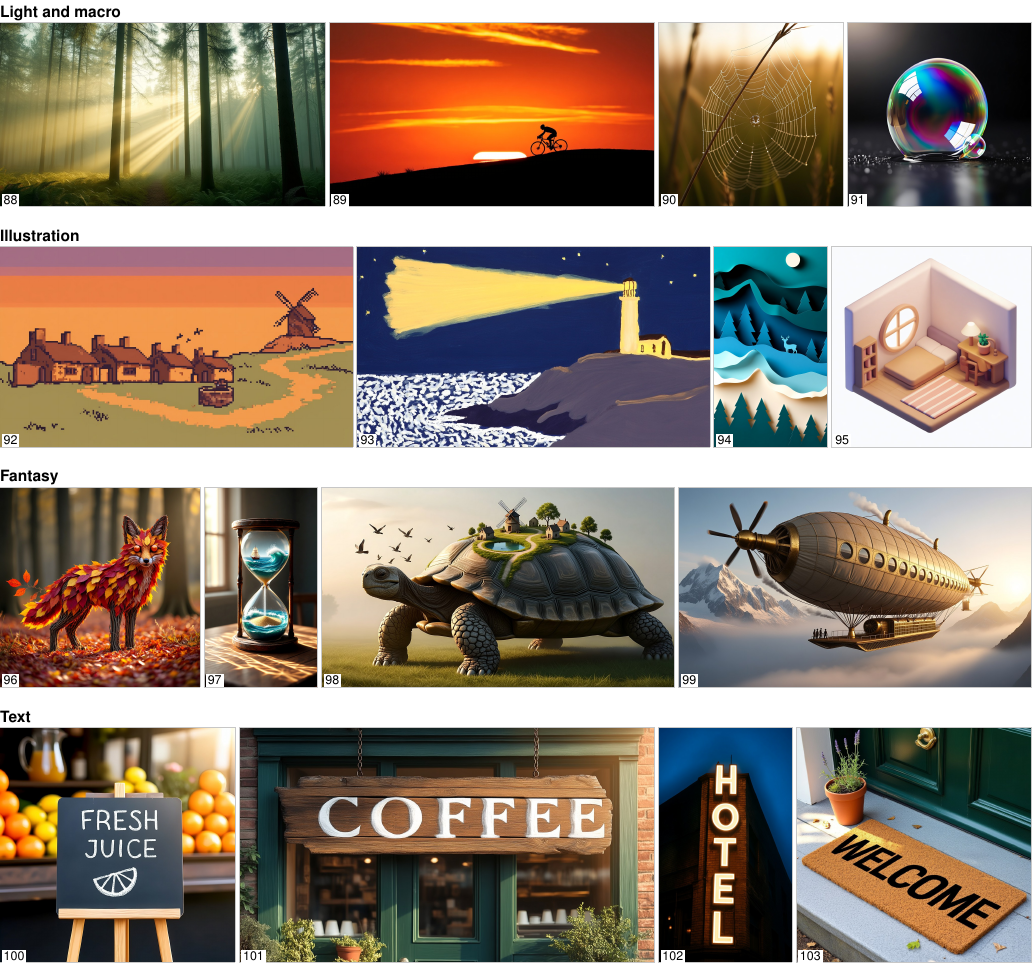}
\caption{Text-to-image samples from DyLoMo-DiT by prompt category, part 3 of 3: samples 88--103, four samples per category. Numbers refer to \cref{tab:a4}.}
\label{fig:a_gal_cat_3}
\end{figure*}

\def\GalAllFigs{\cref{fig:a_gal_all_1,fig:a_gal_all_2,fig:a_gal_all_3}}\def\GalCatFigs{\cref{fig:a_gal_cat_1,fig:a_gal_cat_2,fig:a_gal_cat_3}}

%% file: supp/A_gallery_prompts.tex
{\footnotesize\setlength{\tabcolsep}{4pt}
\begin{longtable}{@{}rlp{5.3in}@{}}
\caption{Prompts of the gallery samples (\GalAllFigs: numbers 1--51; \GalCatFigs: numbers 52--103). Prompts longer than 30 words are cut after the 30th word.}
\label{tab:a4}\\
\toprule
No. & Category & Prompt \\
\midrule
\endfirsthead
\multicolumn{3}{@{}l}{\tablename~\thetable\ (continued)}\\
\toprule
No. & Category & Prompt \\
\midrule
\endhead
\bottomrule
\endfoot
1 & Fashion & A man in his forties with olive skin, salt-and-pepper hair, and a short beard stands against a sun-washed terracotta wall, wearing an unstructured sand-colored linen suit over a white open-collar~\ldots \\
2 & Portrait & A Latino man in his forties with sun-darkened skin, a thick dark mustache, and a straw hat stands at the end of a row of grapevines at dawn, holding a~\ldots \\
3 & Animals & A common kingfisher perches on a thin bare branch above a slow river, seen in profile facing left. Its back and wings are bright metallic blue, its chest is rusty~\ldots \\
4 & Animals & A green sea turtle swims slowly over a coral reef in clear turquoise water, seen from slightly above and to the side. Its shell is patterned with olive and brown~\ldots \\
5 & Food & A bowl of tonkotsu ramen sits on a dark wooden counter, photographed from a high angle. The creamy white broth is topped with two slices of pink-edged pork belly, a~\ldots \\
6 & Food & A breakfast plate on a white marble table, seen from directly above. Two slices of toasted sourdough are topped with mashed avocado, a poached egg with its yolk running onto~\ldots \\
7 & Food & Skewers of marinated chicken and vegetables cook over glowing charcoal on an outdoor grill, seen from a low angle close to the grate. The chicken pieces are golden brown with~\ldots \\
8 & Food & A slice of layered strawberry shortcake stands on a white ceramic plate on a light wooden table. The three layers of pale sponge are separated by whipped cream and halved~\ldots \\
9 & Food & A bakery display case seen from the customer's side, filled with rows of baked goods on wooden trays: golden croissants with flaky layers, pain au chocolat, round sourdough loaves with~\ldots \\
10 & Food & A close-up of two hands dusting flour over fresh tagliatelle on a wooden board, with the long ribbons of pale yellow pasta coiled into loose nests. Beside the board lie~\ldots \\
11 & Product & A rectangular glass perfume bottle stands on a plain beige surface in front of a matching beige background. The bottle is made of thick clear glass with sharp edges and~\ldots \\
12 & Product & A single white leather sneaker with a thick cream rubber sole and gray suede panels stands at a three-quarter angle on a plain light gray surface against a matching seamless~\ldots \\
13 & Product & A pair of matte black wireless earbuds rests in an open charging case on a smooth dark slate surface. The case lid is open at an angle, showing a small~\ldots \\
14 & Product & A cream-colored ceramic candle jar with a lit wick stands on a rough-textured beige linen cloth next to a dried pampas grass stem and a small wooden tray. The candle's~\ldots \\
15 & Product & A frosted glass dropper bottle filled with clear amber serum stands upright on a wet black stone surface, with water droplets scattered around its base. The bottle has a matte~\ldots \\
16 & Product & A set of three stainless-steel cooking pots with glass lids and riveted black handles is arranged on a light wooden countertop against a plain white background, sizes stepping down from~\ldots \\
17 & Product & A hard-shell suitcase in matte sage green with four spinner wheels and a telescoping handle stands upright on a plain pale-gray studio floor, next to a matching smaller carry-on. Both~\ldots \\
18 & Product & A set of children's wooden building blocks in soft pastel colors, including cubes, arches, cylinders, and triangles in mint, peach, pale yellow, and natural beech, stacked into a small tower~\ldots \\
19 & Nature & A calm alpine lake at sunrise reflects a range of snow-capped mountains and a clear sky that shifts from pale orange near the peaks to deep blue above. The water~\ldots \\
20 & Nature & Rugged coastal cliffs drop into the sea under a sunset sky streaked with orange and purple clouds. Waves break white against the dark rocks at the base of the cliffs.~\ldots \\
21 & Nature & Rows of blooming lavender stretch toward a low hill with a single stone farmhouse in the distance, under a clear summer sky. The purple rows converge toward the horizon, and~\ldots \\
22 & Nature & Cherry trees in full bloom lean over a narrow stream, their pale pink blossoms thick on every branch. Fallen petals float on the slow-moving water and gather along the mossy~\ldots \\
23 & City & A small fishing village of wooden houses painted red and white stands on stilts along the edge of a fjord, with steep green mountains rising behind. Several small boats are~\ldots \\
24 & City & A traditional East Asian courtyard seen from inside the gate, with a gray tiled roof, upturned eaves, and red wooden pillars framing a stone-paved yard. A large ceramic pot with~\ldots \\
25 & Interior & A bright living room in a Scandinavian style, with white walls, a light oak floor, and a large window on the left letting in daylight. A gray fabric sofa with~\ldots \\
26 & Interior & A spacious kitchen with a large white marble island, dark green cabinets with brass handles, and a white subway-tile backsplash. On the island are a wooden cutting board with sliced~\ldots \\
27 & Interior & A quiet Japanese-style tea room with tatami mats, sliding paper screens on the left letting in soft light, and a low wooden table with a black iron teapot and two~\ldots \\
28 & Interior & An artist's loft studio with high ceilings, large north-facing windows, and white brick walls. A wooden easel holds a large canvas in progress with abstract blocks of blue and ochre.~\ldots \\
29 & Interior & A bathroom with a freestanding white oval bathtub on a terrazzo floor, a wooden stool holding a folded white towel and a bar of soap, and a large frosted window~\ldots \\
30 & Light and macro & A close-up of a single dandelion seed head backlit by the low evening sun, with each fine seed filament glowing white against a dark, blurred green background. A few seeds~\ldots \\
31 & Light and macro & An extreme close-up of a green leaf with water droplets resting on its surface, each drop reflecting a tiny inverted image of the surroundings. The leaf's veins form a fine~\ldots \\
32 & Light and macro & A macro photo of a honeybee landing on a purple cosmos flower, with its wings slightly blurred and its hairy legs dusted with yellow pollen. The flower's petals are sharp~\ldots \\
33 & Light and macro & Sunlight passes through a half-open wooden window blind and falls across a white wall and a wooden table in a quiet room, creating a pattern of sharp light and shadow~\ldots \\
34 & Light and macro & A close-up of clear ice cubes stacked in a short crystal glass on a dark surface, with a splash of amber liquid being poured in and frozen mid-motion. The cubes~\ldots \\
35 & Light and macro & An extreme macro of a single snowflake resting on a strand of dark blue wool, showing its six symmetrical arms with fine branching plates and sharp hexagonal facets. Tiny air~\ldots \\
36 & Light and macro & A woman in her fifties with gray-streaked dark hair and medium brown skin sits in a wooden chair beside a tall window, reading a handwritten letter held in both hands.~\ldots \\
37 & Illustration & A watercolor illustration of a quiet street in a small European town, with tall narrow houses in soft washes of ochre, rose, and pale blue, their windows shuttered and flower~\ldots \\
38 & Illustration & An oil painting of a field of sunflowers under a bright summer sky, with thick impasto brushstrokes building up the yellow petals and dark brown centers. The nearest flowers are~\ldots \\
39 & Illustration & A 3D rendered cartoon scene of a small orange fox with large round eyes and a fluffy white-tipped tail sitting on a mossy log in a sunny forest clearing. The~\ldots \\
40 & Illustration & A flat vector illustration of a beach scene with a striped umbrella, a folded deck chair, and a surfboard planted upright in the sand, with gentle stylized waves and a~\ldots \\
41 & Illustration & A children's picture book illustration of a bear, a rabbit, and a hedgehog having a tea party at a small table under a large tree, drawn with soft colored-pencil textures~\ldots \\
42 & Illustration & A low-poly 3D render of a mountain landscape at sunrise, with faceted peaks in shades of blue and purple, a flat triangular lake reflecting the sky, and a small pine~\ldots \\
43 & Illustration & A cartoon illustration of a round, chubby penguin wearing a tall white chef's hat and a small red neckerchief, standing on a stool and stirring a big pot of soup~\ldots \\
44 & Fantasy & A floating island drifts among towering cumulus clouds at golden hour, carrying a small stone castle with pointed turrets, a waterfall that pours off the island's edge and dissolves into~\ldots \\
45 & Fantasy & A small green dragon the size of a cat, with folded leathery wings and bright golden eyes, sits curled on a stack of old books on a wooden desk in~\ldots \\
46 & Fantasy & A colossal weathered stone statue of a seated figure, half-buried in desert sand, rises above a line of dunes at sunset. Its surface is cracked and streaked with centuries of~\ldots \\
47 & Fantasy & An underwater city of domed glass buildings and glowing walkways lies on the seabed in clear blue water, with schools of fish swimming between the towers and sunlight from the~\ldots \\
48 & Fantasy & A whimsical cottage shaped like a giant teapot stands in a meadow under a starry night sky, with warm light glowing from its round windows and a thin stream of~\ldots \\
49 & Fantasy & An enormous whale glides slowly through the sky above a modern city at sunset, casting a long shadow across the rooftops, with people on a rooftop terrace in the foreground~\ldots \\
50 & Text & A glowing neon sign reading ``OPEN'' in bright pink cursive letters hangs in the window of a small diner at night, with the glass reflecting the wet street outside. Behind~\ldots \\
51 & Text & A small paper candy bag in pale pink stands on a white surface, with the word ``SWEET'' printed in rounded white letters across the front. The bag is slightly open~\ldots \\
\midrule
52 & Portrait & A studio portrait of a young Black woman with a full natural afro and smooth dark skin, looking straight into the camera with a confident, closed-mouth smile. She wears a~\ldots \\
53 & Portrait & A close-up portrait of a teenage girl with red curly hair and a face covered in freckles, looking slightly past the camera with a thoughtful expression. She wears a plain~\ldots \\
54 & Portrait & A Thai woman in her sixties with gray hair tied back and a wide, warm smile stands behind her fruit stall in a covered daytime market, resting one hand on~\ldots \\
55 & Portrait & A young South Asian man in his twenties stands on a wet city sidewalk after rain, looking directly into the camera with a calm expression. He has short black hair,~\ldots \\
56 & People & An elderly man with white hair and a brown cardigan sits across a small wooden table from a boy of about nine, playing chess. The boy leans forward on his~\ldots \\
57 & People & Three friends in their late twenties sit on a checkered picnic blanket on a grassy lawn under a large oak tree. A Black woman with short natural hair laughs while~\ldots \\
58 & People & A mother and her daughter of about five water plants together on a small apartment balcony. The mother, a woman with long black hair in a loose braid and a~\ldots \\
59 & People & A couple in their forties shops at an outdoor farmers market on a bright morning. The woman, with East Asian features and a short bob, holds a woven basket and~\ldots \\
60 & Fashion & A man in his thirties with brown skin and a short beard walks along the waterline of a sandy beach, wearing an open short-sleeved shirt with a blue-and-white palm leaf~\ldots \\
61 & Fashion & A close-up of a woman's neck and jawline from the side, showing a silk scarf in a navy-and-cream geometric pattern tied loosely at the neck, a large gold hoop earring,~\ldots \\
62 & Fashion & A woman in her late twenties with light brown skin and a short platinum pixie cut leans against a plain brick wall on an autumn street, wearing a fitted black~\ldots \\
63 & Fashion & A young Black man in his early twenties walks along a tree-lined campus path in autumn, wearing a brown herringbone tweed blazer over a cream turtleneck, dark green corduroy trousers,~\ldots \\
64 & Animals & A brown and white tawny owl sits inside the hollow of an old oak tree, its round face framed by the rough bark of the opening, with wide dark eyes~\ldots \\
65 & Animals & A red deer stag with a full set of antlers stands at the edge of a misty pond at dawn, head turned toward the camera, with its front hooves in~\ldots \\
66 & Animals & A scarlet macaw perches on a thick mossy branch in a tropical forest, its head turned to the side to show the bright red plumage, yellow and blue wing feathers,~\ldots \\
67 & Animals & An adult African elephant walks slowly across dry golden grassland with a small calf close at her side, the calf's trunk curled up to touch her front leg. Their gray~\ldots \\
68 & Food & A freshly baked Margherita pizza on a wooden peel just pulled from a wood-fired brick oven, with a blistered, lightly charred crust, bubbling melted mozzarella, bright red tomato sauce, and~\ldots \\
69 & Food & A bubbling hot pot with a divided pot of red spicy broth on one side and clear mushroom broth on the other sits on a portable burner in the center~\ldots \\
70 & Food & A plate of Thai mango sticky rice on a banana leaf, with a neat mound of white sticky rice drizzled with coconut cream, sprinkled with toasted yellow mung beans, and~\ldots \\
71 & Food & A top-down close-up of a matcha latte in a wide ceramic cup with a speckled gray glaze, with a white milk-foam rosetta pattern poured into the bright green surface. The~\ldots \\
72 & Product & A structured handbag in cognac-brown pebbled leather sits on a white marble pedestal against a soft beige backdrop. The bag has two rolled top handles, a gold-toned clasp, and visible~\ldots \\
73 & Product & A single matte red lipstick in a slim black cylindrical case stands upright on a white marble surface, with the cap removed and placed beside it, the bullet of lipstick~\ldots \\
74 & Product & A pair of trail running shoes in coral and dark gray mesh with aggressive black rubber lugs stands side by side on a flat piece of wet slate rock, with~\ldots \\
75 & Product & Four handmade ceramic mugs in a row on a plain white shelf against a seamless light gray wall, glazed in muted terracotta, sage green, cream, and charcoal, each with a~\ldots \\
76 & Nature & Curved terraces of young green rice fill a steep hillside, each step holding a thin sheet of water that mirrors the pale orange sky at sunrise. Low mist drifts in~\ldots \\
77 & Nature & A black sand beach under a dramatic overcast sky, with tall basalt sea stacks rising from the water near the shore. Waves roll in with white foam that spreads thinly~\ldots \\
78 & Nature & The inside of a narrow sandstone slot canyon, with smooth, wave-like walls in layered bands of orange, red, and purple curving upward toward a thin strip of sky. A single~\ldots \\
79 & Nature & A single large oak tree stands alone in the middle of a flat meadow covered in low morning mist, its full green canopy lit from the side by the rising~\ldots \\
80 & City & A row of narrow houses painted in deep red, ochre, forest green, and pale blue stands along a calm canal, with tall windows, white trim, and steep gabled roofs. Their~\ldots \\
81 & City & The exterior of a modern public library built from pale concrete and warm wooden slats, with large windows revealing bookshelves inside. A wide set of steps leads up to the~\ldots \\
82 & City & A small yellow tram climbs a steep narrow cobblestone street between tall old apartment buildings with tiled facades in faded blue, white, and pink. Laundry hangs from a few balconies,~\ldots \\
83 & City & A long suspension bridge spans a wide river at dusk, its two tall towers and cables lit by warm lamps. Cars on the bridge form streaks of red and white~\ldots \\
84 & Interior & A cozy home study with floor-to-ceiling wooden bookshelves packed with books of many colors, a leather armchair in dark brown, and a small side table holding a reading lamp and~\ldots \\
85 & Interior & A dining room with a long oval table of pale oak surrounded by six curved chairs in dark green fabric, set under a row of three round paper pendant lamps.~\ldots \\
86 & Interior & A bright children's bedroom with a low wooden bed under a soft white canopy, a pale green wall, and a round woven rug on the light wood floor. A small~\ldots \\
87 & Interior & A glass conservatory attached to an old house, filled with potted plants of many sizes: tall palms, trailing ivy, ferns on a wooden stand, and small cacti along the windowsill.~\ldots \\
88 & Light and macro & Tall pine trees in a quiet forest at early morning, with beams of sunlight breaking through the mist between the trunks and falling in bright diagonal shafts onto a floor~\ldots \\
89 & Light and macro & A cyclist rides along the crest of a low hill, seen as a sharp black silhouette against an enormous orange and red sunset sky with thin horizontal clouds. The bicycle's~\ldots \\
90 & Light and macro & A close-up of a circular spider web stretched between two dry grass stems, each strand lined with tiny beads of dew that sparkle in the low morning sun coming from~\ldots \\
91 & Light and macro & A macro close-up of a single soap bubble resting on a wet dark surface, its thin film swirling with bands of iridescent color in blue, magenta, green, and gold. The~\ldots \\
92 & Illustration & A pixel art illustration of a small village at sunset, with a row of cottages with chimneys, a stone well, a winding dirt path, and a windmill on a hill,~\ldots \\
93 & Illustration & A gouache painting of a lighthouse on a rocky cliff at night, with a beam of warm yellow light sweeping across a dark blue sky scattered with small painted stars.~\ldots \\
94 & Illustration & A layered paper-cut illustration of a forest at dusk, with silhouetted trees in four stacked layers of paper in deep teal, mid blue, light blue, and cream, each layer casting~\ldots \\
95 & Illustration & An isometric 3D render of a small cozy room floating on a white background, with a bed under a round window, a wooden desk with a lamp and a potted~\ldots \\
96 & Fantasy & A fox standing in a quiet autumn forest, its entire body made of layered red, orange, and yellow fallen leaves, with twigs forming its legs and two bright amber berries~\ldots \\
97 & Fantasy & A tall glass hourglass standing on a wooden table, with a miniature ocean inside the upper bulb, complete with rolling waves, white foam, and a tiny sailing boat, slowly pouring~\ldots \\
98 & Fantasy & An enormous tortoise walks slowly across a grassy plain, carrying a small village on its mossy shell: stone cottages, a windmill, a pond, and a few trees, with a winding~\ldots \\
99 & Fantasy & A large airship with a weathered brass hull, canvas balloon, spinning propellers, and rows of round windows flies between snow-capped mountain peaks at sunrise. Steam trails from two pipes along~\ldots \\
100 & Text & A small chalkboard sign on a wooden easel stands in front of a fruit juice stall, with the words ``FRESH JUICE'' written in neat white chalk capital letters and a~\ldots \\
101 & Text & A rustic wooden sign hangs above the door of a small coffee shop, with the word ``COFFEE'' carved and painted in white serif capital letters. The sign hangs from two~\ldots \\
102 & Text & A tall vertical neon sign reading ``HOTEL'' in glowing warm white capital letters, one letter above the next, is mounted on the corner of an old brick building at night.~\ldots \\
103 & Text & A rectangular coir doormat with the word ``WELCOME'' printed in bold black capital letters lies on a gray stone doorstep in front of a dark green painted wooden door with~\ldots \\
\end{longtable}}

%% file: supp/B_params_compute.tex
\section{Parameters, Compute and Results of All Models and Ablations}
\label{sec:app_b}

This appendix is the listing that \cref{sec:setup} refers to: parameters, forward compute and FID over training of every model and ablation of the ImageNet experiments (\Cref{tab:b_dit,tab:b_scale,tab:b_dm}). It also gives the numbers that \cref{sec:vs_dit,sec:vs_ditmoe,sec:scale_baselines,sec:loop_role,sec:allocation} summarize: the continuation to 1M steps, the sub-figures on parameters added at equal compute and on the XL tier, the trade between layers and passes, the arms with fewer parameters and more compute, and the remaining layouts at the budget of DiT-MoE-B.

\paragraph{Compute accounting.}
The numbers are measured on the instantiated models and include the router scoring (0.07\% of the total for DyLoMo-DiT-XL) but not the score nonlinearity and the top-$k$ selection. Equal compute does not mean an equal number of layers: DyLoMo-DiT-XL has hidden width 1024 (1152 for DiT-XL/2) and 23 layer visits on average (28 layers for DiT-XL/2), so each visit costs on average about 1.2 times one layer of DiT-XL/2, with the excess coming mainly from the sparse experts. Active parameters are the parameters that one token uses, counted per layer visit (a token that selects different experts in different passes of the looped segment uses more than this count): the layers of the looped segment are counted once, the routed experts in proportion to the number selected, and everything else in full.

\begin{table*}[t]
\caption{Models at the budget of DiT and their internal ablations (ImageNet-256, EMA weights, guidance scale 1.5, 50k images). ``Layers'': physical layers, with the size of the looped segment in parentheses. ``Experts'': routing heads $\times$ experts per head (selected per head); for DiT-MoE, experts per layer (selected) plus the always-on shared expert. ``Passes'': 3\,$|$\,1 is three passes for $t<500$ and one for $t\ge500$; 3\,$|$\,2\,$|$\,1 has boundaries at $t=400$ and 600; 1\,$|$\,3 is the reverse of 3\,$|$\,1. Compute of an allocated model is the mean over the 250 sampling steps. ``Seeds'': training seeds behind the 400k value. $^\ast$Evaluated with two passes at every timestep (other schedules: \cref{sec:app_g}). ``w/o adaLN'': total parameters without the adaLN modulation layers of the blocks (one linear layer $d\rightarrow6d$ with bias per physical layer; the layers of the looped segment hold one copy), given for the configurations of \cref{tab:h_cfg}.}
\label{tab:b_dit}
\centering\footnotesize\setlength{\tabcolsep}{3.4pt}
\begin{tabular}{@{}llllrrrrrrrrr@{}}
\toprule
 & & & & \multicolumn{4}{c}{Cost} & \multicolumn{4}{c}{FID $\downarrow$ at step} & \\
\cmidrule(lr){5-8}\cmidrule(lr){9-12}
Model & \hd{Layers\\(looped)} & Experts & Passes & \hd{Total\\(M)} & \hd{w/o adaLN\\(M)} & \hd{Active\\(M)} & GMAC & 100k & 200k & 300k & 400k & Seeds \\
\midrule
\input{supp/gen_B_dit.tex} 
\bottomrule
\end{tabular}
\end{table*}

\paragraph{Models at the budget of DiT.}
\Cref{tab:b_dit} lists the nine rows of \Cref{tab:main} together with the models that share their budget. In \Cref{tab:main}, against the DiT of each tier, sFID, IS and Precision move in the same direction as FID, and Recall differs by at most 0.01; in the B tier DyLoMo-DiT is still behind DiT-B/2 at 100k steps and ahead from 200k steps on. A second B-tier layout at equal parameters and compute, with 16 physical layers and a looped middle segment, reaches 17.30 ($-18\%$ against DiT-B/2); the main text keeps the 12-layer layout, which has the depth of DiT-B/2. A second layout without the loop at the XL tier (28 layers, 5 experts per head with 4 selected) gives 5.13, against 5.02 for the 23-layer layout of \Cref{tab:main}.

\begin{table}[t]
\caption{Continuation to 1M steps under the same protocol (FID-50K, guidance scale 1.5): \Cref{tab:main} at the XL tier (top) and the DyLoMo-DiT of \Cref{tab:ditmoe} at the B tier (bottom). The 400k column repeats the main-text table; from 500k on, each row is one continued run. ``Ratio'': DyLoMo-DiT-XL over DiT-XL/2.}
\label{tab:b_1m}
\centering\footnotesize\setlength{\tabcolsep}{3.2pt}
\begin{tabular}{@{}lrrrrrrr@{}}
\toprule
Steps & 400k & 500k & 600k & 700k & 800k & 900k & 1M \\
\midrule
\input{supp/gen_B_1m.tex} 
\bottomrule
\end{tabular}
\end{table}

\begin{figure}[t]
\centering
\includegraphics[width=0.9\linewidth]{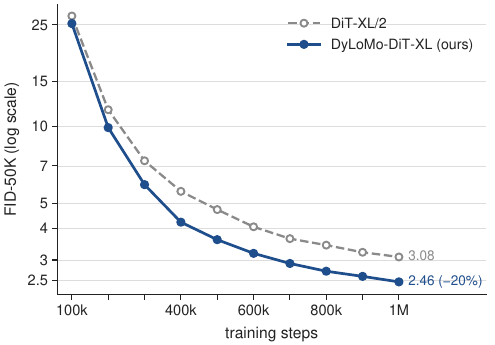}
\caption{FID-50K against training steps up to 1M for DyLoMo-DiT-XL and DiT-XL/2 (guidance scale 1.5; the numbers of \Cref{tab:b_1m}). Up to 400k steps the points are seed means; from 500k on, each curve is one continued run. In parentheses: the difference to DiT-XL/2 at 1M steps.}
\label{fig:b_1m}
\end{figure}

\paragraph{Continuation to 1M steps.}
Trained on to 1M steps under the same protocol, DyLoMo-DiT-XL and DiT-XL/2 reach 2.46 and 3.08 (\Cref{tab:b_1m,fig:b_1m}). The FID of DyLoMo-DiT-XL decreases monotonically from 500k to 1M steps (3.60, 3.19, 2.91, 2.71, 2.59, 2.46), and its ratio to DiT-XL/2 stays between 0.76 and 0.80 over the six evaluation points. At the budget of DiT-MoE-B (\Cref{tab:ditmoe}), DyLoMo-DiT trained on to 1M steps reaches 3.41; its FID decreases monotonically over the six evaluation points (5.72, 4.78, 4.32, 3.86, 3.57 and 3.41 from 500k to 1M steps; 7.01 at 400k). The original configuration of DiT-MoE-B, continued to 500k steps, gives 10.45: DyLoMo-DiT is 45\% below it at 500k steps, after 46\% at 400k steps (12.91, \cref{sec:vs_ditmoe}).

\paragraph{Beyond 1M steps.}
The two XL runs of \Cref{tab:b_1m} were continued further under the same protocol. At 1.6M steps, DyLoMo-DiT-XL and DiT-XL/2 reach 2.18 and 2.67 (\cref{fig:1m}); at 2.2M steps, they reach 2.07 and 2.52. DyLoMo-DiT-XL is 18\% below DiT-XL/2 at both points, and the ratio of the two FIDs is 0.82 at both, after 0.80 at 1M steps, so the relative gap is stable from 1.6M to 2.2M steps. Each of these points is one run per model.

\begin{table*}[t]
\caption{XL tier beyond the budget of DiT-XL/2 (setting and columns of \Cref{tab:b_dit}). ``$\times$P'' and ``$\times$C'': total parameters and compute relative to DiT-XL/2 (675.1M, 118.6 GMAC). The dense models with a loop replace every sparse-expert feed-forward network by a dense one of width 5184 and remove the routing.}
\label{tab:b_scale}
\centering\footnotesize\setlength{\tabcolsep}{2.6pt}
\begin{tabular}{@{}llllrrrrrrrrrr@{}}
\toprule
 & & & & \multicolumn{5}{c}{Cost} & \multicolumn{4}{c}{FID $\downarrow$ at step} & \\
\cmidrule(lr){5-9}\cmidrule(lr){10-13}
Model & \hd{Layers\\(looped)} & Experts & Passes & \hd{Total\\(M)} & \hd{Active\\(M)} & GMAC & $\times$P & $\times$C & 100k & 200k & 300k & 400k & Seeds \\
\midrule
\input{supp/gen_B_scale.tex} 
\bottomrule
\end{tabular}
\end{table*}

\begin{figure*}[t]
\centering
\includegraphics[width=\textwidth]{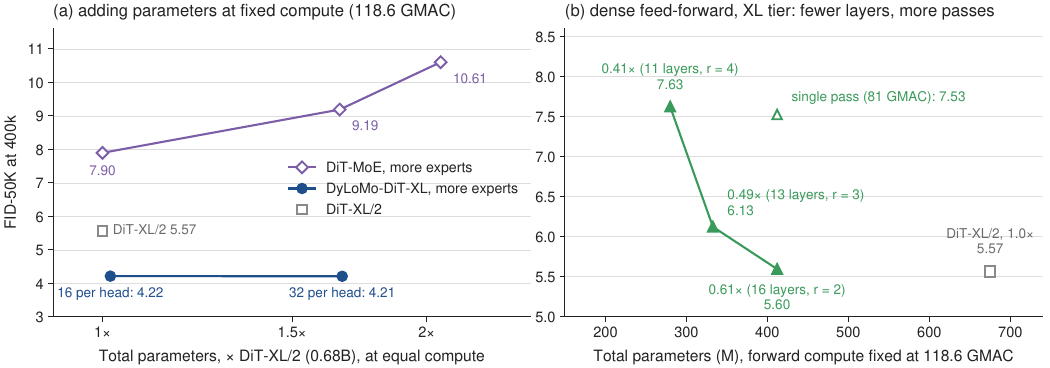}
\caption{Parameters at fixed compute, XL tier (ImageNet-256, 400k steps, 118.6 GMAC). (a) Parameters added through experts: more experts make DiT-MoE worse and DyLoMo-DiT slightly better. (b) Parameters removed from a dense model through the loop, from DiT-XL/2 ($1\times$) down: the $0.61\times$ model matches DiT-XL/2, and at $0.41\times$ the model is back at the level of the $0.61\times$ model trained with a single pass.}
\label{fig:b_params}
\end{figure*}

\paragraph{Larger baselines and parameters added at equal compute.}
\begin{figure}[t]
\centering
\includegraphics[width=0.86\linewidth]{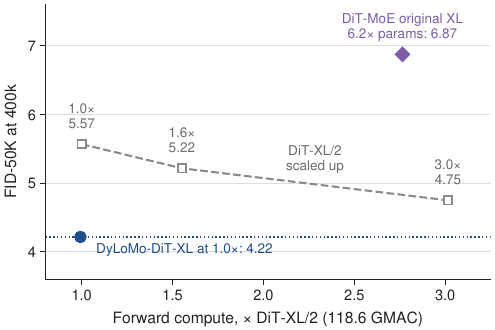}
\caption{Larger baselines against DyLoMo-DiT-XL at $1\times$ (400k steps): DiT-XL/2 scaled up in parameters and compute, and the original XL DiT-MoE ($6.2\times$ the parameters). Neither reaches it. For readability the figure omits the DiT at $2.2\times$ (4.57) and the DiT-MoE at $2.2\times$ (6.65), which are listed in \Cref{tab:b_scale}. DiT-XL/2 and DyLoMo-DiT-XL are means of three and two seeds; the other models are single runs.}
\label{fig:scale}
\end{figure}

The scaled DiT-XL/2 models of \cref{fig:scale} and \cref{sec:scale_baselines} have hidden widths 1440, 1728 and 2016, with parameters and compute growing together (\Cref{tab:b_scale}); the DiT-MoE configurations are described in \cref{sec:app_j}. \Cref{fig:b_params}a is the sub-figure on parameters added at equal compute: from $1.0\times$ to $1.7\times$ and $2.1\times$ the parameters of DiT-XL/2, the FID of DiT-MoE rises from 7.90 to 9.19 and 10.61, whereas DyLoMo-DiT-XL with 32 instead of 16 experts per head ($1.6\times$ its parameters) goes from 4.22 to 4.21 (the latter with two passes at every timestep; the model with 16 experts per head under the same schedule gives 4.36, \Cref{tab:b_scale}).

\paragraph{Parameter floor with a dense feed-forward.}
\Cref{fig:b_params}b is the dense counterpart of \cref{fig:loop}a. With a dense feed-forward at 118.6 GMAC, a model with $0.61\times$ the parameters of DiT-XL/2 (16 layers, two passes) matches it, $0.49\times$ (13 layers, three passes) is 10\% worse and $0.41\times$ (11 layers, four passes) 37\% worse, back at the level of a single pass.

\begin{figure}[t]
\centering
\includegraphics[width=\linewidth]{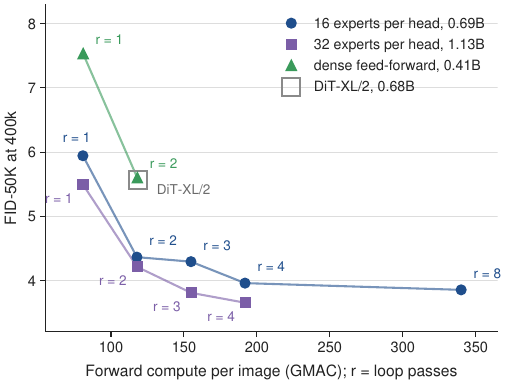}
\caption{Parameters fixed, compute added by passes, XL tier (ImageNet-256, 400k steps): 16 experts per head (0.69B), the variant with 32 experts per head (1.13B) and the dense model with a loop (0.41B); DiT-XL/2 for reference. The gains continue to four passes in both sparse-expert families and, with a smaller step, to eight passes in the family with 16 experts per head.}
\label{fig:b_passes}
\end{figure}

\paragraph{Passes at the XL tier.}
At the XL tier the gains continue beyond four passes (\cref{fig:b_passes}): the model with 16 experts per head (0.69B) gives 5.94, 4.36, 4.29, 3.96 and 3.85 with one, two, three, four and eight passes, four passes being 9\% below two and eight passes 2.6\% below four, and the variant with 32 experts per head (1.13B) gives 5.49, 4.21, 3.81 and 3.65 from one to four. The return diminishes: from two to four passes the compute rises from $1.0\times$ to $1.6\times$ that of DiT-XL/2 and FID falls by 0.40, whereas from four to eight it rises from $1.6\times$ to $2.9\times$ and FID falls by 0.10. The eight-pass model is at 21.81, 8.13, 5.05 and 3.85 after 100k, 200k, 300k and 400k steps (\Cref{tab:b_scale}).

\paragraph{Loop and allocation at each budget.}
\Cref{tab:b_loopalloc} separates the two steps of the loop at the four budgets at which all three models are trained: the model without the loop, the loop with two passes at every timestep, and the loop with passes allocated by noise level, at the same total parameters and compute within a budget. \input{supp/gen_B_loopalloc_txt}

\begin{table}[t]
\caption{Loop and allocation at four budgets (FID at 400k steps). The three models of a row have the same total parameters and compute. The last two columns give the FID change of two passes at every timestep relative to no loop and of allocated passes relative to two passes at every timestep, computed from unrounded values.}
\label{tab:b_loopalloc}
\centering\footnotesize\setlength{\tabcolsep}{2pt}
\begin{tabular}{@{}lrrrrr@{}}
\toprule
Budget & \hd{No\\loop} & \hd{Two passes\\at every $t$} & \hd{Allocated\\by noise} & \hd{Two passes\\vs.\ no loop} & \hd{Allocated vs.\\two passes} \\
\midrule
\input{supp/gen_B_loopalloc.tex} 
\bottomrule
\end{tabular}
\end{table}

\paragraph{Layers against passes.}
With both parameters and compute fixed, layers and passes are interchangeable between two and four passes: in the B tier (single routing head, budget of DiT-MoE-B), 12 layers with three passes and 24 experts per layer give 7.89 against 7.94 for 16 layers with two passes; in the XL tier, 11 layers with four passes give 4.44 against 4.36 for 16 layers with two, while eight passes over 9 layers (5.87) are worse (\Cref{tab:b_dit,tab:b_dm}).

\begin{table*}[t]
\caption{Models at the budget of DiT-MoE (setting and columns of \Cref{tab:b_dit}). The S, B and L rows in bold are the DyLoMo-DiT rows of \Cref{tab:ditmoe}; the ``Active'' column gives the active parameters that \Cref{tab:ditmoe} omits. 5\,$|$\,1 is five passes for $t<500$ and one for $t\ge500$.}
\label{tab:b_dm}
\centering\footnotesize\setlength{\tabcolsep}{3.4pt}
\begin{tabular}{@{}llllrrrrrrrr@{}}
\toprule
 & & & & \multicolumn{3}{c}{Cost} & \multicolumn{4}{c}{FID $\downarrow$ at step} & \\
\cmidrule(lr){5-7}\cmidrule(lr){8-11}
Model & \hd{Layers\\(looped)} & Experts & Passes & \hd{Total\\(M)} & \hd{Active\\(M)} & GMAC & 100k & 200k & 300k & 400k & Seeds \\
\midrule
\input{supp/gen_B_dm.tex} 
\bottomrule
\end{tabular}
\end{table*}

\paragraph{Fewer parameters with more compute; five low-noise passes.}
The three models of \cref{fig:loop}a with fewer parameters at 63 GMAC have 12 layers and three passes ($0.72\times$), 10 layers and four passes ($0.58\times$) and 6 layers and five passes ($0.31\times$); the $0.58\times$ model is still more than 20\% below both DiT-MoE-B configurations. Two further models remove parameters and add compute (\Cref{tab:b_dm}): 14 layers with a looped segment of 6 and three passes ($0.86\times$ the parameters, $1.19\times$ the compute) give 35.19, 16.98, 10.84 and 8.03 at 100k to 400k steps, level with the two-pass model at $1\times$ (7.94), and 12 layers with a looped segment of 8 and three passes ($0.72\times$, $1.28\times$) give 31.42, 15.25, 9.93 and 7.31, not higher than the two-pass model (7.94). At the same compute ($1.28\times$), the 16-layer model with three passes and $1\times$ the parameters gives 6.72; at $0.72\times$ the parameters and $1\times$ the compute (12 layers, looped segment of 5), FID is 9.04. For passes at fixed parameters (\cref{fig:loop}b), the model with five passes for $t<500$ and one for $t\ge500$ (three passes on average) gives 6.47, level with four passes at every timestep (6.54) at 18\% less compute.

\paragraph{Allocation against two passes at every timestep.}
The FID pairs behind the percentages of \cref{sec:allocation} are $-4.2\%$ in the B tier (18.82 vs.\ 19.65), $-4.7\%$ in L (6.41 vs.\ 6.73) and $-3.4\%$ in XL (4.22 vs.\ 4.36) at the DiT budget, all as means over the available seeds (\Cref{tab:h_seeds}), and $-12\%$ for the single-head B model at the DiT-MoE budget (7.01 vs.\ 7.94); \Cref{tab:g_alloc} places them next to the three-band and the reverse allocation.

\paragraph{DiT-MoE budget: evaluation points and other layouts.}
In \Cref{tab:ditmoe}, all four evaluation points go the same way in every tier, with the gap widening from S to B and similar in B and L: the ratio of the FID of DyLoMo-DiT to that of DiT-MoE goes from 0.80 at 100k steps to 0.64 at 400k in the S tier, from 0.62 to 0.44 in B and from 0.56 to 0.44 in L (\Cref{tab:b_dm}). The tiers of DiT-MoE have about 6.1 times the parameters and 2.6 to 2.8 times the compute of the DiT of the same name. \Cref{tab:b_dm} also lists the layouts at the budget of DiT-MoE-B that \Cref{tab:path} leaves out: with three routing heads, 12 layers instead of 16 give 6.12 (46 experts per head) and 6.65 (16 experts per head); a narrower model (width 512, two heads, 38 experts per head) gives 6.35; and the layouts without a loop give 6.05 (18 layers, three heads), 7.76 (16 layers, two heads of 8 experts) and 9.13 (12 layers, compute spent on selecting 12 experts per head).

\begin{table}[t]
\caption{More experts at equal compute in the L and B tiers, two passes at every timestep (setting and columns of \Cref{tab:b_dit}; only the costs and the 400k value are repeated here).}
\label{tab:b_lb32}
\centering\footnotesize\setlength{\tabcolsep}{3pt}
\begin{tabular}{@{}lrrrrr@{}}
\toprule
Model & \hd{Total\\(M)} & \hd{Active\\(M)} & GMAC & \hd{FID $\downarrow$\\(400k)} & Seeds \\
\midrule
\input{supp/gen_B_lb32s.tex} 
\bottomrule
\end{tabular}
\end{table}

\paragraph{More experts in the L and B tiers.}
As at the XL tier, raising the number of experts per head from 17 to 32 at equal compute lowers FID in the L and B tiers (\Cref{tab:b_lb32}).

%% file: supp/gen_B_dit.tex
\tierrow{13}{XL tier (budget of DiT-XL/2)}
DiT-XL/2 & 28 & dense & --- & 675.1 & 452.0 & 675.1 & 118.6 & 27.02 & 11.62 & 7.33 & 5.57 & 3 \\
DyLoMo-DiT-XL & 16 (7) & 4$\times$16 (6) & 3\,$|$\,1 & 686.6 & 585.8 & 411.3 & 117.8 & 25.24 & 9.89 & 5.91 & \textbf{4.22} & 2 \\
\ \ two passes at every $t$ & 16 (7) & 4$\times$16 (6) & 2 & 686.6 & 585.8 & 411.3 & 117.8 & 25.31 & 9.74 & 5.82 & 4.36 & 1 \\
\ \ three bands & 16 (7) & 4$\times$16 (6) & 3\,$|$\,2\,$|$\,1 & 686.6 & 585.8 & 411.3 & 117.8 & 25.53 & 10.06 & 6.08 & 4.39 & 1 \\
\ \ pass count drawn from $\{1,2,3\}$$^\ast$ & 16 (7) & 4$\times$16 (6) & 2 & 686.6 & 585.8 & 411.3 & 117.8 & 27.25 & 11.07 & 6.50 & 4.68 & 1 \\
Same recipe, no loop & 23 & 4$\times$8 (6) & 1 & 678.4 & --- & 595.8 & 117.7 & 26.88 & 11.11 & 6.85 & 5.02 & 1 \\
No loop, second layout & 28 & 4$\times$5 (4) & 1 & 676.4 & --- & 625.3 & 118.0 & 26.92 & 11.07 & 6.83 & 5.13 & 1 \\
11 layers, four passes & 11 (4) & 4$\times$29 (6) & 4 & 686.5 & --- & 279.5 & 117.8 & 27.31 & 10.45 & 6.14 & 4.44 & 1 \\
9 layers, eight passes & 9 (2) & 4$\times$39 (6) & 8 & 681.0 & --- & 226.8 & 117.9 & 35.59 & 14.87 & 8.58 & 5.87 & 1 \\
DiT-MoE scaled to this budget & 28 & 3 (2) + shared & --- & 675.1 & --- & 629.0 & 118.6 & 34.85 & 16.93 & 10.67 & 7.90 & 1 \\
\tierrow{13}{L tier (budget of DiT-L/2)}
DiT-L/2 & 24 & dense & --- & 458.1 & 307.0 & 458.1 & 80.7 & 32.03 & 15.08 & 9.77 & 7.42 & 3 \\
DyLoMo-DiT-L & 12 (6) & 4$\times$17 (6) & 3\,$|$\,1 & 451.3 & 375.7 & 278.3 & 80.1 & 31.52 & 14.02 & 8.62 & \textbf{6.41} & 2 \\
\ \ two passes at every $t$ & 12 (6) & 4$\times$17 (6) & 2 & 451.3 & 375.7 & 278.3 & 80.1 & 32.67 & 15.01 & 9.52 & 6.73 & 3 \\
\ \ three bands & 12 (6) & 4$\times$17 (6) & 3\,$|$\,2\,$|$\,1 & 451.3 & 375.7 & 278.3 & 80.1 & 30.79 & 14.00 & 8.83 & 6.33 & 1 \\
Same recipe, no loop & 18 & 4$\times$17 (14) & 1 & 456.2 & --- & 423.2 & 80.9 & 32.13 & 14.94 & 9.50 & 6.97 & 1 \\
\tierrow{13}{B tier (budget of DiT-B/2)}
DiT-B/2 & 12 & dense & --- & 130.5 & 88.0 & 130.5 & 23.0 & 52.61 & 33.68 & 25.68 & 21.19 & 1 \\
DyLoMo-DiT-B & 12 (6) & 2$\times$17 (6) & 3\,$|$\,1 & 128.5 & 109.6 & 75.8 & 23.1 & 55.18 & 33.51 & 24.36 & \textbf{18.82} & 2 \\
\ \ two passes at every $t$ & 12 (6) & 2$\times$17 (6) & 2 & 128.5 & 109.6 & 75.8 & 23.1 & 57.89 & 35.21 & 25.53 & 19.65 & 2 \\
\ \ three bands & 12 (6) & 2$\times$17 (6) & 3\,$|$\,2\,$|$\,1 & 128.5 & 109.6 & 75.8 & 23.1 & 54.78 & 33.29 & 24.16 & 18.93 & 1 \\
\ \ reverse allocation & 12 (6) & 2$\times$17 (6) & 1\,$|$\,3 & 128.5 & 109.6 & 75.8 & 23.1 & 62.12 & 41.14 & 31.77 & 26.14 & 1 \\
Same recipe, no loop & 18 & 2$\times$8 (6) & 1 & 130.2 & --- & 114.8 & 23.1 & 55.51 & 33.72 & 24.49 & 19.45 & 1 \\
16-layer layout & 16 (6) & 2$\times$17 (6) & 3\,$|$\,1 & 130.6 & --- & 87.1 & 23.0 & 52.57 & 30.86 & 22.14 & 17.30 & 1 \\

%% file: supp/gen_B_1m.tex
DyLoMo-DiT-XL & 4.22 & 3.60 & 3.19 & 2.91 & 2.71 & 2.59 & 2.46 \\
DiT-XL/2 & 5.57 & 4.72 & 4.05 & 3.64 & 3.43 & 3.22 & 3.08 \\
\midrule
Ratio & 0.76 & 0.76 & 0.79 & 0.80 & 0.79 & 0.80 & 0.80 \\
\midrule
DyLoMo-DiT-B & 7.01 & 5.72 & 4.78 & 4.32 & 3.86 & 3.57 & 3.41 \\

%% file: supp/gen_B_scale.tex
\tierrow{14}{Larger baselines}
DiT-XL/2, width 1440 & 28 & dense & --- & 1054.2 & 1054.2 & 184.0 & 1.56 & 1.55 & 25.40 & 10.87 & 6.98 & 5.22 & 1 \\
DiT-XL/2, width 1728 & 28 & dense & --- & 1517.3 & 1517.3 & 263.7 & 2.25 & 2.22 & 22.52 & 9.34 & 5.93 & 4.57 & 1 \\
DiT-XL/2, width 2016 & 28 & dense & --- & 2064.6 & 2064.6 & 357.7 & 3.06 & 3.02 & 22.55 & 9.69 & 6.18 & 4.75 & 1 \\
DiT-MoE, width 1536 & 28 & 3 (2) + shared & --- & 1517.0 & 1412.8 & 263.7 & 2.25 & 2.22 & --- & --- & --- & 6.65 & 1 \\
DiT-MoE, original XL & 28 & 8 (2) + shared & --- & 4168.2 & 1492.7 & 328.0 & 6.17 & 2.76 & 31.62 & 14.98 & 9.53 & 6.87 & 1 \\
\tierrow{14}{Parameters added at equal compute}
DiT-MoE, experts 768, shared 1536 & 28 & 8 (2) + shared & --- & 1121.1 & 675.2 & 118.7 & 1.66 & 1.00 & 37.20 & 18.96 & 12.37 & 9.19 & 1 \\
DiT-MoE, experts 1232, shared 616 & 28 & 8 (2) + shared & --- & 1391.3 & 676.0 & 118.9 & 2.06 & 1.00 & 40.73 & 21.34 & 14.12 & 10.61 & 1 \\
DyLoMo-DiT-XL, 32 per head & 16 (7) & 4$\times$32 (6) & 2 & 1127.2 & 411.5 & 117.9 & 1.67 & 0.99 & 25.93 & 9.90 & 5.76 & 4.21 & 3 \\
\tierrow{14}{Compute added by passes at fixed parameters}
16 experts per head, one pass & 16 (7) & 4$\times$16 (6) & 1 & 686.6 & 411.3 & 80.8 & 1.02 & 0.68 & 30.38 & 13.19 & 8.11 & 5.94 & 1 \\
\ \ two passes & 16 (7) & 4$\times$16 (6) & 2 & 686.6 & 411.3 & 117.8 & 1.02 & 0.99 & 25.31 & 9.74 & 5.82 & 4.36 & 1 \\
\ \ three passes & 16 (7) & 4$\times$16 (6) & 3 & 686.6 & 411.3 & 154.8 & 1.02 & 1.31 & 25.00 & 9.44 & 5.81 & 4.29 & 1 \\
\ \ four passes & 16 (7) & 4$\times$16 (6) & 4 & 686.6 & 411.3 & 191.8 & 1.02 & 1.62 & 22.86 & 8.55 & 5.11 & 3.96 & 1 \\
\ \ eight passes & 16 (7) & 4$\times$16 (6) & 8 & 686.6 & 411.3 & 339.9 & 1.02 & 2.87 & 21.81 & 8.13 & 5.05 & 3.85 & 1 \\
32 experts per head, one pass & 16 (7) & 4$\times$32 (6) & 1 & 1127.2 & 411.5 & 80.8 & 1.67 & 0.68 & 31.24 & 13.06 & 7.71 & 5.49 & 2 \\
\ \ three passes & 16 (7) & 4$\times$32 (6) & 3 & 1127.2 & 411.5 & 154.9 & 1.67 & 1.31 & 23.53 & 8.63 & 5.09 & 3.81 & 1 \\
\ \ four passes & 16 (7) & 4$\times$32 (6) & 4 & 1127.2 & 411.5 & 192.0 & 1.67 & 1.62 & 22.41 & 8.01 & 4.72 & 3.65 & 1 \\
\tierrow{14}{Dense feed-forward with a loop (no experts)}
16 layers, two passes & 16 (7) & dense & 2 & 412.0 & 412.0 & 118.0 & 0.61 & 1.00 & 24.95 & 11.53 & 7.34 & 5.60 & 2 \\
16 layers, one pass & 16 (7) & dense & 1 & 412.0 & 412.0 & 80.9 & 0.61 & 0.68 & 30.64 & 15.00 & 9.91 & 7.53 & 2 \\
13 layers, three passes & 13 (5) & dense & 3 & 332.7 & 332.7 & 118.0 & 0.49 & 1.00 & 25.10 & 12.05 & 8.03 & 6.13 & 1 \\
11 layers, four passes & 11 (4) & dense & 4 & 279.9 & 279.9 & 118.0 & 0.41 & 1.00 & 28.63 & 14.55 & 10.01 & 7.63 & 1 \\
\tierrow{14}{Other members with 32 experts per head}
No loop, 11 selected per head & 16 & 4$\times$32 (11) & 1 & 1127.2 & 549.2 & 116.0 & 1.67 & 0.98 & 28.62 & 11.51 & 6.91 & 4.98 & 1 \\
11 layers, four passes & 11 (4) & 4$\times$32 (6) & 4 & 740.3 & --- & 117.9 & 1.10 & 0.99 & 29.71 & 12.42 & 7.12 & 5.16 & 1 \\

%% file: supp/gen_B_loopalloc_txt.tex
Against the model without the loop, the loop with two passes at every timestep changes FID by $-13.1\%$, $-3.4\%$ and $+1.0\%$ at the budgets of DiT-XL/2, DiT-L/2 and DiT-B/2 and by $+0.1\%$ at the budget of DiT-MoE-B; allocating the passes by noise level then changes it by $-3.4\%$, $-4.7\%$, $-4.2\%$ and $-11.7\%$.

%% file: supp/gen_B_loopalloc.tex
DiT-XL/2 & 5.02 & 4.36 & 4.22 & $-13.1\%$ & $-3.4\%$ \\
DiT-L/2 & 6.97 & 6.73 & 6.41 & $-3.4\%$ & $-4.7\%$ \\
DiT-B/2 & 19.45 & 19.65 & 18.82 & $+1.0\%$ & $-4.2\%$ \\
DiT-MoE-B & 7.93 & 7.94 & 7.01 & $+0.1\%$ & $-11.7\%$ \\

%% file: supp/gen_B_dm.tex
\tierrow{12}{S tier (budget of DiT-MoE-S)}
DiT-MoE-S, 16 experts, 4 selected & 12 & 16 (4) + shared & --- & 199.3 & 71.9 & 16.0 & 71.30 & 47.45 & 36.96 & 30.83 & 1 \\
DyLoMo-DiT, single head & 16 (6) & 1$\times$16 (2) & 3\,$|$\,1 & 199.6 & 56.9 & 16.1 & 56.78 & 33.99 & 24.81 & \textbf{19.79} & 1 \\
\tierrow{12}{L tier (budget of DiT-MoE-L)}
DiT-MoE-L, 16 experts, 4 selected & 24 & 16 (4) + shared & --- & 2824.0 & 1012.0 & 222.5 & 35.64 & 17.27 & 10.94 & 7.92 & 1 \\
DyLoMo-DiT, single head & 16 (6) & 1$\times$16 (2) & 3\,$|$\,1 & 2821.8 & 712.0 & 222.5 & 20.11 & 7.60 & 4.56 & \textbf{3.50} & 1 \\
\tierrow{12}{B tier (budget of DiT-MoE-B): baselines}
DiT-MoE-B, 16 experts, 4 selected & 12 & 16 (4) + shared & --- & 795.9 & 286.3 & 62.9 & 49.82 & 29.26 & 20.73 & 16.06 & 1 \\
DiT-MoE-B, original (8 experts, 2 selected) & 12 & 8 (2) + shared & --- & 795.9 & 286.3 & 62.9 & 45.39 & 24.90 & 16.91 & 12.91 & 1 \\
\tierrow{12}{B tier: single routing head (rows 2--4 of \Cref{tab:path}; pass and parameter ladders of \cref{fig:loop})}
No loop, 22 layers & 22 & 1$\times$10 (2) & 1 & 779.7 & 316.7 & 62.9 & 34.44 & 16.36 & 10.57 & 7.93 & 1 \\
16 layers, one pass & 16 (6) & 1$\times$16 (2) & 1 & 796.2 & 229.0 & 45.2 & 40.90 & 20.64 & 13.57 & 10.06 & 1 \\
\ \ two passes at every $t$ & 16 (6) & 1$\times$16 (2) & 2 & 796.2 & 229.0 & 62.9 & 35.11 & 16.87 & 10.90 & 7.94 & 1 \\
\ \ three passes at every $t$ & 16 (6) & 1$\times$16 (2) & 3 & 796.2 & 229.0 & 80.6 & 31.40 & 14.56 & 9.36 & 6.72 & 1 \\
\ \ four passes at every $t$ & 16 (6) & 1$\times$16 (2) & 4 & 796.2 & 229.0 & 98.3 & 29.85 & 13.86 & 8.93 & 6.54 & 1 \\
\ \ allocated, two passes on average & 16 (6) & 1$\times$16 (2) & 3\,$|$\,1 & 796.2 & 229.0 & 62.9 & 31.05 & 14.78 & 9.72 & \textbf{7.01} & 1 \\
\ \ allocated, three passes on average & 16 (6) & 1$\times$16 (2) & 5\,$|$\,1 & 796.2 & 229.0 & 80.6 & 29.18 & 13.25 & 8.67 & 6.47 & 1 \\
12 layers, three passes, 24 experts & 12 (5) & 1$\times$24 (2) & 3 & 807.1 & 170.5 & 62.9 & 33.97 & 16.05 & 10.68 & 7.89 & 1 \\
12 layers, three passes & 12 (5) & 1$\times$16 (2) & 3 & 575.6 & 170.4 & 62.9 & 35.25 & 17.81 & 12.10 & 9.04 & 1 \\
10 layers, four passes & 10 (4) & 1$\times$16 (2) & 4 & 465.3 & 141.2 & 62.9 & 37.40 & 19.62 & 13.40 & 10.11 & 1 \\
6 layers, five passes & 6 (4) & 1$\times$16 (2) & 5 & 244.6 & 82.6 & 62.9 & 38.83 & 21.03 & 14.48 & 11.20 & 1 \\
14 layers, 6 looped, three passes & 14 (6) & 1$\times$16 (2) & 3 & 685.9 & 199.7 & 74.7 & 35.19 & 16.98 & 10.84 & 8.03 & 1 \\
12 layers, 8 looped, three passes & 12 (8) & 1$\times$16 (2) & 3 & 575.6 & 170.4 & 80.6 & 31.42 & 15.25 & 9.93 & 7.31 & 1 \\
\tierrow{12}{B tier: several routing heads (rows 5--6 of \Cref{tab:path}; other layouts)}
Three heads, 16 experts per head & 16 (6) & 3$\times$16 (2) & 3\,$|$\,1 & 796.7 & 229.5 & 63.1 & 32.19 & 13.42 & 7.98 & 5.61 & 1 \\
Three heads, 46 experts per head & 16 (6) & 3$\times$46 (6) & 3\,$|$\,1 & 797.3 & 229.6 & 63.1 & 32.31 & 12.68 & 7.55 & 5.28 & 1 \\
12 layers, 46 experts per head & 12 (6) & 3$\times$46 (6) & 3\,$|$\,1 & 797.0 & 199.8 & 63.1 & 35.79 & 15.27 & 8.93 & 6.12 & 1 \\
12 layers, 16 experts per head & 12 (6) & 3$\times$16 (2) & 3\,$|$\,1 & 800.8 & 199.6 & 63.0 & 34.97 & 15.66 & 9.49 & 6.65 & 1 \\
Width 512, 16 layers & 16 (7) & 2$\times$38 (6) & 3\,$|$\,1 & 793.1 & 187.5 & 62.7 & 36.23 & 15.61 & 9.13 & 6.35 & 1 \\
No loop, 18 layers & 18 & 3$\times$27 (6) & 1 & 806.6 & 305.0 & 63.0 & 33.02 & 14.41 & 8.60 & 6.05 & 1 \\
No loop, 16 layers, two heads & 16 & 2$\times$8 (2) & 1 & 795.1 & 298.1 & 62.9 & 36.69 & 17.28 & 10.71 & 7.76 & 1 \\
No loop, 12 layers, 12 selected per head & 12 & 3$\times$46 (12) & 1 & 797.0 & 289.4 & 63.8 & 39.77 & 19.45 & 12.51 & 9.13 & 1 \\

%% file: supp/gen_B_lb32s.tex
\tierrow{6}{L tier}
17 experts per head & 451.3 & 278.3 & 80.1 & 6.73 & 3 \\
32 experts per head & 687.4 & 278.4 & 80.1 & 5.88 & 2 \\
\tierrow{6}{B tier}
17 experts per head & 128.5 & 75.8 & 23.1 & 19.65 & 2 \\
32 experts per head & 200.5 & 75.9 & 23.1 & 17.40 & 1 \\

%% file: supp/C_image_to_video.tex
\section{From Image to Video}
\label{sec:app_c}

This appendix supports the statement of \cref{sec:intro,sec:conclusion} that the design transfers to video. It tests whether the weights of the text-to-image model of \cref{sec:t2i} can serve directly as the starting point for video generation. The architecture is unchanged and no parameters are added; the input becomes video latents: an 81-frame video is encoded by the same VAE into 21 latent frames, 9,408 tokens per video. The task is image-to-video: the first latent frame is kept clean and excluded from the loss, which covers only the remaining 20 frames. Training uses an in-house video set, with a separate held-out portion as the validation set; the video stage loads weights only from an image checkpoint, with a global batch of 96 videos, learning rate $5\times10^{-5}$ and cosine decay. Evaluation uses the validation loss and a small number of sample videos.

\begin{table}[t]
\caption{Validation loss after training with the same video recipe from different image checkpoints (lower is better). Stages 3 and 4 are successive stages of the image training.}
\label{tab:c1}
\centering\footnotesize
\begin{tabular}{lcc}
\toprule
Image starting point & Video, 1,000 steps & Video, 4,000 steps \\
\midrule
End of stage 3 & 0.2425 & 0.2139 \\
Stage 4, 37\% & 0.2401 & --- \\
Stage 4, 50\% & 0.2396 & 0.2117 \\
Stage 4, 75\% & --- & 0.2109 \\
End of stage 4 & 0.2374 & 0.2100 \\
\bottomrule
\end{tabular}
\end{table}

Video training started directly from the image weights converges normally. The longer the image stage is trained, the lower the video validation loss: at both evaluation points, 1,000 and 4,000 video steps, the loss improves monotonically with the image starting point. Training from the end of stage 4 to 8,000 steps brings the validation loss down to 0.1993, and it is still decreasing.

\begin{figure*}[t]
\centering
\includegraphics[width=\textwidth]{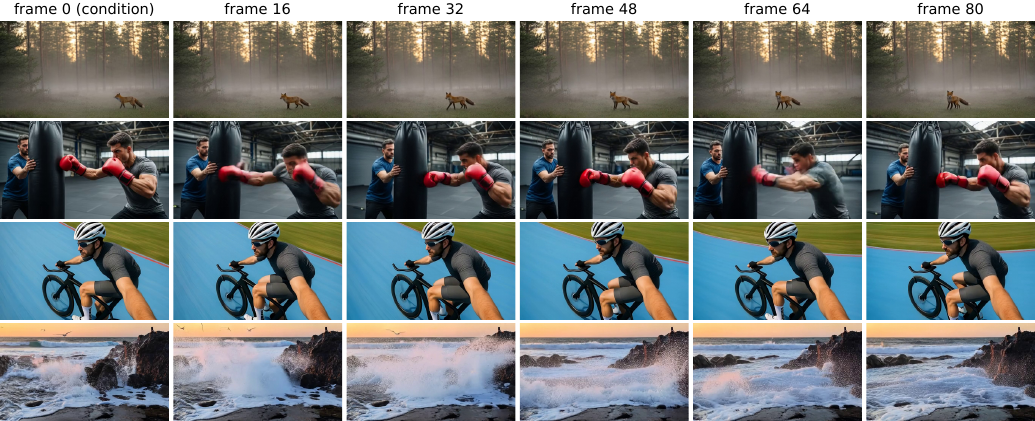}
\caption{Image-to-video samples after 8,000 steps of video training from the end of stage 4. Each row is one video (81 frames, 16 fps), showing frames 0, 16, 32, 48, 64 and 80 from left to right. Frame 0 is the conditioning frame, generated from the text description by the text-to-image model (end of stage 4), not taken from a real video.}
\label{fig:video}
\end{figure*}

%% file: supp/D_in256_pairs.tex
\section{Further ImageNet Sample Pairs}
\label{sec:app_d}

\begin{figure*}[t]
\centering
\includegraphics[width=\textwidth]{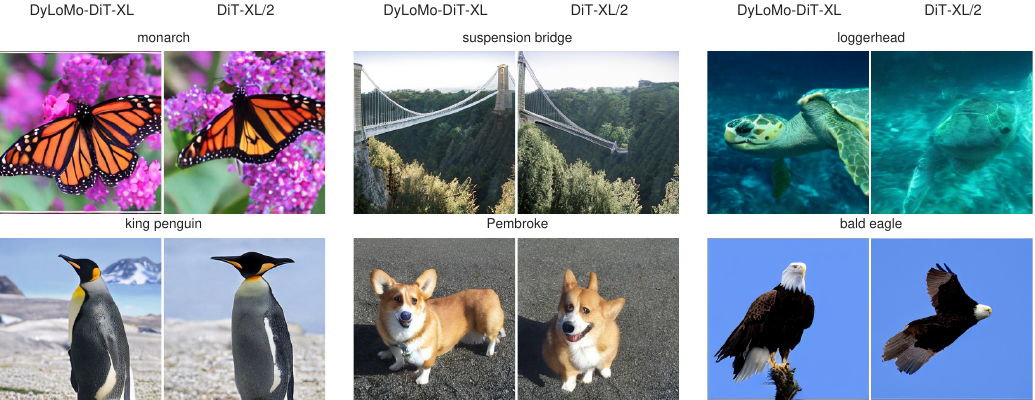}
\caption{Additional samples for the left half of \cref{fig:compare}, in the same setting (compute of DiT-XL/2; 400k-step EMA weights, guidance scale 4.0, 250 DDPM steps). In each pair, DyLoMo-DiT-XL is on the left and DiT-XL/2 on the right.}
\label{fig:d_xl}
\end{figure*}

\begin{figure*}[t]
\centering
\includegraphics[width=\textwidth]{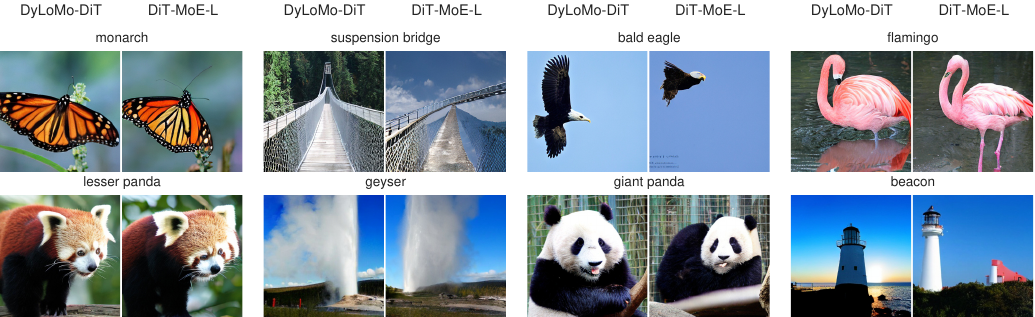}
\caption{Further same-noise pairs at the budget of DiT-MoE-L (2.82B parameters, 222 GMAC; 400k-step EMA weights, guidance scale 4.0, 250 DDPM steps). In each pair, the DyLoMo-DiT of \Cref{tab:ditmoe} (L row) is on the left and DiT-MoE-L on the right.}
\label{fig:d_ditmoe}
\end{figure*}

This appendix adds the sample pairs that \cref{fig:compare} leaves out: six further pairs against DiT-XL/2 (\cref{fig:d_xl}) and eight further pairs against DiT-MoE-L (\cref{fig:d_ditmoe}). The setting is that of \cref{fig:compare}: 400k-step EMA weights, guidance scale 4.0, 250 DDPM steps, and the two images of a pair use the same class label, the same initial noise and the same per-step noise. As in the main text, most pairs share a composition (monarch, king penguin, flamingo, lesser panda) and differ in the integrity of the subject and in the details; in the loggerhead pair of \cref{fig:d_xl} the DiT-XL/2 sample is barely recognizable as a turtle, and in the bald eagle pair of \cref{fig:d_ditmoe} the DiT-MoE-L sample has text-like artifacts at the lower left. These are single samples per noise and serve only as a qualitative reference.

%% file: supp/E_scale_curve.tex
\section{Scaling Curve of the Text-to-Image Architecture}
\label{sec:app_e}

This appendix supports the statement of \cref{sec:intro} that the design scales in the text-to-image setting (\cref{sec:t2i}). It tests whether the same architecture keeps improving with more parameters across scales. All three scales use the architecture of the text-to-image model (the middle half of the layers executed in two passes, sparse experts routed per head, and a multi-stream residual with 4 streams), changing only the width, the number of layers and the number of experts (\Cref{tab:e1}). All are trained from random initialization for 6,000 steps with the same data, batch and learning-rate schedule; the validation set is fixed, and the timestep and noise of every image are fixed and identical across the three scales sample by sample. For each scale a control without multi-stream residual is also trained. We report the validation loss at 6,000 steps.

\begin{table*}[t]
\caption{The three configurations of the scaling curve and their validation loss at 6,000 steps. In all three, the looped segment is the middle half of the layers, executed in two passes, and each head selects 6 experts. Active parameters are counted per image token, with the layers of the looped segment counted once. oHC: the multi-stream residual of \cref{sec:app_i}.}
\label{tab:e1}
\centering\footnotesize\setlength{\tabcolsep}{4pt}
\begin{tabular}{lrrcrrcc}
\toprule
Scale & Width & \hd{Physical\\layers} & \hd{Heads $\times$\\experts per head} & \hd{Total\\parameters} & \hd{Active\\parameters} & \hd{Validation loss\\(with oHC)} & \hd{Validation loss\\(without oHC)} \\
\midrule
s & 512 & 24 & $2\times64$ & 2.87B & 0.35B & 0.2904 & 0.2927 \\
m & 1024 & 32 & $4\times128$ & 15.48B & 1.05B & 0.2828 & 0.2847 \\
l & 1536 & 32 & $6\times192$ & 34.73B & 1.74B & 0.2797 & 0.2813 \\
\bottomrule
\end{tabular}
\end{table*}

\begin{figure*}[t]
\centering
\includegraphics[width=\textwidth]{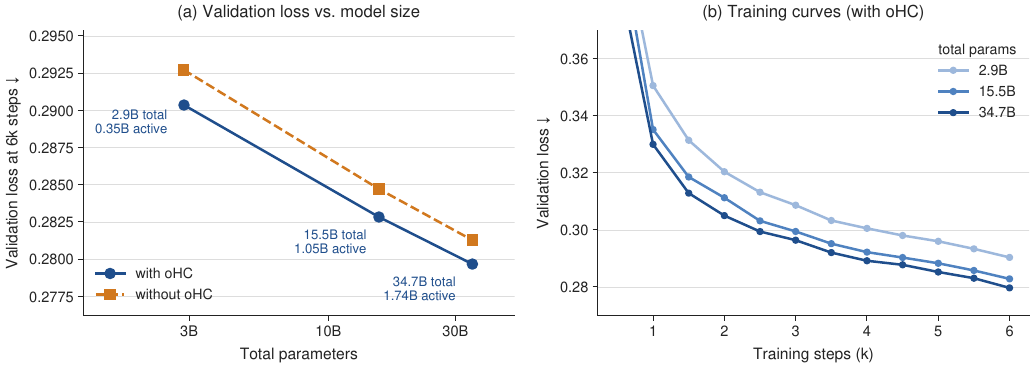}
\caption{Scaling curve of the text-to-image architecture. (a) Validation loss at 6,000 steps against total parameters; (b) validation-loss curves of the three scales with multi-stream residual, evaluated every 500 steps.}
\label{fig:e_scale}
\end{figure*}

Validation loss decreases monotonically with scale: 2.6\% lower from 2.87B to 15.48B, and a further 1.1\% lower at 34.73B (\cref{fig:e_scale}a); the training curves keep the same order at all 12 evaluation points (\cref{fig:e_scale}b). The group without multi-stream residual also decreases monotonically with scale. At all three scales the version with multi-stream residual has the lower validation loss, by 0.8\%, 0.7\% and 0.6\% respectively. These differences are reported for description only.

%% file: supp/F_routing_extra.tex
\section{Mechanism Analysis: Protocol and Controls}
\label{sec:app_f}

This appendix gives what \cref{sec:mechanism} refers to: the measurement protocol, the control trained with two passes at every timestep, the routing change by group of layers, the position at which the timestep embedding changes abruptly, and the routing switch of a model without the loop.

\paragraph{Protocol.}
All statistics of \cref{sec:mechanism} use the 400k-step EMA weights of the DyLoMo-DiT-XL of \Cref{tab:main}; the inputs are 2,048 images drawn at random from the training set, each noised once at each of the ten timesteps $t=50,150,\ldots,950$ and run through the conditional branch only, without guidance. The number of experts selected again and the cosine and norm ratio of the residual writes are computed per token (and per head), then averaged with equal weight over layers, timesteps and heads; for the allocated-pass model they compare the first and the second pass and are counted at $t\le450$, where the model runs more than one pass. Expert distributions, their JS divergence and the number of experts in use (the exponential of the entropy of the distribution) are computed per layer and head, and the main text reports the mean over all layers and heads. ``More than 80\% of its calls in one range'' compares the low-noise range $t\le550$ with the high-noise range $t\ge650$ on the average number of calls per timestep, because the two ranges contain different numbers of grid timesteps, and with the calls of the looped layers counted per pass, because the allocated model runs three passes on one side of $t=500$ and one on the other. The finer grid has step 10 from $t=460$ to 640 and includes $t=499$ and 500.

\begin{table}[t]
\caption{Routing statistics of the allocated-pass DyLoMo-DiT-XL of \Cref{tab:main} and of the control of the same structure trained with two passes at every timestep (400k-step EMA weights). Each head selects 6 of 16 experts, and the 14 sparse-expert layers hold 896 experts in total.}
\label{tab:f_ctrl}
\centering\footnotesize\setlength{\tabcolsep}{3pt}
\begin{tabular}{@{}p{4.45cm}rr@{}}
\toprule
Statistic & Allocated & Control \\
\midrule
Experts selected again in the second pass ($t\le450$) & 4.61 & 4.57 \\
\ \ same position in another image & 2.57 & 2.61 \\
\ \ uniform random selection & 2.25 & 2.25 \\
JS divergence, adjacent grid timesteps inside a range (nats) & 0.005 & 0.005 \\
JS divergence, $t=550\rightarrow650$ (nats) & 0.30 & 0.26 \\
Switch on the finer grid & $550\rightarrow560$ & $560\rightarrow570$ \\
\ \ JS divergence at the switch (nats) & 0.287 & 0.261 \\
Experts with more than 80\% of their calls in one range & 47\% & 36\% \\
\ \ in the low-noise / high-noise range & 300 / 121 & 209 / 112 \\
Experts in use per head, $t=50$ / $t=950$ & 15.0 / 9.2 & 15.0 / 10.6 \\
\bottomrule
\end{tabular}
\end{table}

\begin{figure}[t]
\centering
\includegraphics[width=\linewidth]{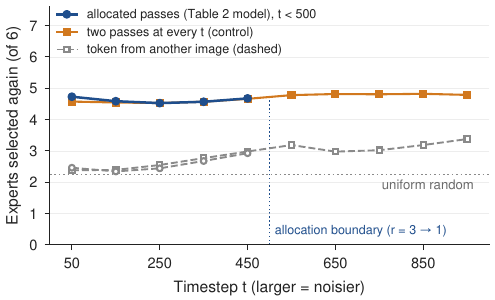}
\caption{Number of identical experts selected by the same token in two passes of the same layer (6 of 16 per head), for the allocated-pass model, which runs more than one pass only for $t<500$, and for the control trained with two passes at every timestep; dashed: the token at the same position in another image; dotted: uniform random selection. The second pass keeps most of the experts of the first at every timestep.}
\label{fig:f_overlap}
\end{figure}

\paragraph{Two-pass control.}
\Cref{tab:f_ctrl} places the allocated-pass model next to the control, and \cref{fig:f_overlap} shows the number of experts selected again per timestep. In the allocated model the same token selects 4.61 of the same experts in the first and the second pass (4.34--4.96 over layers and timesteps), 4.16 in the first and the third, and 4.71 in the second and the third; the control gives 4.57. The two passes write in the same direction: the cosine between their residual writes is 0.68 for attention and 0.62 for the feed-forward, positive for more than 99.9\% of tokens, and the second write has 1.42 and 1.49 times the norm of the first. The control also splits the denoising process into two ranges within one step ($560\rightarrow570$, against $550\rightarrow560$ for the allocated model), with a smaller share of experts that are specific to one range (36\% against 47\%).

\begin{table}[t]
\caption{JS divergence (nats) of the expert distribution between adjacent timesteps of the finer grid, by group of layers (layers 1 and 16 have dense feed-forward networks and no routing). ``Switch'': $550\rightarrow560$ for the allocated model and $560\rightarrow570$ for the control; at every other pair of the grid except $499\rightarrow500$ the overall value is at most 0.0005.}
\label{tab:f_groups}
\centering\footnotesize\setlength{\tabcolsep}{3.2pt}
\begin{tabular}{@{}lrrrr@{}}
\toprule
 & \multicolumn{2}{c}{Allocated} & \multicolumn{2}{c}{Control} \\
\cmidrule(lr){2-3}\cmidrule(lr){4-5}
Layers & Switch & \hd{$499$\\$\rightarrow500$} & Switch & \hd{$499$\\$\rightarrow500$} \\
\midrule
Before the loop (2--4) & 0.260 & 0.0011 & 0.354 & 0.0000 \\
Looped segment (5--11) & 0.266 & 0.0026 & 0.173 & 0.0000 \\
After the loop (12--15) & 0.344 & 0.0012 & 0.345 & 0.0000 \\
All & 0.287 & 0.0019 & 0.261 & 0.0000 \\
\bottomrule
\end{tabular}
\end{table}

\begin{figure}[t]
\centering
\includegraphics[width=0.93\linewidth]{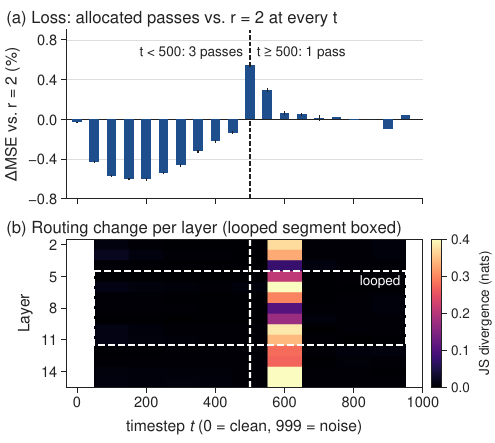}
\caption{Routing change per layer (DyLoMo-DiT-XL). (a) repeats the per-timestep loss of \cref{fig:noise}a so that the timestep axes align. (b) JS divergence (nats) of the expert distribution between adjacent grid timesteps, for each layer (dashed box: looped segment; dashed line: allocation boundary $t=500$). Every layer switches between $t=550$ and 650; elsewhere the divergence is about 0.005, too small to show on this color scale.}
\label{fig:f_layers}
\end{figure}

\paragraph{JS divergence by group of layers.}
Every layer of the allocated model switches at the same place, before, inside and after the looped segment (\Cref{tab:f_groups}; per layer on the coarse grid: \cref{fig:f_layers}b). The switch does not sit at the allocation boundary: the JS divergence from $t=499$ to 500 is 0.002 overall and 0.001 in the layers before the loop, whose routing does not depend on the number of passes, two orders of magnitude below the main switch.

\begin{table}[t]
\caption{Position of the routing switch and of the largest change of the learned timestep embedding between adjacent timesteps, for every model in which both were measured (400k-step EMA weights). The routing grid has step 10 for the first four models and step 1 for the last two; ``change'' is the relative change of the embedding at its largest step.}
\label{tab:f_temb}
\centering\footnotesize\setlength{\tabcolsep}{2.6pt}
\begin{tabular}{@{}p{3.55cm}ccr@{}}
\toprule
Model & \hd{Routing\\switch} & \hd{Embedding,\\largest step} & Change \\
\midrule
DyLoMo-DiT-XL (\Cref{tab:main}) & $550\rightarrow560$ & $555\rightarrow556$ & 0.60 \\
\ \ two passes at every $t$ & $560\rightarrow570$ & $567\rightarrow568$ & 0.76 \\
B, three heads (\Cref{tab:path}, row 5) & $600\rightarrow610$ & $600\rightarrow601$ & 0.75 \\
B, single head (\Cref{tab:ditmoe}) & $570\rightarrow580$ & $572\rightarrow573$ & 0.58 \\
XL, 32 experts per head, two passes & $558\rightarrow559$ & $558\rightarrow559$ & 0.78 \\
\ \ one pass (no loop) & $563\rightarrow564$ & $563\rightarrow564$ & 0.82 \\
\bottomrule
\end{tabular}
\end{table}

\begin{figure}[t]
\centering
\includegraphics[width=\linewidth]{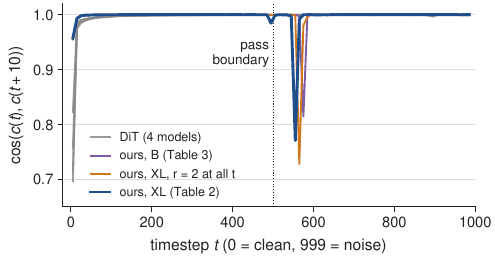}
\caption{Cosine between the learned timestep embeddings of two adjacent timesteps ($t$ and $t+10$; 400k-step EMA weights): the DyLoMo-DiT-XL of \Cref{tab:main}, its control with two passes at every timestep, the DyLoMo-DiT of \Cref{tab:ditmoe} at the B tier, and four DiT models trained under the same setting (three XL/2 seeds and one L/2, gray). Each DyLoMo-DiT model has one abrupt change between $t=550$ and 580, where its routing switches (\Cref{tab:f_temb}); DiT changes abruptly only between $t=0$ and 10. The allocated XL model also has a small dip at its pass boundary (cosine 0.98 from $t=490$ to 500).}
\label{fig:f_temb}
\end{figure}

\paragraph{Timestep embedding.}
In every model in which we measured both, the learned timestep embedding changes abruptly at the same place as the routing (\Cref{tab:f_temb}): the largest step of the embedding falls inside the grid interval of the routing switch, between $t=555$ and 601 depending on the model. A DiT trained under the same setting has a single abrupt change near $t=0$ (\cref{fig:f_temb}). We report the coincidence and draw no causal conclusion.

\begin{figure}[t]
\centering
\includegraphics[width=\linewidth]{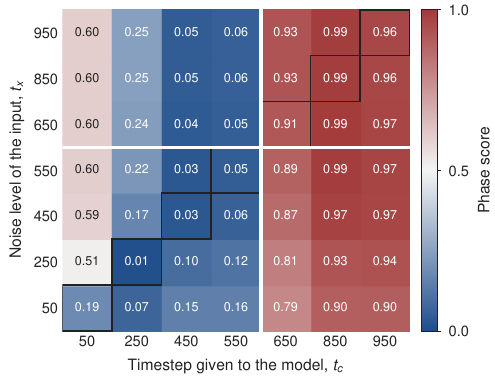}
\caption{Routing under a swapped timestep: the DyLoMo-DiT-XL of \Cref{tab:main} (400k-step EMA weights), the three sparse-expert layers before the loop. The input is noised to $t_x$ while the timestep embedding receives $t_c$. Each cell is the phase score of the expert distribution, averaged over layers and heads: 0 for the routing of the low-noise range ($t\le550$) and 1 for that of the high-noise range ($t\ge650$). Outlined cells are the unswapped diagonal; white lines mark the routing switch.}
\label{fig:f_swap}
\end{figure}

\paragraph{Swapping the timestep.}
Without a timestep input, the router can tell the noise level in two ways: from the timestep, which reaches it through the adaLN modulation, or from the noisy latent itself. We separate the two by noising the input to $t_x$ while giving $t_c$ to the timestep embedding, for $t_x,t_c\in\{50,250,450,550,650,850,950\}$, with the inputs and noise of the protocol above (pairs with $t_x\neq t_c$ never occur in training); on the diagonal $t_x=t_c$, the expert counts equal those of the original analysis exactly. We use the three sparse-expert layers before the loop, whose routing does not depend on the number of passes. For the 24 pairs with $t_x$ and $t_c$ on opposite sides of the switch, the expert distribution is closer to that of $t_c$ than to that of $t_x$ in 95.5\% of the 288 layer--head--pair cases (mean JS divergence 0.017 against 0.207 nats), and for all 12 routing heads on average; $t_c$ accounts for 97\% of the variance of the phase score in \cref{fig:f_swap} (the JS divergence to the mean diagonal distribution of the low-noise range, divided by the sum of the divergences to those of the low- and high-noise ranges), $t_x$ for 1\%. The routing thus follows the timestep the model is given, not the noise level of its input: the noise-level specialization is read from the timestep conditioning. The exception is $t_c=50$ with a noisier input, where the routing moves partway toward the high-noise experts (phase score about 0.6).

\paragraph{A model without the loop.}
The switch is not induced by the loop either. The XL model with 32 experts per head trained with a single pass (16 layers, no loop; \Cref{tab:b_scale}) also switches within one step: on a grid of step 1 over $t=500$ to 624, the JS divergence between adjacent timesteps peaks at $563\rightarrow564$ (0.329 nats), the second-largest value is 0.008, and all 14 sparse-expert layers and all 56 layer--head pairs peak at this pair. The model of the same structure trained with two passes peaks at $558\rightarrow559$ (0.327).

%% file: supp/G_loop_count.tex
\section{Pass Schedules, Position of the Looped Segment, and a Video Model}
\label{sec:app_g}

This appendix gives what \cref{sec:allocation} refers to (the three-band and reverse allocations, the six inference schedules of the model trained with mixed pass counts, and the cost of running a fixed-schedule model under another schedule), the ablation of the position of the looped segment cited in \cref{sec:loop}, and the return on passes in a video model cited in \cref{sec:alloc}.

\begin{table}[t]
\caption{Pass schedules, each trained separately and run under its own schedule (ImageNet-256, 400k steps, FID-50K, guidance scale 1.5). Columns give the passes of the looped segment: 2 at every timestep; 3\,$|$\,1, three passes for $t<500$ and one for $t\ge500$ (the allocation of the main text); 3\,$|$\,2\,$|$\,1, three passes for $t<400$, two for $400\le t<600$ and one for $t\ge600$; 1\,$|$\,3, the reverse of 3\,$|$\,1. The schedules of a row share the architecture and the mean compute; the last row is the single-head model of \Cref{tab:ditmoe}.}
\label{tab:g_alloc}
\centering\footnotesize\setlength{\tabcolsep}{4pt}
\begin{tabular}{@{}lrrrrr@{}}
\toprule
Model & 2 & 3\,$|$\,1 & vs.\ 2 & 3\,$|$\,2\,$|$\,1 & 1\,$|$\,3 \\
\midrule
\input{supp/gen_G_alloc.tex} 
\bottomrule
\end{tabular}
\end{table}

\paragraph{Three bands and the reverse allocation.}
A three-band allocation (three passes at low noise, two in the middle, one at high noise) is level with the two-band allocation of the main text (B 18.93, L 6.33, XL 4.39; \Cref{tab:g_alloc}), and the reverse allocation (one pass at low noise, three at high noise) is 33\% worse than two passes at every timestep in the B tier (26.14). The per-timestep comparison of \cref{fig:noise}a uses the same 10k validation images and the same noise for both models (bars: paired SEM); just above the boundary the error of the allocated model is higher by 0.5\% and 0.3\% at the first two grid timesteps: the extra pass pays only at low noise, and the pass saved at high noise costs almost nothing.

\begin{figure*}[t]
\centering
\includegraphics[width=\textwidth]{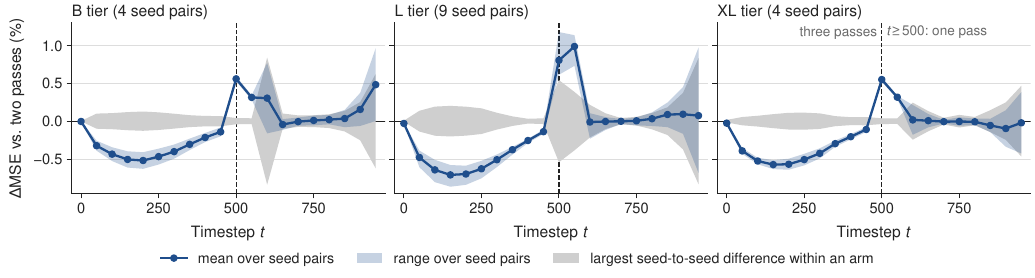}
\caption{Per-timestep loss of allocated passes relative to two passes at every $t$, in the three tiers and over training seeds (protocol of \cref{fig:noise}a: 400k-step EMA weights, the same 10k validation images and noise for every checkpoint). Each seed of the allocated model is paired with each seed of the two-pass model; the grey band is the largest difference between two seeds of the same model. Values are in \% of the two-pass model's error at that timestep.}
\label{fig:g_tloss}
\end{figure*}

\paragraph{The loss difference across tiers and seeds.}
\Cref{fig:g_tloss} repeats the comparison of \cref{fig:noise}a in the three tiers and over the available training seeds of both models. At every grid timestep from $t=50$ to $450$, the allocated model has the lower error in all seed pairs of all three tiers. Over the range $t<500$, its error is lower by 0.28\%, 0.40\% and 0.33\% in the B, L and XL tiers, while two seeds of the same model differ by at most 0.07\%, 0.11\% and 0.04\%. Over $t\ge500$, where it runs one pass instead of two, its error is higher by 0.35\%, 0.58\% and 0.31\%, and over all timesteps it is lower by 0.27\%, 0.38\% and 0.31\% (error averaged over the grid timesteps of the range, relative to the two-pass model). With two or three seeds per model, the bands are ranges, not confidence intervals.

\begin{figure}[t]
\centering
\includegraphics[width=\linewidth]{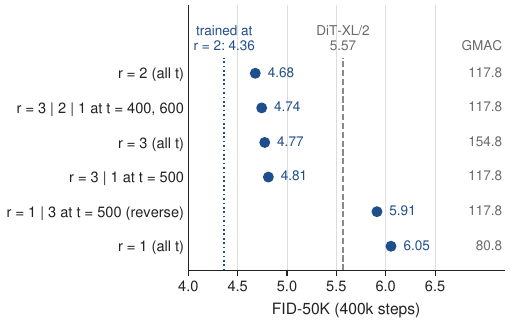}
\caption{One DyLoMo-DiT-XL trained with the pass count drawn from $\{1,2,3\}$ at every step, run under six inference schedules (ImageNet-256, 400k steps; right-hand column: mean compute over the sampling steps). The schedules that place the extra passes at low noise are within 3\% of one another; the reverse allocation and a single pass are 25--30\% worse.}
\label{fig:g_sched}
\end{figure}

\paragraph{Choosing the depth at inference.}
One DyLoMo-DiT-XL is trained with the pass count drawn from $\{1,2,3\}$ at every optimization step (one value for the whole batch) and then run under six schedules (\cref{fig:g_sched}): two passes at every timestep 4.68, three bands 4.74, three passes at every timestep 4.77, two bands 4.81, the reverse 5.91, one pass at every timestep 6.05. The four schedules that place the passes at low noise are within 3\% of one another and interchangeable; three passes at every timestep spend 31\% more compute for no gain; the reverse is as poor as a single pass.

\paragraph{A fixed-schedule model must be run under its schedule.}
Training and inference pass counts must match for a model trained at one schedule: run with one pass at every timestep, the allocated-pass weights of the single-head B model of \Cref{tab:ditmoe} give an FID of 257 (S 213, L 248), and with two passes at every timestep 57.6; changing the depth at inference requires training with mixed pass counts.

\paragraph{Relation to step-wise adaptive computation.}
Prior step-wise adaptive computation in diffusion models acts at inference: AdaDiff~\cite{tang2023adadiff} exits early at some timesteps, DyDiT~\cite{zhao2024dydit} changes the width per timestep, and DDSM~\cite{yang2023ddsm} selects a network size per step. The allocation of passes here is part of the model and of its training, and total parameters and mean compute stay equal to those of the baseline.

\paragraph{Sampling steps against passes at equal compute.}
A looped model spends its sampling compute on passes; a single-pass model can spend the same compute on more sampling steps. To separate the two, the XL models trained with one to four passes at every timestep (\Cref{tab:b_scale}) are each sampled with their own pass count and with the number of DDPM steps set so that the total compute of sampling, the number of steps times the compute of one forward pass, is equal across pass counts (\Cref{tab:g_steps}): equal to two passes at 250 steps (29.4 TMAC per image) and to four passes at 250 steps (48.0 TMAC per image), within 0.4\%. The outcome is the same at both budgets: at equal sampling compute, the compute is better spent on passes than on steps. At 29.4 TMAC, four passes at 153 steps give 3.91 against 4.36 for two passes at 250 steps and 5.85 for one pass at 365; at 48.0 TMAC, four passes at 250 steps give 3.96, three passes at 310 steps 4.33, two at 407 steps 4.41 and one at 594 steps 5.89. The number of steps is saturated in this range: at a fixed pass count, FID moves by at most 0.10 between 150 and 600 steps and without a monotone trend (one pass: 5.94, 5.85, 5.88 and 5.89 at 250, 365, 500 and 594 steps), so a single-pass model does not close the gap by taking more steps. All points are single seeds from the 400k checkpoints of \Cref{tab:b_scale}; differences below 0.1 are within noise. Guidance doubles the forward passes of every point alike and is left out of the counts.

\begin{table}[t]
\caption{Sampling steps against passes at equal sampling compute (DyLoMo-DiT-XL trained with one to four passes at every timestep, 400k steps, 50k images, guidance scale 1.5, single seeds). Each model is sampled with its own pass count; the number of DDPM steps is set so that steps $\times$ GMAC per forward pass is equal within 0.4\%: 29.4 TMAC per image (two passes at 250 steps) and 48.0 TMAC (four passes at 250 steps). Last column: 250 steps at every pass count (\Cref{tab:b_scale}).}
\label{tab:g_steps}
\centering\footnotesize\setlength{\tabcolsep}{3.6pt}
\begin{tabular}{@{}lrrrrrr@{}}
\toprule
 & & \multicolumn{2}{c}{29.4 TMAC} & \multicolumn{2}{c}{48.0 TMAC} & 250 steps \\
\cmidrule(lr){3-4}\cmidrule(lr){5-6}
Passes & \hd{GMAC per\\forward} & Steps & FID & Steps & FID & FID \\
\midrule
\input{supp/gen_G_steps.tex} 
\bottomrule
\end{tabular}
\end{table}

\paragraph{Fewer sampling steps.}
The comparison with DiT does not depend on the 250 sampling steps of the protocol (\Cref{tab:g_few}). With the DDPM sampler respaced to 100 and 50 steps and all other settings unchanged, the first seeds of DyLoMo-DiT-XL and DiT-XL/2 (\Cref{tab:h_seeds}) reach 4.68 against 6.33 and 5.85 against 8.23, which is 26\% and 29\% lower, against 22\% for the same two checkpoints at 250 steps (4.34 against 5.59). The respaced timesteps fall on the two sides of the allocation boundary in equal numbers, so DyLoMo-DiT-XL still runs two passes on average and the compute of the two models stays equal.

\begin{table}[t]
\caption{Fewer sampling steps (first seed of each model, 400k steps, EMA weights, guidance scale 1.5, FID-50K; the DDPM sampler is respaced to the given number of steps).}
\label{tab:g_few}
\centering\footnotesize\setlength{\tabcolsep}{5pt}
\begin{tabular}{@{}rrrr@{}}
\toprule
Steps & DiT-XL/2 & DyLoMo-DiT-XL & vs.\ DiT \\
\midrule
50 & 8.23 & \textbf{5.85} & $-$29\% \\
100 & 6.33 & \textbf{4.68} & $-$26\% \\
250 & 5.59 & \textbf{4.34} & $-$22\% \\
\bottomrule
\end{tabular}
\end{table}

\begin{table*}[t]
\caption{Position of the looped segment, measured on the XL variant with 32 experts per head and two passes at every timestep (setting and columns of \Cref{tab:b_dit}). ``All layers'' keeps 12 physical layers and loops all of them, including the two layers with dense feed-forward networks.}
\label{tab:g_pos}
\centering\footnotesize\setlength{\tabcolsep}{3.4pt}
\begin{tabular}{@{}llllrrrrrrrr@{}}
\toprule
 & & & & \multicolumn{3}{c}{Cost} & \multicolumn{4}{c}{FID $\downarrow$ at step} & \\
\cmidrule(lr){5-7}\cmidrule(lr){8-11}
Looped segment & \hd{Layers\\(looped)} & Experts & Passes & \hd{Total\\(M)} & \hd{Active\\(M)} & GMAC & 100k & 200k & 300k & 400k & Seeds \\
\midrule
\input{supp/gen_G_pos.tex} 
\bottomrule
\end{tabular}
\end{table*}

\paragraph{Position of the looped segment.}
The method places the looped segment in the middle of the backbone (layers 5--11 at the XL tier). On the XL variant with 32 experts per head and two passes at every timestep, we additionally train three other placements: the looped segment at the front (layers 2--8), at the back (layers 9--15), and a model that keeps only 12 physical layers and loops all of them. The first two only move the looped segment and leave the architecture unchanged. The middle is best: 4.21, against 4.63 at the front, 4.70 at the back and 4.72 with all layers looped (\Cref{tab:g_pos}).

\begin{table}[t]
\caption{Validation loss of one set of weights of an in-house video generation model evaluated with different numbers of passes (lower is better). Each additional pass executes the 12 layers of the looped segment once more.}
\label{tab:g_video}
\centering\footnotesize\setlength{\tabcolsep}{4pt}
\begin{tabular}{lcccc}
\toprule
Passes & 1 & 2 & 3 & 4 \\
\midrule
Layers executed & 24 & 36 & 48 & 60 \\
Validation loss & 0.24563 & 0.23371 & 0.23009 & 0.22904 \\
Decrease from previous & --- & 0.01192 & 0.00362 & 0.00105 \\
\bottomrule
\end{tabular}
\end{table}

\paragraph{Return on passes in a video model.}
The return on extra passes at fixed parameters (\cref{sec:loop_role}) is also observed on a separate in-house video generation model: image-to-video, 24 layers, 64 experts per head, with the multi-stream residual of \cref{sec:app_i}, 2.87B parameters in total, looping layers 7 to 18. At every training step the number of passes is drawn uniformly at random from $\{1,2,3,4\}$, for 10B tokens in total. The same weights are then evaluated with one to four passes by the validation loss on 2,000 held-out video clips (the flow-matching~\cite{lipman2022flow} mean squared error, with fixed noise, so the four settings are comparable per sample; \Cref{tab:g_video}). With each additional pass the gain falls to about three tenths of the previous one: the third pass executes 33\% more layers and lowers the validation loss by 0.0036; the fourth adds another 25\% and lowers it by 0.0011.

Two points bear on reading \Cref{tab:g_video}. During training each of the four settings received only a quarter of the steps, and the single pass is diluted the most, so the decrease at the second pass is inflated (when the same architecture is trained with the pass count drawn only from $\{1,2\}$, this term is 0.0074); the increments at the third and fourth passes are diluted about equally on both sides of the comparison and can be compared with each other. The differences in the table are obtained by spending more compute on the same weights and are not a comparison at equal compute; the ImageNet models trained separately at a fixed number of passes (\cref{fig:loop}b, \cref{fig:b_passes}) show the same diminishing return. In addition, LOOM~\cite{he2026loom}, with a separate router and residual scaling per pass, still gains from pushing the loop beyond five passes in language models, so the diminishing return at the third and fourth passes may depend on the recipe and need not be a limit of looping itself.

%% file: supp/gen_G_alloc.tex
B tier, DiT budget & 19.65 & 18.82 & $-4.2\%$ & 18.93 & 26.14 \\
L tier, DiT budget & 6.73 & 6.41 & $-4.7\%$ & 6.33 & --- \\
XL tier, DiT budget & 4.36 & 4.22 & $-3.4\%$ & 4.39 & --- \\
B tier, DiT-MoE budget & 7.94 & 7.01 & $-11.7\%$ & --- & --- \\

%% file: supp/gen_G_steps.tex
1 & 80.8 & 365 & 5.85 & 594 & 5.89 & 5.94 \\
2 & 117.8 & 250 & 4.36 & 407 & 4.41 & 4.36 \\
3 & 154.8 & 190 & 4.35 & 310 & 4.33 & 4.29 \\
4 & 191.8 & 153 & 3.91 & 250 & 3.96 & 3.96 \\

%% file: supp/gen_G_pos.tex
Middle (layers 5--11) & 16 (7) & 4$\times$32 (6) & 2 & 1127.2 & 411.5 & 117.9 & 25.93 & 9.90 & 5.76 & 4.21 & 3 \\
Front (layers 2--8) & 16 (7) & 4$\times$32 (6) & 2 & 1127.2 & 411.5 & 117.9 & 28.45 & 11.19 & 6.48 & 4.63 & 1 \\
Back (layers 9--15) & 16 (7) & 4$\times$32 (6) & 2 & 1127.2 & 411.5 & 117.9 & 27.40 & 11.06 & 6.48 & 4.70 & 1 \\
All layers (12 layers, all looped) & 12 (12) & 4$\times$32 (6) & 2 & 817.1 & --- & 119.2 & 29.27 & 11.66 & 6.55 & 4.72 & 1 \\

%% file: supp/H_arch_details.tex
\section{Architecture and Protocol Details}
\label{sec:app_h}

\begin{figure*}[t]
  \centering
  \includegraphics[width=\textwidth]{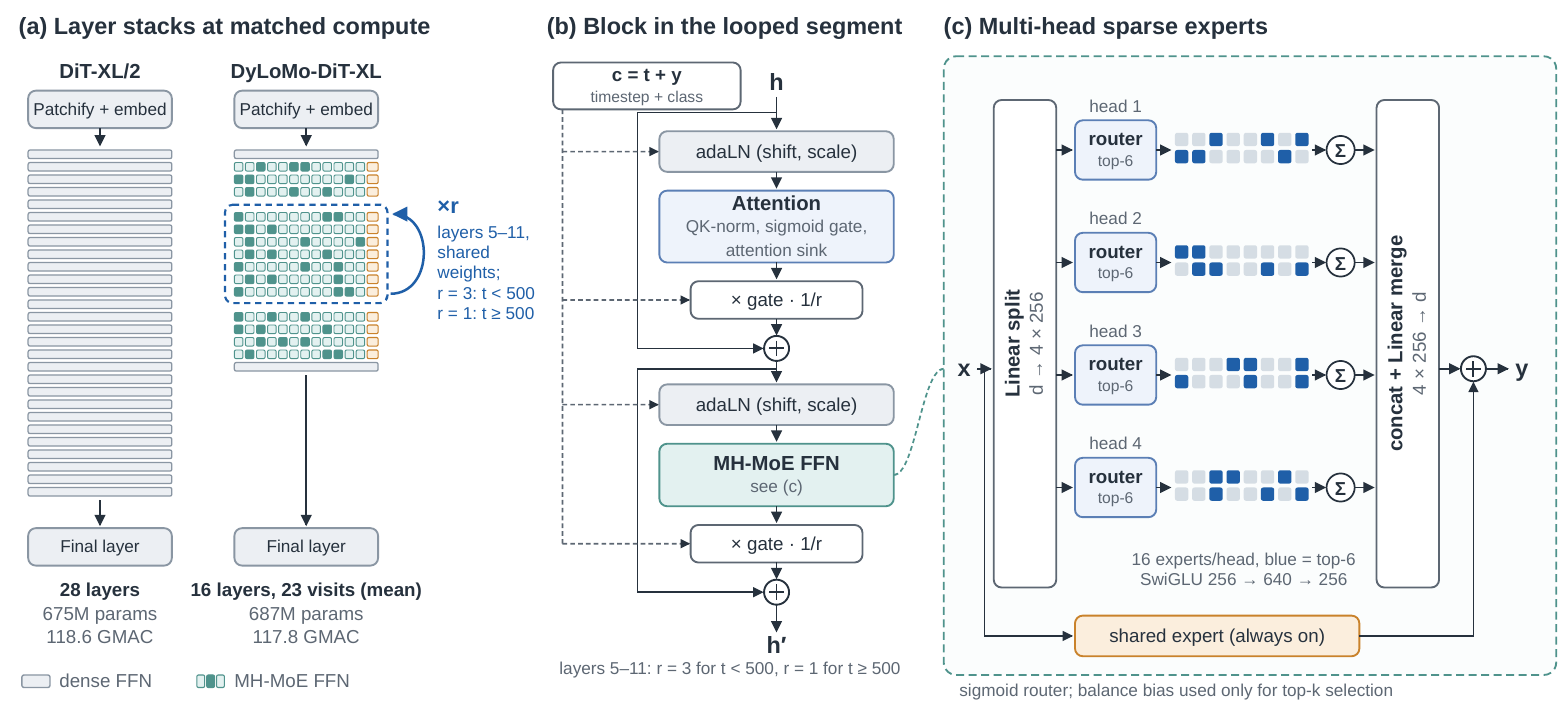}
  \caption{\cref{fig:arch} instantiated at the XL tier, with all values. (a) DiT-XL/2: 28 layers, 0.68B parameters, 118.6 GMAC; DyLoMo-DiT-XL: 16 physical layers, layers 5--11 sharing weights and executed three times for $t<500$ and once for $t\ge500$ (23 layer visits on average), 0.69B parameters, 117.8 GMAC. (c) The input is split into 4 heads of 256 dimensions; the router of each head selects 6 of its own 16 experts; each expert is a SwiGLU $256\to640\to256$.}
  \label{fig:arch_detail}
\end{figure*}

This appendix completes \cref{sec:method} and \cref{sec:setup}: the architecture with all values (\cref{fig:arch_detail}), the configurations of all tiers, the remaining details of the layers (position embedding, normalization, initialization, routing weights, activation), the training and evaluation protocol, the agreement of the DiT baselines with the DiT paper, and the number of seeds behind every number.

\begin{table*}[t]
  \centering
  \footnotesize
  \caption{Configurations at the budget of DiT (\Cref{tab:main}) and the DiT references. Layer indices count from 1. ``Layer visits'': at low noise ($t<500$, three passes) / at high noise ($t\ge500$, one pass) / mean over the sampling steps. ``Single-pass member'': the ``same recipe, no loop'' row of \Cref{tab:main}, which differs from the row above it only in the entries listed.}
  \label{tab:h_cfg}
  \setlength{\tabcolsep}{4pt}
  \begin{tabular}{@{}p{3.3cm} p{2.45cm} p{2.45cm} p{2.45cm} p{1.45cm} p{1.45cm} p{1.45cm}@{}}
    \toprule
     & DyLoMo-DiT-XL & DyLoMo-DiT-L & DyLoMo-DiT-B & DiT-XL/2 & DiT-L/2 & DiT-B/2 \\
    \midrule
    Hidden width $d$ & 1024 & 1024 & 512 & 1152 & 1024 & 768 \\
    Physical layers $L$ & 16 & 12 & 12 & 28 & 24 & 12 \\
    Looped segment & layers 5--11 & layers 4--9 & layers 4--9 & --- & --- & --- \\
    Layer visits & 30 / 16 / 23 & 24 / 12 / 18 & 24 / 12 / 18 & 28 & 24 & 12 \\
    Attention heads $\times$ head dimension & $8\times128$ & $8\times128$ & $4\times128$ & $16\times72$ & $16\times64$ & $12\times64$ \\
    Normalization & RMSNorm & RMSNorm & RMSNorm & LayerNorm & LayerNorm & LayerNorm \\
    Feed-forward & first and last layers dense (width 2688), the rest multi-head sparse experts & same as XL & first and last layers dense (width 1280), the rest multi-head sparse experts & dense & dense & dense \\
    Sparse experts: $H\times E$ ($k$) & $4\times16$ (6) & $4\times17$ (6) & $2\times17$ (6) & --- & --- & --- \\
    Width $f$ of experts and shared expert & 640 & 512 & 312 & --- & --- & --- \\
    Single-pass member & 23 layers, $4\times8$ (6) & 18 layers, $4\times17$ (14), $f=224$ & 18 layers, $2\times8$ (6) & --- & --- & --- \\
    Total parameters (M) & 686.6 & 451.3 & 128.5 & 675.1 & 458.1 & 130.5 \\
    \bottomrule
  \end{tabular}
\end{table*}

\begin{table}[t]
  \centering
  \footnotesize
  \caption{Single-head configurations at the budget of DiT-MoE (\Cref{tab:ditmoe}): 16 physical layers with the middle 6 looped (layers 6--11), one routing head with 16 experts per layer and 2 selected, three passes for $t<500$ and one for $t\ge500$ (28 / 16 / 22 layer visits).}
  \label{tab:h_cfg_dm}
  \setlength{\tabcolsep}{5pt}
  \begin{tabular}{@{}lrrr@{}}
    \toprule
     & S & B & L \\
    \midrule
    Hidden width $d$ & 384 & 768 & 1024 \\
    Total parameters (M) & 199.6 & 796.2 & 2821.8 \\
    Active parameters (M) & 56.9 & 229.0 & 712.0 \\
    Compute (GMAC) & 16.1 & 62.9 & 222.5 \\
    \bottomrule
  \end{tabular}
\end{table}

\paragraph{Configurations.}
\Cref{tab:h_cfg} lists the configurations at the budget of DiT and \Cref{tab:h_cfg_dm} the single-head configurations at the budget of DiT-MoE; the layout of every other arm is given with its results in \cref{sec:app_b}. The middle segment is about half of the layers: at the XL tier $L=16$ with layers 5 to 11 looped; at the B and L tiers $L=12$ with layers 4 to 9 looped, so that a token passes through 24 layers at low noise and 12 at high noise. At the XL tier $H=4$, $E=16$ and $k=6$, the intermediate width of both the experts and the shared expert is 640, and each token activates 24 routed experts per layer. In the single-head B model the experts have intermediate width 1256 and the shared expert 760; the three-head model of \Cref{tab:path} keeps the expert width and has a shared expert of width 776.

\paragraph{Input, conditioning and normalization.}
The input is the VAE latent, split into $2\times2$ patches and linearly projected into tokens; the timestep embedding and the class embedding are summed into a conditioning vector $c$, which every sublayer injects through adaLN-Zero; positional information is supplied by rotary position embeddings~\cite{su2021roformer} inside attention; the output layer predicts the noise and the variance. All normalization in the network is RMSNorm, applied once each to $q$ and $k$; the DiT baselines keep the official implementation. The rotary position encoding is a two-dimensional Fourier RoPE with 16 frequency bands, rotating only the first $3/4$ of the dimensions of each attention head; the sigmoid gate of each attention head reads the normalized input. Normalization is computed in FP32 with eps $=10^{-6}$.

\paragraph{Initialization.}
Router weights are drawn from a normal distribution with standard deviation 0.02. The experts' up-projection has standard deviation $256^{-1/2}$ and the down-projection $f^{-1/2}$. The remaining linear layers have standard deviation equal to the input dimension to the power $-1/2$, none with bias; the exceptions are the adaLN modulation layers (with bias, zero-initialized) and the patch embedding layer (PyTorch default initialization, with bias). The load-balancing biases start at zero. The RMSNorm $w$, the attention sink and the output layer are also zero-initialized. The timestep and class embeddings use the PyTorch default initialization.

\paragraph{Routing weights.}
The constant $\alpha=4.9$ that multiplies the normalized routing weights in \cref{sec:mhmoe} is the default of the implementation and was not tuned.

\paragraph{Activation function.}
The experts, the shared expert and the dense feed-forward networks of the first and last layers all use a clamped SwiGLU: the gate branch $g$ is clipped to at most 7, the other branch $u$ is clipped to $[-7, 7]$, and the output is $g\cdot\sigma(1.702\cdot g)\cdot(u+1)$. This SwiGLU~\cite{shazeer2020glu} variant with clipping and a residual term appears in gpt-oss~\cite{openai2025gptoss}.

\paragraph{Protocol details.}
Images are encoded with sd-vae-ft-ema~\cite{rombach2022ldm} with scaling factor 0.18215 into $32\times32\times4$ latents. Training uses a global batch of 256 for 400k steps, AdamW~\cite{loshchilov2019adamw} with $\beta=(0.9, 0.999)$, a constant learning rate of $10^{-4}$, no weight decay, and an EMA coefficient of 0.9999. DDPM~\cite{ho2020ddpm} uses a 1000-step linear noise schedule, predicting the noise and learning the variance~\cite{nichol2021improved}; evaluation follows \cref{sec:setup}, using ADM's~\cite{dhariwal2021adm} TensorFlow evaluator and reference statistics.

\begin{table}[t]
  \centering
  \footnotesize
  \caption{FID-50K without guidance (400k steps, EMA weights, 250 DDPM steps): the DiT baselines against the published values under the same training setting, and DyLoMo-DiT. Table~4 of the DiT paper~\cite{peebles2023dit} uses the ft-MSE decoder; all evaluations in this paper use the ft-EMA decoder. Seeds without guidance: DiT three (XL), two (L), one (B); DyLoMo-DiT one. The rows at guidance scale 1.5 repeat \Cref{tab:main}.}
  \label{tab:h_nocfg}
  \setlength{\tabcolsep}{3.5pt}
  \begin{tabular}{@{}lrrr@{}}
    \toprule
     & XL & L & B \\
    \midrule
    DiT, no guidance & 18.91 & 22.73 & 43.32 \\
    \ \ DiT paper, Table~4 & 19.47 & 23.33 & 43.47 \\
    \ \ official PyTorch repository & 18.1 & --- & --- \\
    DyLoMo-DiT, no guidance & 14.22 & 18.10 & 38.47 \\
    \ \ vs.\ DiT & $-$25\% & $-$20\% & $-$11\% \\
    \midrule
    DiT, guidance scale 1.5 & 5.57 & 7.42 & 21.19 \\
    DyLoMo-DiT, guidance scale 1.5 & 4.22 & 6.41 & 18.82 \\
    \ \ vs.\ DiT & $-$24\% & $-$14\% & $-$11\% \\
    \bottomrule
  \end{tabular}
\end{table}

\paragraph{Agreement of the DiT baselines with the DiT paper.}
Without guidance, the FID of our DiT-XL/2, DiT-L/2 and DiT-B/2 baselines at 400k steps is within 3\% of the values that Table~4 of the DiT paper reports under the same training setting (\Cref{tab:h_nocfg}).

\begin{table}[t]
  \centering
  \footnotesize
  \caption{Models trained with more than one seed: FID-50K at 400k steps per seed and the mean used in the paper. Every other model has one seed (column ``Seeds'' of \Cref{tab:b_dit,tab:b_scale,tab:b_dm}). $^\ddagger$See text.}
  \label{tab:h_seeds}
  \setlength{\tabcolsep}{3pt}
  \begin{tabular}{@{}p{3.6cm}rlr@{}}
    \toprule
    Model & Seeds & FID per seed & Mean \\
    \midrule
    \input{supp/gen_H_seeds.tex} 
    \midrule
    ProMoE-L (\cref{sec:app_j}) & 3 & 5.94, 6.98, 5.91 & 6.28 \\
    DyLoMo-DiT at the budget of ProMoE-L, four heads & 3 & 5.15, 4.88, 5.29 & 5.11 \\
    \ \ same, one head & 2 & 5.60, 5.79 & 5.70 \\
    \bottomrule
  \end{tabular}
\end{table}

\paragraph{Seeds.}
All numbers in the paper are means over the available seeds (\cref{sec:setup}); for a model with one seed, this is the value of that seed. \Cref{tab:h_seeds} lists the models with more than one seed and their per-seed values; all other models, including DiT-B/2 and the DyLoMo-DiT rows of \Cref{tab:ditmoe}, have one seed. $^\ddagger$The three DiT-L/2 runs were trained from two snapshots of the training code (seed 42 from the earlier one, seeds 1 and 2 from the later one). The spread of the three DiT-XL/2 seeds is below 2\%, so the 14\% spread of DiT-L/2 should not be read as the typical seed-to-seed variation.

%% file: supp/gen_H_seeds.tex
DiT-XL/2 & 3 & 5.59, 5.50, 5.60 & 5.57 \\
DiT-L/2$^\ddagger$ & 3 & 6.88, 7.85, 7.51 & 7.42 \\
DyLoMo-DiT-XL & 2 & 4.34, 4.10 & 4.22 \\
DyLoMo-DiT-L & 2 & 6.25, 6.57 & 6.41 \\
DyLoMo-DiT-B & 2 & 19.16, 18.48 & 18.82 \\
\ \ two passes at every $t$ (L) & 3 & 6.91, 6.42, 6.85 & 6.73 \\
\ \ two passes at every $t$ (B) & 2 & 20.10, 19.20 & 19.65 \\
DyLoMo-DiT-XL, 32 experts per head, two passes & 3 & 4.14, 4.24, 4.26 & 4.21 \\
\ \ same, one pass & 2 & 4.97, 6.02 & 5.49 \\
DyLoMo-DiT-L, 32 experts per head, two passes & 2 & 5.58, 6.17 & 5.88 \\
Dense feed-forward with a loop, two passes (XL) & 2 & 5.37, 5.83 & 5.60 \\
\ \ same, one pass & 2 & 7.42, 7.64 & 7.53 \\

%% file: supp/I_ohc_full.tex
\section{Full Results for the Multi-Stream Residual}
\label{sec:app_i}

This appendix gives the evaluation of the multi-stream residual that \cref{sec:related} and \cref{sec:setup} refer to: on ImageNet it is on par when training completes normally but makes training unstable together with adaLN, so it is not part of the method; without adaLN it trains normally and brings a gain, so the text-to-image model keeps it.

\paragraph{Multi-stream residual.}
Hyper-Connections~\cite{zhu2024hyper} widen the residual stream into several streams and learn the connections between them; its appendix reports on DiT-XL/2 that static HC (SHC$\times$2) improves FID from 2.36 to 2.18 (a baseline with FP16 and QK normalization, 1400 epochs, guidance scale 1.5). mHC~\cite{xie2025mhc} constrains the mixing matrix between streams to be doubly stochastic; oHC~\cite{guo2026ohc} constrains it to the rotation group and parameterizes it with a pair of unit quaternions; other models fix it to the identity (iHC), such as the language model Hy4-preview~\cite{hy4preview} and Chimera~\cite{ge2026chimera}. The experiments of mHC and oHC are only on language models. In diffusion models, MAGI-2~\cite{magi2} uses mHC; Chimera uses iHC together with sparse experts and reports that both improve the compute efficiency of pretraining (measured by the compute needed to reach the same training loss); Liang~\etal~\cite{liang2026residual} compare single-stream and four-stream mHC on ImageNet-256 and find that the gain depends on the prediction target: at the B tier in pixel space, FID goes from 10.19 to 9.78 with $x$-prediction and from 139.83 to 25.36 with $v$-prediction. We evaluate oHC separately as an optional component on top of DyLoMo-DiT, with the results below.

The DyLoMo-DiT of this paper has no multi-stream residual. We additionally trained versions with oHC, in which the residual stream is widened to 4 streams mixed by an $\mathrm{SO}(4)$ rotation, with the training and evaluation protocol unchanged and 1.2\% more compute. These experiments were run on the XL family with 32 experts per head and two passes at every timestep (1.13B parameters; \cref{sec:app_b}) and on its ablations. When training completes normally, the version with oHC is on par with the version without: the final FID is 4.20, against 4.21 without. Under this recipe, however, training with oHC is unstable: of the 9 runs across the arms with oHC, 7 have FID rising rather than falling at later evaluation points (back up to tens, at most over three hundred); in the 11 runs of these arms without oHC (9 at the XL tier, 2 at the L tier), FID decreases monotonically with training steps. In these abnormal runs the model itself degrades in the same way: the quality of individual images drops while diversity does not. Evaluating the abnormal checkpoints directly with the raw (non-EMA) weights gives FID from 23.96 to 46.24, Precision from 0.43 to 0.58 and Recall from 0.62 to 0.69; the normally completed checkpoints give 7.03 to 10.50, 0.70 to 0.74 and 0.55 to 0.59 respectively. In 2 of these runs (300.19 and 255.67) the EMA weights collapse further on top of this, with FID in the hundreds and Recall close to 0. The phenomenon also appears in the arms with dense feed-forward and no routing, so it does not depend on the routing of the sparse experts. We therefore did not include oHC in the method proper.

\begin{table}[t]
\caption{The family with the multi-stream residual (oHC), EMA FID at 400k steps at the XL tier. Several numbers in one cell are different runs; $\dagger$ marks a run whose FID rose rather than fell at later evaluation points. All rows have 32 experts per head. The two runs with oHC at the L tier behave the same way: the sparse-expert model 255.67$^\dagger$ (5.88 without oHC), the dense feed-forward 10.59$^\dagger$.}
\label{tab:i1}
\centering\footnotesize\setlength{\tabcolsep}{3pt}
\begin{tabular}{@{}p{2.95cm} p{1.95cm} p{2.9cm}@{}}
\toprule
Model & Without oHC & With oHC \\
\midrule
Sparse experts, two passes & 4.14; 4.24; 4.26 & 4.20; 31.66$^\dagger$; 15.33$^\dagger$ \\
Sparse experts, one pass & 4.97; 6.02 & 5.30 \\
Dense feed-forward, two passes & 5.37; 5.83 & 8.02$^\dagger$; 37.29$^\dagger$ \\
Dense feed-forward, one pass & 7.42; 7.64 & 300.19$^\dagger$ \\
\bottomrule
\end{tabular}
\end{table}

\begin{table}[t]
\caption{Total parameters and forward compute per pass of the models with oHC (in parentheses, the difference relative to the DiT of the same tier). The corresponding values without oHC are in \Cref{tab:b_scale,tab:b_lb32}; oHC adds only the parameters and compute of the mixing between streams (0.7--2.4\% of the parameters and about 1\% of the compute).}
\label{tab:i2}
\centering\footnotesize\setlength{\tabcolsep}{4pt}
\begin{tabular}{lrl}
\toprule
Model & \hd{Total\\parameters (M)} & GMAC \\
\midrule
\tierrow{3}{XL tier}
Sparse experts, two passes & 1,137 & 119.31 ($+$0.58\%) \\
Sparse experts, one pass & 1,137 & 81.82 ($-$31.0\%) \\
Dense feed-forward, two passes & 422 & 119.48 ($+$0.73\%) \\
Dense feed-forward, one pass & 422 & 81.94 ($-$30.9\%) \\
\tierrow{3}{L tier}
Sparse experts, two passes & 696 & 81.26 ($+$0.70\%) \\
Dense feed-forward, two passes & 288 & 81.39 ($+$0.86\%) \\
\bottomrule
\end{tabular}
\end{table}

Taking the run with the lowest FID in each cell, the comparisons in the family with oHC go in the same direction as without: adding the loop goes from 5.30 to 4.20, and sparse experts against dense feed-forward is 4.20 vs.\ 8.02.

In our experiments the instability appears only when oHC is used together with adaLN. With the optimizer recipe unchanged (no weight decay, no gradient clipping, no learning-rate decay) and only the conditioning changed to a single addition at the input instead of adaLN, all 7 runs with oHC train without the instability (the no-loop model among them was trained only to 200k, with both evaluation points still decreasing), FID decreases monotonically in each, and oHC helps: two runs of the full model reach FID 5.01 and 5.17 at 400k, below the 5.67 without oHC. The text-to-image model likewise does not use adaLN and therefore includes this component; its optimizer additionally uses weight decay and gradient clipping, and the phenomenon did not occur during its training.

%% file: supp/J_ditmoe.tex
\section{Sparse-Expert Baselines}
\label{sec:app_j}

This appendix documents the DiT-MoE baseline of \cref{sec:setup} and \cref{sec:vs_ditmoe,sec:scale_baselines}: its two configurations (both in the S, B and L tiers; the original one also at XL), how the implementation was checked against the released code, how our training differs from the recipe of the DiT-MoE paper, and what two independent works report for the same kind of routing.

\begin{table*}[t]
\caption{The two DiT-MoE configurations under our protocol (ImageNet-256, 400k steps, EMA weights, guidance scale 1.5, 50k images) and the DyLoMo-DiT configured for the same total parameters and compute (\Cref{tab:ditmoe}). Both configurations keep the width, depth and attention heads of the DiT of the same name and have one always-on shared expert; within a tier they have the same expert parameters and differ only in the size of the router. ``---'': not trained.}
\label{tab:j_cfg}
\centering\footnotesize\setlength{\tabcolsep}{5pt}
\begin{tabular}{@{}llrrrrrr@{}}
\toprule
Tier & DiT-MoE configuration & \hd{Total\\(M)} & \hd{Active\\(M)} & GMAC & FID $\downarrow$ & \hd{DyLoMo-DiT\\FID $\downarrow$} & \hd{vs.\\DiT-MoE} \\
\midrule
\input{supp/gen_J_cfg.tex} 
\bottomrule
\end{tabular}
\end{table*}

\paragraph{Implementation and the two configurations.}
DiT-MoE is implemented from the code released by its authors~\cite{fei2024ditmoe}. A model keeps the width, depth and attention heads of the DiT of the same name; the feed-forward network of every layer is replaced by routed experts plus one shared expert, with SwiGLU experts. The router takes a softmax over the experts, selects those with the highest probability and weights them by the probabilities without renormalization; the shared expert is computed for every token with weight 1. Load balancing uses an expert-level auxiliary loss with coefficient 0.005, the value in the text of the DiT-MoE paper (the default of the released code is 0.01), with statistics computed locally in each data-parallel process. The training protocol, optimizer, EMA, sampling and evaluation are those of all other models (\cref{sec:setup}). We train two configurations (\Cref{tab:j_cfg}). The original configuration has 8 experts per layer with 2 selected, each of intermediate width $4d$, and a shared expert of width $2d$ (the released code implements the two shared experts of the paper as one feed-forward network of this width). The second configuration cuts every expert in two: 16 experts per layer of width $2d$ with 4 selected and the same shared expert, so that total parameters, active parameters and compute are unchanged up to the router; it has the number of experts per layer of the DyLoMo-DiT it is compared with in \Cref{tab:ditmoe}. At the B tier, the original configuration (12.91) is 20\% below the configuration with 16 experts (16.06), and DyLoMo-DiT (7.01) is 46\% and 56\% below the two. At the S tier the two configurations are level: the original one gives 30.80 and the one with 16 experts 30.83, and DyLoMo-DiT (19.79) is 36\% below both, so the advantage of the original configuration at the B tier does not carry over to the S tier. At the L tier the original configuration is at 34.23, 16.38, 10.33 and 7.50 after 100k, 200k, 300k and 400k steps, against 35.64, 17.27, 10.94 and 7.92 for the configuration with 16 experts, so it is 5\% below at 400k, and DyLoMo-DiT (3.50) is 53\% and 56\% below the two. Over the three tiers the original configuration is therefore level with or below the one with 16 experts, and DyLoMo-DiT is 36\%, 46\% and 53\% below the stronger of the two at S, B and L.

\paragraph{Configurations at other budgets.}
The DiT-MoE models of \cref{sec:scale_baselines} use the same code (costs and FID over training: \Cref{tab:b_scale,tab:b_dit}). Scaled down to the budget of DiT-XL/2, DiT-MoE has width 1024 and 3 experts per layer with 2 selected (expert width 536, shared expert 2768; 7.90). At the compute of DiT-XL/2 with more parameters, it has width 1152 and 8 experts with 2 selected, either with experts of width 768 and a shared expert as wide as the two selected experts together (1536; $1.7\times$ the parameters, 9.19) or with the proportions of the released code, in which the shared expert is half as wide as a routed expert (1232 and 616; $2.1\times$, 10.61). The model scaled to $2.2\times$ in both parameters and compute has width 1536 (expert width 808, shared expert 4136; 6.65), and the original XL configuration has experts of width 4608 and a shared expert of width 2304 (4.17B parameters, 328.0 GMAC; 6.87). Without guidance, the two eight-expert models at the compute of DiT-XL/2 reach 26.89 and 29.24 at 400k steps, against 18.91 for DiT-XL/2 (\Cref{tab:h_nocfg}).

\paragraph{Differences from the recipe of the DiT-MoE paper.}
The DiT-MoE paper trains with a batch size of 1024 for 500k, 1M and 7M iterations and reports, with guidance and after long training, an FID of 1.72 for its XL configuration against 2.27 for DiT-XL/2. Our runs use the protocol of all other models in this paper (batch 256, 400k steps, DDPM, 250 sampling steps, guidance scale 1.5). The DiT-MoE numbers of this paper therefore characterize its recipe under this protocol and say nothing about the results in its paper. Training every model under one protocol is also what ProMoE~\cite{wei2025promoe} and DiffMoE~\cite{shi2025diffmoe} do when they compare with this kind of routing; both in addition reconfigure the sparse-expert baselines to the active parameters of the dense model, whereas \cref{sec:vs_ditmoe} keeps the width, depth, routing, shared expert and auxiliary loss of DiT-MoE and configures our model for its budget.

\paragraph{Verification against the released code.}
The checks were made on the two eight-expert models at the compute of DiT-XL/2. Given the same inputs and weights, our implementation and the released code agree bitwise in BF16 on the forward output of the training branch and on the gradients of all parameters; both expert widths were checked (the check at expert width 1232 used a shared expert of width 608; the shared expert is an ordinary dense feed-forward network and does not go through the expert kernel). The remaining checks were done on the model with experts of width 768: a full-model FP32 forward pass on CPU with the 100k-step EMA weights gives a relative $L_2$ error of $2.4\times10^{-5}$ to $3.5\times10^{-4}$, and the gradient injected by the auxiliary loss agrees bitwise with explicit differentiation (CPU, FP32). A control trained with the original implementation of the released code (without kernel fusion) reaches FID 36.14 at 100k steps, comparable to the 37.20 of our implementation. The ordering of the training loss across evaluation points agrees with that of FID. FID evaluated with the raw (non-EMA) weights at 100k steps is 41.36, worse than the 37.20 with EMA weights, so the gap to DiT does not come from a mismatch between EMA weights and routing. The routing of the two models stays close to uniform over training: the normalized entropy, averaged over layers and positions, is between 0.96 and 0.98 at every checkpoint, and no expert is ever unused; the share of selections that goes to one expert (12.5\% when uniform) ranges from 7\% to 18\% in the model with experts of width 1232 and from 1.3\% to 29.4\% at 400k steps in the model with experts of width 768.

\paragraph{Independent evidence.}
Two independent works that retrain this kind of routing next to a dense DiT under one protocol find the same direction as \cref{sec:vs_ditmoe,sec:scale_baselines}. ProMoE (Table~10 of~\cite{wei2025promoe}; DDPM, L tier, 500k steps) reports FID 23.12 without guidance and 7.55 at guidance scale 1.5 for its DiT-MoE-L, which has 8 experts with 1 selected and no shared expert, against 20.81 and 6.29 for the dense DiT-L (its Table~4). DiffMoE (Table~11 of~\cite{shi2025diffmoe}; DDPM, L tier, 700k steps) reports 20.81 for token-choice routing with 16 experts against 17.87 for the dense DiT-L. In both, token-choice sparse experts are worse than the dense model at equal active parameters.

\paragraph{A controlled comparison with ProMoE.}
DiT-MoE is the baseline of the main text because the later models start from its design: ProMoE~\cite{wei2025promoe}, DiffMoE and EC-DiT keep a DiT backbone with a single pass over distinct layers and change how tokens are routed to the experts, a different axis from the budgeting of depth, parameters and compute studied here. We also compare with ProMoE at equal total parameters and compute. Its model code is ported from the authors' release and checked against it: with the same inputs and weights, the forward pass and all gradients are bit-identical. The sampling and evaluation pipeline is checked with their released ProMoE-L/2 checkpoint (500k steps, rectified flow): sampled with 250 Euler steps, guidance 1.5, the ft-MSE VAE and 50k images, it gives FID 2.82, sFID 4.62 and IS 241.6 against 2.79, 4.58 and 244.2 reported by the authors, and our evaluator run on their released 50k samples gives 2.7925, identical to their log. We train the expert-choice variant, which the authors recommend for DDPM training, in the L configuration of their paper: the DiT-L/2 backbone (24 layers, width 1024) with a sparse feed-forward on every other layer, each with 12 routed experts selected by prototype routing, one shared and one unconditional expert, expert width 2048; 1.063B parameters in total, 0.458B active, 80.6 GMAC without the one-hot products that index tokens in its routing (82.2 with them). Two DyLoMo-DiT models are configured for this budget; they differ only in the number of routing heads. Both have 16 layers with the middle six looped, 24 experts per head with 2 selected, expert width 784 and shared width 880, and passes allocated by noise level as in \cref{tab:main}; one has a single routing head, the recipe of \cref{tab:ditmoe}, and the other four heads of width 256, the head layout of the models of \cref{tab:main} (1.066B parameters, $+0.3\%$, and 80.6 GMAC, $-0.02\%$, for both; the heads split the hidden width, so the number of heads changes neither). All three are trained under our protocol (DDPM, batch 256, the ft-EMA VAE, no gradient clipping), which differs from theirs in the objective, the VAE and the clipping, so the numbers are not comparable to the rectified-flow numbers of their paper. The first seed of ProMoE-L reaches 30.07, 12.49, 7.89, 5.94 and 4.86 after 100k, 200k, 300k, 400k and 500k steps (400k as in our other comparisons, 500k as in the DDPM results of their paper). At 400k steps it is below the original configuration of DiT-MoE-L (7.50, \Cref{tab:j_cfg}) with 38\% of its parameters and 36\% of its compute. Its gain over the dense model is in line with their paper: at 400k steps under our protocol, this seed is 19.8\% below DiT-L/2 (5.94 against 7.42) and the mean of three seeds 15\% below it (6.28), and their DDPM comparison at 500k steps reports 18.6\% (5.12 against 6.29, their Table~4). With the released checkpoint, this indicates that the port is faithful and that our protocol does not put ProMoE at a disadvantage. At the same five steps, the first seed of the four-head DyLoMo-DiT reaches 30.77, 12.03, 7.11, 5.15 and 4.05 and that of the single-head one 26.53, 11.84, 7.51, 5.60 and 4.56: they are 13\% and 6\% below ProMoE-L at 400k steps and 17\% and 6\% at 500k steps. At 400k steps the three models are repeated with further seeds (\Cref{tab:h_seeds}): ProMoE-L reaches 5.94, 6.98 and 5.91 (mean 6.28), the four-head DyLoMo-DiT 5.15, 4.88 and 5.29 (mean 5.11) and the single-head one 5.60 and 5.79 (mean 5.70). In the seed means, the four-head model is 19\% and the single-head model 9\% below ProMoE-L, and four heads are 10\% below one head; the seeds of ProMoE-L span 1.06 and those of the four-head model 0.42. We also trained ProMoE-XL, the XL configuration of their paper in the same variant ($2.32\times$ the parameters of DiT-XL/2; its compute within 0.2\% of DiT-XL/2 without the one-hot products and 1.6\% above it with them), for 400k steps under our protocol. It reaches 4.78 with one seed, 14\% below DiT-XL/2 (5.57). The DyLoMo-DiT-XL of \Cref{tab:main}, with 44\% of the parameters of ProMoE-XL at the same compute, reaches 4.22, 12\% below it.

%% file: supp/gen_J_cfg.tex
S & 8 experts, 2 selected (original) & 199.3 & 71.9 & 16.0 & 30.80 & 19.79 & $-$36\% \\
S & 16 experts, 4 selected & 199.3 & 71.9 & 16.0 & 30.83 & 19.79 & $-$36\% \\
\midrule
B & 8 experts, 2 selected (original) & 795.9 & 286.3 & 62.9 & 12.91 & 7.01 & $-$46\% \\
B & 16 experts, 4 selected & 795.9 & 286.3 & 62.9 & 16.06 & 7.01 & $-$56\% \\
\midrule
L & 8 experts, 2 selected (original) & 2823.8 & 1011.8 & 222.5 & 7.50 & 3.50 & $-$53\% \\
L & 16 experts, 4 selected & 2824.0 & 1012.0 & 222.5 & 7.92 & 3.50 & $-$56\% \\
\midrule
XL & 8 experts, 2 selected (original) & 4168.2 & 1492.7 & 328.0 & 6.87 & --- & --- \\
XL & 16 experts, 4 selected & --- & --- & --- & --- & --- & --- \\

%% file: supp/K_flow.tex
\section{Flow-Matching Objective}
\label{sec:app_k}

The ImageNet experiments of the main paper use the DDPM objective of the DiT protocol (\cref{sec:setup}). To test whether the comparison with DiT depends on this objective, we retrain the L-tier and XL-tier models of \Cref{tab:main}, DiT-L/2 and DiT-XL/2 and the DyLoMo-DiT of each tier, with a flow-matching objective~\cite{lipman2022flow,liu2022rectified,ma2024sit}: the input is the linear interpolation $x_\sigma=(1-\sigma)\cdot x_0+\sigma\cdot\epsilon$ with $\sigma$ drawn uniformly from $[0,1]$, the network predicts the velocity $\epsilon-x_0$, and the time embedding receives $1000\,\sigma$. The architectures, their parameters and compute, and all other training settings are those of the DDPM runs. DyLoMo-DiT keeps its pass schedule with the boundary at $\sigma=0.5$: three passes for $\sigma<0.5$ and one otherwise, two passes on average. All models are sampled from the EMA weights with 250 Euler steps uniformly spaced in $\sigma$ and evaluated as in \cref{sec:setup}.

\begin{table}[t]
\caption{Flow-matching objective in the L and XL tiers (ImageNet-256, 400k steps, EMA weights, 250 Euler steps, FID-50K). The L-tier entries are means over two seeds; the XL tier has one seed. Parameters and compute are those of \Cref{tab:main}.}
\label{tab:k_flow}
\centering\footnotesize\setlength{\tabcolsep}{4pt}
\begin{tabular}{@{}lrrrr@{}}
\toprule
 & \hd{Total\\params (M)} & GMAC & \multicolumn{2}{c}{FID $\downarrow$} \\
\cmidrule(l){4-5}
Model & & & \hd{guidance\\1.5} & \hd{no\\guidance} \\
\midrule
DiT-L/2 & 458.1 & 80.7 & 6.66 & 20.57 \\
DyLoMo-DiT-L & 451.3 & 80.1 & \textbf{5.95} & \textbf{17.61} \\
\ \ vs.\ DiT & & & $-$11\% & $-$14\% \\
\midrule
DiT-XL/2 & 675.1 & 118.6 & 5.27 & 17.39 \\
DyLoMo-DiT-XL & 686.6 & 117.8 & \textbf{4.51} & \textbf{14.40} \\
\ \ vs.\ DiT & & & $-$15\% & $-$17\% \\
\bottomrule
\end{tabular}
\end{table}

At guidance scale 1.5, DyLoMo-DiT is 11\% below DiT in the L tier and 15\% below it in the XL tier; without guidance it is 14\% and 17\% below (\Cref{tab:k_flow}). Under the DDPM objective the gaps are 14\% and 24\% with guidance and 20\% and 25\% without (\Cref{tab:main}, \cref{sec:app_h}). The L-tier entries are means over two seeds. At guidance scale 1.5, the two DiT-L/2 runs reach 6.75 and 6.57 and the two DyLoMo-DiT-L runs 5.90 and 6.00: both DyLoMo-DiT-L runs are below both DiT-L/2 runs, and the two seeds of a model differ by at most 0.19, against 0.71 between the means of the two models. The XL tier has one seed. The gain therefore carries over to a flow-matching objective; with this few seeds, its direction is better established than its size. The two objectives are sampled differently, so FID values are compared only within an objective.

%% file: main.bbl
\begin{thebibliography}{82}
\providecommand{\natexlab}[1]{#1}
\providecommand{\url}[1]{\texttt{#1}}
\expandafter\ifx\csname urlstyle\endcsname\relax
  \providecommand{\doi}[1]{doi: #1}\else
  \providecommand{\doi}{doi: \begingroup \urlstyle{rm}\Url}\fi

\bibitem[Akiti et~al.(2026)Akiti, Modukuri, Nagarapu, Akiti, and
  Liu]{akiti2026nucleus}
Chandan Akiti, Ajay Modukuri, Murali~Nandan Nagarapu, Gunavardhan Akiti, and
  Haozhe Liu.
\newblock {Nucleus-Image: Sparse MoE for Image Generation}.
\newblock \emph{arXiv preprint arXiv:2604.12163}, 2026.

\bibitem[Bae et~al.(2025)Bae, Kim, Bayat, Kim, Ha, Schuster, Fisch,
  Harutyunyan, Ji, Courville, and Yun]{bae2025mor}
Sangmin Bae, Yujin Kim, Reza Bayat, Sungnyun Kim, Jiyoun Ha, Tal Schuster, Adam
  Fisch, Hrayr Harutyunyan, Ziwei Ji, Aaron Courville, and Se-Young Yun.
\newblock {Mixture-of-Recursions: Learning Dynamic Recursive Depths for
  Adaptive Token-Level Computation}.
\newblock In \emph{Advances in Neural Information Processing Systems
  (NeurIPS)}, 2025.

\bibitem[Bai et~al.(2025)Bai, Chen, Liu, Wang, Ge, Song, Dang, Wang, Wang,
  Tang, Zhong, Zhu, Yang, Li, Wan, Wang, Ding, Fu, Xu, Ye, Zhang, Xie, Cheng,
  Zhang, Yang, Xu, and Lin]{bai2025qwen25vl}
Shuai Bai, Keqin Chen, Xuejing Liu, Jialin Wang, Wenbin Ge, Sibo Song, Kai
  Dang, Peng Wang, Shijie Wang, Jun Tang, Humen Zhong, Yuanzhi Zhu, Mingkun
  Yang, Zhaohai Li, Jianqiang Wan, Pengfei Wang, Wei Ding, Zheren Fu, Yiheng
  Xu, Jiabo Ye, Xi Zhang, Tianbao Xie, Zesen Cheng, Hang Zhang, Zhibo Yang,
  Haiyang Xu, and Junyang Lin.
\newblock {Qwen2.5-VL Technical Report}.
\newblock \emph{arXiv preprint arXiv:2502.13923}, 2025.

\bibitem[Bai and Melas-Kyriazi(2024)]{bai2024fpdm}
Xingjian Bai and Luke Melas-Kyriazi.
\newblock {Fixed Point Diffusion Models}.
\newblock In \emph{Proceedings of the IEEE/CVF Conference on Computer Vision
  and Pattern Recognition (CVPR)}, 2024.

\bibitem[Balaji et~al.(2022)Balaji, Nah, Huang, Vahdat, Song, Zhang, Kreis,
  Aittala, Aila, Laine, Catanzaro, Karras, and Liu]{balaji2022ediffi}
Yogesh Balaji, Seungjun Nah, Xun Huang, Arash Vahdat, Jiaming Song, Qinsheng
  Zhang, Karsten Kreis, Miika Aittala, Timo Aila, Samuli Laine, Bryan
  Catanzaro, Tero Karras, and Ming-Yu Liu.
\newblock {eDiff-I: Text-to-Image Diffusion Models with an Ensemble of Expert
  Denoisers}.
\newblock \emph{arXiv preprint arXiv:2211.01324}, 2022.

\bibitem[{Black Forest Labs}(2024)]{flux2024}
{Black Forest Labs}.
\newblock {{FLUX}}.
\newblock GitHub repository, black-forest-labs/flux, 2024.

\bibitem[Chen et~al.(2024)Chen, Yu, Ge, Yao, Xie, Wang, Kwok, Luo, Lu, and
  Li]{chen2023pixartalpha}
Junsong Chen, Jincheng Yu, Chongjian Ge, Lewei Yao, Enze Xie, Zhongdao Wang,
  James Kwok, Ping Luo, Huchuan Lu, and Zhenguo Li.
\newblock {PixArt-$\alpha$: Fast Training of Diffusion Transformer for
  Photorealistic Text-to-Image Synthesis}.
\newblock In \emph{International Conference on Learning Representations
  (ICLR)}, 2024.

\bibitem[Chen et~al.(2026{\natexlab{a}})Chen, Li, Huang, Yin, Shang, and
  Qin]{chen2026loopmoe}
Wenkai Chen, Tianshu Li, Wenyong Huang, Yichun Yin, Lifeng Shang, and Chengwei
  Qin.
\newblock {LoopMoE: Unifying Iterative Computation with Mixture-of-Experts for
  Language Modeling}.
\newblock In \emph{Proceedings of the 2026 Conference on Empirical Methods in
  Natural Language Processing (EMNLP)}, 2026{\natexlab{a}}.

\bibitem[Chen et~al.(2026{\natexlab{b}})Chen, Goyal, and
  Krishnamoorthi]{chen2026scalinglooped}
Yanbei Chen, Anirudh Goyal, and Raghuraman Krishnamoorthi.
\newblock {Scaling Laws for Looped Mixture of Experts}.
\newblock \emph{arXiv preprint arXiv:2609.40316}, 2026{\natexlab{b}}.

\bibitem[Cheng et~al.(2022)Cheng, Misra, Schwing, Kirillov, and
  Girdhar]{cheng2022mask2former}
Bowen Cheng, Ishan Misra, Alexander~G. Schwing, Alexander Kirillov, and Rohit
  Girdhar.
\newblock {Masked-Attention Mask Transformer for Universal Image Segmentation}.
\newblock In \emph{Proceedings of the IEEE/CVF Conference on Computer Vision
  and Pattern Recognition (CVPR)}, 2022.

\bibitem[Cheng et~al.(2025)Cheng, He, Yu, Tu, Zhu, Wang, Gao, and
  Hu]{diffmoe_timestep}
Kun Cheng, Xiao He, Lei Yu, Zhijun Tu, Mingrui Zhu, Nannan Wang, Xinbo Gao, and
  Jie Hu.
\newblock {Diff-MoE: Diffusion Transformer with Time-Aware and Space-Adaptive
  Experts}.
\newblock In \emph{International Conference on Machine Learning (ICML)}, 2025.

\bibitem[Chng et~al.(2026)Chng, Chen, Tong, Diao, Cai, Yang, Liu, Lu, Lin, and
  Huang]{chng2026loopeddit}
Yong~Xien Chng, Tianyi Chen, Wenwen Tong, Haiwen Diao, Zhongang Cai, Lei Yang,
  Ziwei Liu, Lewei Lu, Dahua Lin, and Gao Huang.
\newblock {Looped Diffusion Transformer}.
\newblock \emph{arXiv preprint arXiv:2609.40305}, 2026.

\bibitem[Csord{\'a}s et~al.(2024)Csord{\'a}s, Irie, Schmidhuber, Potts, and
  Manning]{csordas2024moeut}
R{\'o}bert Csord{\'a}s, Kazuki Irie, J{\"u}rgen Schmidhuber, Christopher Potts,
  and Christopher~D. Manning.
\newblock {MoEUT: Mixture-of-Experts Universal Transformers}.
\newblock In \emph{Advances in Neural Information Processing Systems
  (NeurIPS)}, 2024.

\bibitem[Cui et~al.(2026)Cui, Jackson, Herrera, T{\'a}rano, and
  Kerner]{cui2026mhlatentmoe}
Chenwei Cui, Rockwell Jackson, Benjamin~Joseph Herrera, Ana~Mar{\'i}a
  T{\'a}rano, and Hannah Kerner.
\newblock {Multi-Head LatentMoE and Head Parallel: Communication-Efficient and
  Deterministic MoE Parallelism}.
\newblock \emph{arXiv preprint arXiv:2602.04870}, 2026.

\bibitem[Dai et~al.(2024)Dai, Deng, Zhao, Xu, Gao, Chen, Li, Zeng, Yu, Wu, Xie,
  Li, Huang, Luo, Ruan, Sui, and Liang]{dai2024deepseekmoe}
Damai Dai, Chengqi Deng, Chenggang Zhao, R.~X. Xu, Huazuo Gao, Deli Chen,
  Jiashi Li, Wangding Zeng, Xingkai Yu, Y. Wu, Zhenda Xie, Y.~K. Li, Panpan
  Huang, Fuli Luo, Chong Ruan, Zhifang Sui, and Wenfeng Liang.
\newblock {DeepSeekMoE: Towards Ultimate Expert Specialization in
  Mixture-of-Experts Language Models}.
\newblock In \emph{Proceedings of the 62nd Annual Meeting of the Association
  for Computational Linguistics (ACL)}, 2024.

\bibitem[{DeepSeek-AI}(2024)]{deepseek2024v3}
{DeepSeek-AI}.
\newblock {DeepSeek-V3 Technical Report}.
\newblock \emph{arXiv preprint arXiv:2412.19437}, 2024.

\bibitem[Dehghani et~al.(2019)Dehghani, Gouws, Vinyals, Uszkoreit, and
  Kaiser]{dehghani2019universal}
Mostafa Dehghani, Stephan Gouws, Oriol Vinyals, Jakob Uszkoreit, and {\L}ukasz
  Kaiser.
\newblock {Universal Transformers}.
\newblock In \emph{International Conference on Learning Representations
  (ICLR)}, 2019.

\bibitem[Deng et~al.(2009)Deng, Dong, Socher, Li, Li, and
  Fei-Fei]{deng2009imagenet}
Jia Deng, Wei Dong, Richard Socher, Li-Jia Li, Kai Li, and Li Fei-Fei.
\newblock {{ImageNet}: A Large-Scale Hierarchical Image Database}.
\newblock In \emph{Proceedings of the IEEE Conference on Computer Vision and
  Pattern Recognition (CVPR)}, 2009.

\bibitem[Derakhshani et~al.(2026)Derakhshani, Curvo, Burghouts, van~de Meent,
  and Snoek]{derakhshani2026lift}
Mohammad~Mahdi Derakhshani, Pedro M.~P. Curvo, Gertjan~J. Burghouts, Jan-Willem
  van~de Meent, and Cees G.~M. Snoek.
\newblock {LiFT: Loop Flow Transformers}.
\newblock \emph{arXiv preprint arXiv:2610.05538}, 2026.

\bibitem[Dhariwal and Nichol(2021)]{dhariwal2021adm}
Prafulla Dhariwal and Alex Nichol.
\newblock {Diffusion Models Beat GANs on Image Synthesis}.
\newblock In \emph{Advances in Neural Information Processing Systems
  (NeurIPS)}, 2021.

\bibitem[Esser et~al.(2024)Esser, Kulal, Blattmann, Entezari, M{\"u}ller,
  Saini, Levi, Lorenz, Sauer, Boesel, Podell, Dockhorn, English, and
  Rombach]{esser2024sd3}
Patrick Esser, Sumith Kulal, Andreas Blattmann, Rahim Entezari, Jonas
  M{\"u}ller, Harry Saini, Yam Levi, Dominik Lorenz, Axel Sauer, Frederic
  Boesel, Dustin Podell, Tim Dockhorn, Zion English, and Robin Rombach.
\newblock {Scaling Rectified Flow Transformers for High-Resolution Image
  Synthesis}.
\newblock In \emph{International Conference on Machine Learning (ICML)}, 2024.

\bibitem[Fang et~al.(2024)Fang, Ma, and Wang]{fang2024remixdit}
Gongfan Fang, Xinyin Ma, and Xinchao Wang.
\newblock {Remix-DiT: Mixing Diffusion Transformers for Multi-Expert
  Denoising}.
\newblock In \emph{Advances in Neural Information Processing Systems
  (NeurIPS)}, 2024.

\bibitem[Fei et~al.(2024)Fei, Fan, Yu, Li, and Huang]{fei2024ditmoe}
Zhengcong Fei, Mingyuan Fan, Changqian Yu, Debang Li, and Junshi Huang.
\newblock {Scaling Diffusion Transformers to 16 Billion Parameters}.
\newblock \emph{arXiv preprint arXiv:2407.11633}, 2024.

\bibitem[Feng et~al.(2023)Feng, Zhang, Yu, Fang, Li, Chen, Lu, Liu, Yin, Feng,
  Sun, Chen, Tian, Wu, and Wang]{feng2023ernievilg}
Zhida Feng, Zhenyu Zhang, Xintong Yu, Yewei Fang, Lanxin Li, Xuyi Chen, Yuxiang
  Lu, Jiaxiang Liu, Weichong Yin, Shikun Feng, Yu Sun, Li Chen, Hao Tian, Hua
  Wu, and Haifeng Wang.
\newblock {{ERNIE-ViLG} 2.0: Improving Text-to-Image Diffusion Model with
  Knowledge-Enhanced Mixture-of-Denoising-Experts}.
\newblock In \emph{Proceedings of the IEEE/CVF Conference on Computer Vision
  and Pattern Recognition (CVPR)}, 2023.

\bibitem[Gao et~al.(2026)Gao, Chen, Xiao, Yang, Tao, Zhou, and
  Dai]{gao2026loopie}
Zitian Gao, Yilong Chen, Yihao Xiao, Xinyu Yang, Ran Tao, Joey Zhou, and Bryan
  Dai.
\newblock {Loop the Loopies!}
\newblock \emph{arXiv preprint arXiv:2607.16051}, 2026.

\bibitem[Ge et~al.(2026)Ge, Jiang, Wang, Gu, Xu, Chen, Liu, Shi, Hong, Cai,
  Jin, and Tan]{ge2026chimera}
Chongjian Ge, Hanwen Jiang, Tianyu Wang, Jiuxiang Gu, Yiran Xu, Ziwen Chen,
  Shaoteng Liu, Jing Shi, Yicong Hong, Zefan Cai, Hailin Jin, and Hao Tan.
\newblock {Chimera: Designing and Chinchilla-Scaling Hybrid Visual Diffusion
  Transformers}.
\newblock \emph{arXiv preprint arXiv:2607.28611}, 2026.

\bibitem[Ghosh et~al.(2023)Ghosh, Hajishirzi, and Schmidt]{ghosh2023geneval}
Dhruba Ghosh, Hanna Hajishirzi, and Ludwig Schmidt.
\newblock {GenEval: An Object-Focused Framework for Evaluating Text-to-Image
  Alignment}.
\newblock In \emph{Advances in Neural Information Processing Systems (NeurIPS),
  Datasets and Benchmarks Track}, 2023.

\bibitem[Goyal et~al.(2026)Goyal, Agrawal, Anil, Jain, Paul, and
  Kusupati]{goyal2026elt}
Sahil Goyal, Swayam Agrawal, Gautham~Govind Anil, Prateek Jain, Sujoy Paul, and
  Aditya Kusupati.
\newblock {ELT: Elastic Looped Transformers for Visual Generation}.
\newblock In \emph{European Conference on Computer Vision (ECCV)}, 2026.

\bibitem[Guo et~al.(2026)Guo, Chen, Ke, Lei, Xu, Feng, {Ximen}, and
  Luo]{guo2026ohc}
Haoqiang Guo, Xuyi Chen, Bo Ke, Yishu Lei, Ziyang Xu, Shikun Feng, {Ximen}, and
  Wenhan Luo.
\newblock {oHC: Orthogonal Hyper-Connections on SO(4) via Quaternions}.
\newblock \emph{arXiv preprint arXiv:2609.02672}, 2026.

\bibitem[He et~al.(2026{\natexlab{a}})He, Li, Chang, Meng, Yin, and
  Liu]{he2026loom}
Di He, Pengxiang Li, Da Chang, Qingyan Meng, Lu Yin, and Shiwei Liu.
\newblock {Looping Beyond Twice: A Scalable Recipe for Looped
  Mixture-of-Experts}.
\newblock \emph{arXiv preprint arXiv:2610.01153}, 2026{\natexlab{a}}.

\bibitem[He et~al.(2026{\natexlab{b}})He, Chai, Chen, Zhu, Wu, Luo, Li, and
  Cheng]{he2026latentloop}
Zefeng He, Yunkai Chai, Xiaoyang Chen, Tong Zhu, Zongliang Wu, Yun Luo, Yafu
  Li, and Yu Cheng.
\newblock {LatentLoop: Scaling Looped MoE Models with Asymmetric Compute
  Allocation}.
\newblock \emph{arXiv preprint arXiv:2610.19975}, 2026{\natexlab{b}}.

\bibitem[Ho and Salimans(2022)]{ho2022cfg}
Jonathan Ho and Tim Salimans.
\newblock {Classifier-Free Diffusion Guidance}.
\newblock \emph{arXiv preprint arXiv:2207.12598}, 2022.

\bibitem[Ho et~al.(2020)Ho, Jain, and Abbeel]{ho2020ddpm}
Jonathan Ho, Ajay Jain, and Pieter Abbeel.
\newblock {Denoising Diffusion Probabilistic Models}.
\newblock In \emph{Advances in Neural Information Processing Systems
  (NeurIPS)}, 2020.

\bibitem[Hu et~al.(2024)Hu, Wang, Fang, Fu, Cheng, and Yu]{hu2024ella}
Xiwei Hu, Rui Wang, Yixiao Fang, Bin Fu, Pei Cheng, and Gang Yu.
\newblock {ELLA: Equip Diffusion Models with LLM for Enhanced Semantic
  Alignment}.
\newblock \emph{arXiv preprint arXiv:2403.05135}, 2024.

\bibitem[Jaggi(2026)]{jaggi2026tying}
Martin Jaggi.
\newblock {Tying the Loop -- Tied Expert Layers in Mixture-of-Experts Language
  Models}.
\newblock \emph{arXiv preprint arXiv:2606.16825}, 2026.

\bibitem[Jia et~al.(2026)Jia, Wang, Huang, Zhang, Cambria, and Li]{jia2026mmoe}
Yanhao Jia, Jiepeng Wang, Haibin Huang, Chi Zhang, Erik Cambria, and Xuelong
  Li.
\newblock {MMOE: Modernizing Diffusion Transformers with Efficient Expert
  Design}.
\newblock \emph{arXiv preprint arXiv:2607.24665}, 2026.

\bibitem[Kynk{\"a}{\"a}nniemi et~al.(2019)Kynk{\"a}{\"a}nniemi, Karras, Laine,
  Lehtinen, and Aila]{kynkaanniemi2019pr}
Tuomas Kynk{\"a}{\"a}nniemi, Tero Karras, Samuli Laine, Jaakko Lehtinen, and
  Timo Aila.
\newblock {Improved Precision and Recall Metric for Assessing Generative
  Models}.
\newblock In \emph{Advances in Neural Information Processing Systems
  (NeurIPS)}, 2019.

\bibitem[Lee et~al.(2026)Lee, Biloki, Hu, and May]{lee2026sparse}
Ryan Lee, Jacob Biloki, Edward~J. Hu, and Jonathan May.
\newblock {Sparse Layers are Critical to Scaling Looped Language Models}.
\newblock \emph{arXiv preprint arXiv:2605.09165}, 2026.

\bibitem[Li et~al.(2022)Li, Xu, Tian, Wang, Yan, Bi, Ye, Chen, Xu, Cao, Zhang,
  Huang, Huang, Zhou, and Si]{li2022mplug}
Chenliang Li, Haiyang Xu, Junfeng Tian, Wei Wang, Ming Yan, Bin Bi, Jiabo Ye,
  Hehong Chen, Guohai Xu, Zheng Cao, Ji Zhang, Songfang Huang, Fei Huang,
  Jingren Zhou, and Luo Si.
\newblock {mPLUG: Effective and Efficient Vision-Language Learning by
  Cross-modal Skip-connections}.
\newblock In \emph{Proceedings of the 2022 Conference on Empirical Methods in
  Natural Language Processing (EMNLP)}, 2022.

\bibitem[Liang et~al.(2026)Liang, Kou, Xi, Singh, Zhou, Deng, Cloninger, Wang,
  and Parhi]{liang2026residual}
Tongtong Liang, Siqi Kou, Ziqiao Xi, Esha Singh, Kun Zhou, Zhijie Deng,
  Alexander Cloninger, Yu-Xiang Wang, and Rahul Parhi.
\newblock {Residual-Stream Burden Shapes Representation Learning in Diffusion
  Transformers}.
\newblock \emph{arXiv preprint arXiv:2609.33895}, 2026.

\bibitem[Lin et~al.(2014)Lin, Maire, Belongie, Hays, Perona, Ramanan,
  Doll{\'a}r, and Zitnick]{lin2014coco}
Tsung-Yi Lin, Michael Maire, Serge Belongie, James Hays, Pietro Perona, Deva
  Ramanan, Piotr Doll{\'a}r, and C.~Lawrence Zitnick.
\newblock {Microsoft COCO: Common Objects in Context}.
\newblock In \emph{European Conference on Computer Vision (ECCV)}, 2014.

\bibitem[Lipman et~al.(2023)Lipman, Chen, Ben-Hamu, Nickel, and
  Le]{lipman2022flow}
Yaron Lipman, Ricky T.~Q. Chen, Heli Ben-Hamu, Maximilian Nickel, and Matt Le.
\newblock {Flow Matching for Generative Modeling}.
\newblock In \emph{International Conference on Learning Representations
  (ICLR)}, 2023.

\bibitem[Liu et~al.(2023)Liu, Gong, and Liu]{liu2022rectified}
Xingchao Liu, Chengyue Gong, and Qiang Liu.
\newblock {Flow Straight and Fast: Learning to Generate and Transfer Data with
  Rectified Flow}.
\newblock In \emph{International Conference on Learning Representations
  (ICLR)}, 2023.

\bibitem[Loshchilov and Hutter(2019)]{loshchilov2019adamw}
Ilya Loshchilov and Frank Hutter.
\newblock {Decoupled Weight Decay Regularization}.
\newblock In \emph{International Conference on Learning Representations
  (ICLR)}, 2019.

\bibitem[Ma et~al.(2024)Ma, Goldstein, Albergo, Boffi, Vanden-Eijnden, and
  Xie]{ma2024sit}
Nanye Ma, Mark Goldstein, Michael~S. Albergo, Nicholas~M. Boffi, Eric
  Vanden-Eijnden, and Saining Xie.
\newblock {SiT: Exploring Flow and Diffusion-based Generative Models with
  Scalable Interpolant Transformers}.
\newblock In \emph{European Conference on Computer Vision (ECCV)}, 2024.

\bibitem[Nash et~al.(2021)Nash, Menick, Dieleman, and Battaglia]{nash2021sfid}
Charlie Nash, Jacob Menick, Sander Dieleman, and Peter~W. Battaglia.
\newblock {Generating Images with Sparse Representations}.
\newblock In \emph{International Conference on Machine Learning (ICML)}, 2021.

\bibitem[Nichol and Dhariwal(2021)]{nichol2021improved}
Alex Nichol and Prafulla Dhariwal.
\newblock {Improved Denoising Diffusion Probabilistic Models}.
\newblock In \emph{International Conference on Machine Learning (ICML)}, 2021.

\bibitem[{OpenAI}(2025)]{openai2025gptoss}
{OpenAI}.
\newblock {gpt-oss-120b \& gpt-oss-20b Model Card}.
\newblock \emph{arXiv preprint arXiv:2508.10925}, 2025.

\bibitem[Park et~al.(2024{\natexlab{a}})Park, Go, Kim, Woo, Ham, and
  Kim]{park2024switchdit}
Byeongjun Park, Hyojun Go, Jin-Young Kim, Sangmin Woo, Seokil Ham, and Changick
  Kim.
\newblock {Switch Diffusion Transformer: Synergizing Denoising Tasks with
  Sparse Mixture-of-Experts}.
\newblock In \emph{European Conference on Computer Vision (ECCV)},
  2024{\natexlab{a}}.

\bibitem[Park et~al.(2024{\natexlab{b}})Park, Woo, Go, Kim, and
  Kim]{park2024dtr}
Byeongjun Park, Sangmin Woo, Hyojun Go, Jin-Young Kim, and Changick Kim.
\newblock {Denoising Task Routing for Diffusion Models}.
\newblock In \emph{International Conference on Learning Representations
  (ICLR)}, 2024{\natexlab{b}}.

\bibitem[Parmar et~al.(2022)Parmar, Zhang, and Zhu]{parmar2022cleanfid}
Gaurav Parmar, Richard Zhang, and Jun-Yan Zhu.
\newblock {On Aliased Resizing and Surprising Subtleties in GAN Evaluation}.
\newblock In \emph{Proceedings of the IEEE/CVF Conference on Computer Vision
  and Pattern Recognition (CVPR)}, 2022.

\bibitem[Peebles and Xie(2023)]{peebles2023dit}
William Peebles and Saining Xie.
\newblock {Scalable Diffusion Models with Transformers}.
\newblock In \emph{Proceedings of the IEEE/CVF International Conference on
  Computer Vision (ICCV)}, 2023.

\bibitem[Podell et~al.(2024)Podell, English, Lacey, Blattmann, Dockhorn,
  M{\"u}ller, Penna, and Rombach]{podell2023sdxl}
Dustin Podell, Zion English, Kyle Lacey, Andreas Blattmann, Tim Dockhorn, Jonas
  M{\"u}ller, Joe Penna, and Robin Rombach.
\newblock {SDXL: Improving Latent Diffusion Models for High-Resolution Image
  Synthesis}.
\newblock In \emph{International Conference on Learning Representations
  (ICLR)}, 2024.

\bibitem[Qiu et~al.(2026)Qiu, Acikalin, Lovelace, Belardi, Mulchandani, Gomes,
  and Weinberger]{qiu2026more}
Eric~S. Qiu, Utku~Umur Acikalin, Justin Lovelace, Christian Belardi, Arjun~B.
  Mulchandani, Carla~P. Gomes, and Kilian~Q. Weinberger.
\newblock {MoRE: Mixture of Reused Experts}.
\newblock In \emph{Conference on Language Modeling (COLM)}, 2026.

\bibitem[Radford et~al.(2021)Radford, Kim, Hallacy, Ramesh, Goh, Agarwal,
  Sastry, Askell, Mishkin, Clark, Krueger, and Sutskever]{radford2021clip}
Alec Radford, Jong~Wook Kim, Chris Hallacy, Aditya Ramesh, Gabriel Goh,
  Sandhini Agarwal, Girish Sastry, Amanda Askell, Pamela Mishkin, Jack Clark,
  Gretchen Krueger, and Ilya Sutskever.
\newblock {Learning Transferable Visual Models From Natural Language
  Supervision}.
\newblock In \emph{International Conference on Machine Learning (ICML)}, 2021.

\bibitem[Rombach et~al.(2022)Rombach, Blattmann, Lorenz, Esser, and
  Ommer]{rombach2022ldm}
Robin Rombach, Andreas Blattmann, Dominik Lorenz, Patrick Esser, and Bj{\"o}rn
  Ommer.
\newblock {High-Resolution Image Synthesis with Latent Diffusion Models}.
\newblock In \emph{Proceedings of the IEEE/CVF Conference on Computer Vision
  and Pattern Recognition (CVPR)}, 2022.

\bibitem[Sadat et~al.(2025)Sadat, Hilliges, and Weber]{sadat2024apg}
Seyedmorteza Sadat, Otmar Hilliges, and Romann~M. Weber.
\newblock {Eliminating Oversaturation and Artifacts of High Guidance Scales in
  Diffusion Models}.
\newblock In \emph{International Conference on Learning Representations
  (ICLR)}, 2025.

\bibitem[Salimans et~al.(2016)Salimans, Goodfellow, Zaremba, Cheung, Radford,
  and Chen]{salimans2016is}
Tim Salimans, Ian Goodfellow, Wojciech Zaremba, Vicki Cheung, Alec Radford, and
  Xi Chen.
\newblock {Improved Techniques for Training GANs}.
\newblock In \emph{Advances in Neural Information Processing Systems
  (NeurIPS)}, 2016.

\bibitem[{Sand AI}(2026)]{magi2}
{Sand AI}.
\newblock {{MAGI-2} Preview: Scaling Video Generation Models Efficiently}.
\newblock Technical blog, 2026.

\bibitem[Shazeer(2020)]{shazeer2020glu}
Noam Shazeer.
\newblock {GLU Variants Improve Transformer}.
\newblock \emph{arXiv preprint arXiv:2002.05202}, 2020.

\bibitem[Shazeer et~al.(2017)Shazeer, Mirhoseini, Maziarz, Davis, Le, Hinton,
  and Dean]{shazeer2017moe}
Noam Shazeer, Azalia Mirhoseini, Krzysztof Maziarz, Andy Davis, Quoc Le,
  Geoffrey Hinton, and Jeff Dean.
\newblock {Outrageously Large Neural Networks: The Sparsely-Gated
  Mixture-of-Experts Layer}.
\newblock In \emph{International Conference on Learning Representations
  (ICLR)}, 2017.

\bibitem[Shi et~al.(2025)Shi, Yuan, Yang, Wang, Zheng, Tao, Zhao, Zheng, Zhou,
  Lu, Wan, Zhang, and Gai]{shi2025diffmoe}
Minglei Shi, Ziyang Yuan, Haotian Yang, Xintao Wang, Mingwu Zheng, Xin Tao,
  Wenliang Zhao, Wenzhao Zheng, Jie Zhou, Jiwen Lu, Pengfei Wan, Di Zhang, and
  Kun Gai.
\newblock {DiffMoE: Dynamic Token Selection for Scalable Diffusion
  Transformers}.
\newblock \emph{arXiv preprint arXiv:2503.14487}, 2025.

\bibitem[Su et~al.(2024)Su, Ahmed, Lu, Pan, Bo, and Liu]{su2021roformer}
Jianlin Su, Murtadha Ahmed, Yu Lu, Shengfeng Pan, Wen Bo, and Yunfeng Liu.
\newblock {RoFormer: Enhanced Transformer with Rotary Position Embedding}.
\newblock \emph{Neurocomputing}, 568:\penalty0 127063, 2024.

\bibitem[Sun et~al.(2025)Sun, Lei, Zhang, Li, Huang, Pang, Dai, and
  Du]{sun2025ecdit}
Haotian Sun, Tao Lei, Bowen Zhang, Yanghao Li, Haoshuo Huang, Ruoming Pang, Bo
  Dai, and Nan Du.
\newblock {EC-DIT: Scaling Diffusion Transformers with Adaptive Expert-Choice
  Routing}.
\newblock In \emph{International Conference on Learning Representations
  (ICLR)}, 2025.

\bibitem[Tan et~al.(2023)Tan, Shen, Chen, Courville, and Gan]{tan2023sut}
Shawn Tan, Yikang Shen, Zhenfang Chen, Aaron Courville, and Chuang Gan.
\newblock {Sparse Universal Transformer}.
\newblock In \emph{Proceedings of the 2023 Conference on Empirical Methods in
  Natural Language Processing (EMNLP)}, 2023.

\bibitem[Tang et~al.(2024)Tang, Wang, Ding, Liang, Li, and Xu]{tang2023adadiff}
Shengkun Tang, Yaqing Wang, Caiwen Ding, Yi Liang, Yao Li, and Dongkuan Xu.
\newblock {AdaDiff: Accelerating Diffusion Models through Step-Wise Adaptive
  Computation}.
\newblock In \emph{European Conference on Computer Vision (ECCV)}, 2024.

\bibitem[{Tencent Hy Team}(2026)]{hy4preview}
{Tencent Hy Team}.
\newblock {{Hy4-preview} Model Card}.
\newblock Hugging Face model card, tencent/Hy4-preview, 2026.

\bibitem[Wan et~al.(2025)Wan, Wang, Ai, Wen, Mao, Xie, Chen, Yu, Zhao, Yang,
  Zeng, Wang, Zhang, Zhou, Wang, Chen, Zhu, Zhao, Yan, Huang, Feng, Zhang, Li,
  Wu, Chu, Feng, Zhang, Sun, Fang, Wang, Gui, Weng, Shen, Lin, Wang, Wang,
  Zhou, Wang, Shen, Yu, Shi, Huang, Xu, Kou, Lv, Li, Liu, Wang, Zhang, Huang,
  Li, Wu, Liu, Pan, Zheng, Hong, Shi, Feng, Jiang, Han, Wu, and Liu]{wan2025}
Team Wan, Ang Wang, Baole Ai, Bin Wen, Chaojie Mao, Chen-Wei Xie, Di Chen,
  Feiwu Yu, Haiming Zhao, Jianxiao Yang, Jianyuan Zeng, Jiayu Wang, Jingfeng
  Zhang, Jingren Zhou, Jinkai Wang, Jixuan Chen, Kai Zhu, Kang Zhao, Keyu Yan,
  Lianghua Huang, Mengyang Feng, Ningyi Zhang, Pandeng Li, Pingyu Wu, Ruihang
  Chu, Ruili Feng, Shiwei Zhang, Siyang Sun, Tao Fang, Tianxing Wang, Tianyi
  Gui, Tingyu Weng, Tong Shen, Wei Lin, Wei Wang, Wei Wang, Wenmeng Zhou, Wente
  Wang, Wenting Shen, Wenyuan Yu, Xianzhong Shi, Xiaoming Huang, Xin Xu, Yan
  Kou, Yangyu Lv, Yifei Li, Yijing Liu, Yiming Wang, Yingya Zhang, Yitong
  Huang, Yong Li, You Wu, Yu Liu, Yulin Pan, Yun Zheng, Yuntao Hong, Yupeng
  Shi, Yutong Feng, Zeyinzi Jiang, Zhen Han, Zhi-Fan Wu, and Ziyu Liu.
\newblock {Wan: Open and Advanced Large-Scale Video Generative Models}.
\newblock \emph{arXiv preprint arXiv:2503.20314}, 2025.

\bibitem[{Wan Team}(2025)]{wan2025wan22}
{Wan Team}.
\newblock {{Wan2.2}}.
\newblock GitHub repository, Wan-Video/Wan2.2, 2025.

\bibitem[Wang et~al.(2024)Wang, Gao, Zhao, Sun, and Dai]{wang2024auxfree}
Lean Wang, Huazuo Gao, Chenggang Zhao, Xu Sun, and Damai Dai.
\newblock {Auxiliary-Loss-Free Load Balancing Strategy for Mixture-of-Experts}.
\newblock \emph{arXiv preprint arXiv:2408.15664}, 2024.

\bibitem[Wang et~al.(2026{\natexlab{a}})Wang, Li, Zhu, Zhang, Huang, Yan, and
  Li]{wang2026loopmup}
Shaowen Wang, Bingrui Li, Tansheng Zhu, Ge Zhang, Wenhao Huang, Shen Yan, and
  Jian Li.
\newblock {Loop-$\mu$P: Residual Scaling and Hyperparameter Transfer in Looped
  Transformers}.
\newblock \emph{arXiv preprint arXiv:2606.18524}, 2026{\natexlab{a}}.

\bibitem[Wang et~al.(2026{\natexlab{b}})Wang, Zhang, Luo, Wu, Liu, Liu, Huang,
  Yan, and Li]{wang2026smelt}
Shaowen Wang, Ge Zhang, Kairong Luo, Yuhao Wu, Shaofan Liu, Jiaheng Liu, Wenhao
  Huang, Shen Yan, and Jian Li.
\newblock {SMELT: Scaling Laws for Compute-Matched MoE Looped Transformers}.
\newblock \emph{arXiv preprint arXiv:2609.01343}, 2026{\natexlab{b}}.

\bibitem[Wang et~al.(2026{\natexlab{c}})Wang, Xiong, Zhang, Zhu, Jiang, Xiao,
  Zeng, Chen, Liu, Hu, Chen, Wang, Liu, Lian, LV, Zhao, Gao, Huang, Shen, and
  You]{wang2026loopreg}
Zipeng Wang, Anping Xiong, Jing Zhang, Xiaoya Zhu, Yi Jiang, Yanbing Xiao, Zige
  Zeng, Zezhou Chen, Xiang Liu, Huan Hu, Ping Chen, Kai Wang, Zhaoxiang Liu,
  Shiguo Lian, Ye LV, Fang Zhao, Huanlin Gao, Zhengrui Huang, Bo Shen, and
  Yuren You.
\newblock {LoopREG: Representation Supervision Enables Test-Time Recurrent
  Depth in Diffusion Transformers}.
\newblock \emph{arXiv preprint arXiv:2610.17877}, 2026{\natexlab{c}}.

\bibitem[Wei et~al.(2026)Wei, Zhang, Yuan, Han, Chen, Wang, Zou, Liu, Zhang,
  Liu, and Shan]{wei2025promoe}
Yujie Wei, Shiwei Zhang, Hangjie Yuan, Yujin Han, Zhekai Chen, Jiayu Wang,
  Difan Zou, Xihui Liu, Yingya Zhang, Yu Liu, and Hongming Shan.
\newblock {Routing Matters in MoE: Scaling Diffusion Transformers with Explicit
  Routing Guidance}.
\newblock In \emph{International Conference on Learning Representations
  (ICLR)}, 2026.

\bibitem[Wu et~al.(2024)Wu, Huang, Wang, Ma, Dong, and Wei]{wu2024mhmoe}
Xun Wu, Shaohan Huang, Wenhui Wang, Shuming Ma, Li Dong, and Furu Wei.
\newblock {Multi-Head Mixture-of-Experts}.
\newblock In \emph{Advances in Neural Information Processing Systems
  (NeurIPS)}, 2024.

\bibitem[Xie et~al.(2025{\natexlab{a}})Xie, Chen, Chen, Cai, Tang, Lin, Zhang,
  Li, Zhu, Lu, and Han]{xie2024sana}
Enze Xie, Junsong Chen, Junyu Chen, Han Cai, Haotian Tang, Yujun Lin, Zhekai
  Zhang, Muyang Li, Ligeng Zhu, Yao Lu, and Song Han.
\newblock {SANA: Efficient High-Resolution Text-to-Image Synthesis with Linear
  Diffusion Transformers}.
\newblock In \emph{International Conference on Learning Representations
  (ICLR)}, 2025{\natexlab{a}}.

\bibitem[Xie et~al.(2025{\natexlab{b}})Xie, Wei, Cao, Zhao, Deng, Li, Dai, Gao,
  Chang, Yu, Zhao, Zhou, Xu, Zhang, Zeng, Hu, Wang, Yuan, Wang, and
  Liang]{xie2025mhc}
Zhenda Xie, Yixuan Wei, Huanqi Cao, Chenggang Zhao, Chengqi Deng, Jiashi Li,
  Damai Dai, Huazuo Gao, Jiang Chang, Kuai Yu, Liang Zhao, Shangyan Zhou, Zhean
  Xu, Zhengyan Zhang, Wangding Zeng, Shengding Hu, Yuqing Wang, Jingyang Yuan,
  Lean Wang, and Wenfeng Liang.
\newblock {mHC: Manifold-Constrained Hyper-Connections}.
\newblock \emph{arXiv preprint arXiv:2512.24880}, 2025{\natexlab{b}}.

\bibitem[Yang et~al.(2024)Yang, Chen, Wang, Liu, and Chen]{yang2023ddsm}
Shuai Yang, Yukang Chen, Luozhou Wang, Shu Liu, and Ying-Cong Chen.
\newblock {Denoising Diffusion Step-aware Models}.
\newblock In \emph{International Conference on Learning Representations
  (ICLR)}, 2024.

\bibitem[Yuan et~al.(2025)Yuan, Wang, Huang, Zhu, Zhou, Yu, and
  Min]{yuan2025race}
Yike Yuan, Ziyu Wang, Zihao Huang, Defa Zhu, Xun Zhou, Jingyi Yu, and Qiyang
  Min.
\newblock {Expert Race: A Flexible Routing Strategy for Scaling Diffusion
  Transformer with Mixture of Experts}.
\newblock In \emph{International Conference on Machine Learning (ICML)}, 2025.

\bibitem[Zhao et~al.(2025)Zhao, Han, Tang, Wang, Song, Huang, Wang, and
  You]{zhao2024dydit}
Wangbo Zhao, Yizeng Han, Jiasheng Tang, Kai Wang, Yibing Song, Gao Huang, Fan
  Wang, and Yang You.
\newblock {Dynamic Diffusion Transformer}.
\newblock In \emph{International Conference on Learning Representations
  (ICLR)}, 2025.

\bibitem[Zheng et~al.(2026)Zheng, Wang, Ma, Chen, Zhang, Balaji, Chen, Liu,
  Zhu, and Zhang]{zheng2025rcm}
Kaiwen Zheng, Yuji Wang, Qianli Ma, Huayu Chen, Jintao Zhang, Yogesh Balaji,
  Jianfei Chen, Ming-Yu Liu, Jun Zhu, and Qinsheng Zhang.
\newblock {Large Scale Diffusion Distillation via Score-Regularized
  Continuous-Time Consistency}.
\newblock In \emph{International Conference on Learning Representations
  (ICLR)}, 2026.

\bibitem[Zhu et~al.(2025)Zhu, Huang, Huang, Zeng, Mao, Wu, Min, and
  Zhou]{zhu2024hyper}
Defa Zhu, Hongzhi Huang, Zihao Huang, Yutao Zeng, Yunyao Mao, Banggu Wu, Qiyang
  Min, and Xun Zhou.
\newblock {Hyper-Connections}.
\newblock In \emph{International Conference on Learning Representations
  (ICLR)}, 2025.

\end{thebibliography}
